%% file: main.tex
\documentclass{article}

\usepackage{microtype}
\usepackage{graphicx}
\usepackage{subfigure}
\usepackage{booktabs} %

\usepackage{hyperref}

\usepackage[accepted]{icml2024}

\usepackage{amsmath}
\usepackage{amssymb}
\usepackage{mathtools}
\usepackage{amsthm}

\usepackage[capitalize,noabbrev]{cleveref}

\theoremstyle{plain}

\input{custom/custom_package}

\input{custom/math_commands}

\icmltitlerunning{Fine-Grained Causal Dynamics Learning}

\begin{document}

\twocolumn[
\icmltitle{Fine-Grained Causal Dynamics Learning with Quantization \\ for Improving Robustness in Reinforcement Learning}

\icmlsetsymbol{equal}{*}
\icmlsetsymbol{advise}{$\dagger$}

\begin{icmlauthorlist}
\icmlauthor{Inwoo Hwang}{aiis}
\icmlauthor{Yunhyeok Kwak}{aiis}
\icmlauthor{Suhyung Choi}{aiis}
\icmlauthor{Byoung-Tak Zhang}{aiis}
\icmlauthor{Sanghack Lee}{aiis,gsds}
\end{icmlauthorlist}

\icmlaffiliation{aiis}{AI Institute, Seoul National University}
\icmlaffiliation{gsds}{Graduate School of Data Science, Seoul National University}

\icmlcorrespondingauthor{Sanghack Lee}{sanghack@snu.ac.kr}
\icmlcorrespondingauthor{Byoung-Tak Zhang}{btzhang@snu.ac.kr}

\icmlkeywords{Machine Learning, ICML}

\vskip 0.3in
]

\printAffiliationsAndNotice{}  %

\begin{abstract}
\input{sections/00-abs}
\end{abstract}

\section{Introduction}
\input{sections/01-intro}

\section{Preliminaries}
\label{sec:preliminary}
\input{sections/02-preliminary}

\section{Fine-Grained Causal Dynamics Learning}
\label{sec:method}

\input{sections/03-method}

\section{Experiments}
\label{sec:experiments}
\input{sections/04-experiments}

\section{Discussions and Future Works}
\label{sec:discussion}
\input{sections/05-discussions}

\section{Conclusion}\label{sec:conclusion}
\input{sections/06-conclusion}

\section*{Impact Statement}
In real-world applications, model-based RL requires a large amount of data. As a large-scale dataset may contain sensitive information, it would be advisable to discreetly evaluate the models within simulated environments before their real-world deployment.

\section*{Acknowledgements}
We would like to thank Sujin Jeon and Hyundo Lee for the useful discussions. We also thank anonymous reviewers for their constructive comments. This work was partly supported by the IITP (RS-2021-II212068-AIHub/10\%, RS-2021-II211343-GSAI/10\%, 2022-0-00951-LBA/10\%, 2022-0-00953-PICA/20\%), NRF (RS-2023-00211904/20\%, RS-2023-00274280/10\%, RS-2024-00353991/10\%), and KEIT (RS-2024-00423940/10\%) grant funded by the Korean government.

\bibliography{mainbib}
\bibliographystyle{icml2024}

\newpage
\appendix
\onecolumn

\input{sections/90-appendix_A}
\input{sections/91-appendix_B}

\input{sections/92-appendix_C}

\end{document}

%% file: custom/custom_package.tex
\usepackage{amsthm}
\usepackage{amsmath, amssymb, dsfont}
\usepackage{mathrsfs}
\usepackage{graphicx}
\usepackage{wrapfig}
\usepackage{enumitem}

\usepackage{commath}
\usepackage{bbm}
\usepackage{booktabs}
\usepackage{verbatim}
\usepackage{subfigure}
\usepackage{multirow}
\usepackage{makecell}
\usepackage{bbm}
\usepackage{mathtools}
\usepackage{capt-of}
\usepackage{lipsum}
\usepackage{adjustbox}
\usepackage{arydshln}

\usepackage{thmtools,thm-restate}
\usepackage{apxproof}

\newtheorem{definition}{Definition}

\newtheorem{lemma}{Lemma}
\newtheorem{assumption}{Assumption}
\newtheorem{remark}{Remark}
\newtheorem{corollary}{Corollary}

\usepackage{cleveref}
\Crefname{figure}{Fig.}{Figs.}
\Crefname{section}{Sec.}{Sec.}
\Crefname{equation}{Eq.}{Eqs.}
\Crefname{proposition}{Prop.}{Props.}
\Crefname{prop}{Prop.}{Props.}
\Crefname{theorem}{Thm.}{Thms}
\Crefname{lemma}{Lemma}{Lemmas}
\Crefname{definition}{Def.}{Defs}
\Crefname{algorithm}{Alg.}{Algs}
\Crefname{remark}{Remark}{Remarks}
\Crefname{example}{Example}{Examples}
\Crefname{corollary}{Corollary}{Corollary}
\Crefname{Appendix}{App.}{Apps.}
\Crefname{assumption}{Assumption}{Assumptions}

\usepackage{pifont}
\usepackage{xcolor}
\usepackage{pdfrender}
\usepackage{colortbl}
\definecolor{forestgreen}{rgb}{0.13, 0.55, 0.13}

\usepackage{algorithm}
\usepackage{algorithmic}
\usepackage{fancyhdr}

\newcommand{\stdv}[1]{\scriptsize$\pm$#1}

\usepackage{color, colortbl}
\definecolor{Gray}{gray}{0.9}
\definecolor{lightgray}{rgb}{0.83, 0.83, 0.83}

\newcommand{\paragrapht}[1]{\noindent\textbf{#1}}

\newcommand\Perp{\protect\mathpalette{\protect\independenT}{\perp}} 
\def\independenT#1#2{\mathrel{\rlap{$#1#2$}\mkern3mu{#1#2}}}

\def\*#1{\mathbf{#1}}

%% file: custom/math_commands.tex
\usepackage{amsmath,amsfonts,bm}

\def\eqref#1{equation~\ref{#1}}

\def\1{\bm{1}}

\def\rvx{{\mathbf{x}}}

\def\rmT{{\mathbf{T}}}
\def\rmU{{\mathbf{U}}}
\def\rmV{{\mathbf{V}}}

\def\rmX{{\mathbf{X}}}
\def\rmY{{\mathbf{Y}}}

\DeclareMathAlphabet{\mathsfit}{\encodingdefault}{\sfdefault}{m}{sl}
\SetMathAlphabet{\mathsfit}{bold}{\encodingdefault}{\sfdefault}{bx}{n}

\def\gA{{\mathcal{A}}}

\def\gC{{\mathcal{C}}}
\def\gD{{\mathcal{D}}}
\def\gE{{\mathcal{E}}}
\def\gF{{\mathcal{F}}}
\def\gG{{\mathcal{G}}}
\def\gH{{\mathcal{H}}}
\def\gI{{\mathcal{I}}}

\def\gL{{\mathcal{L}}}
\def\gM{{\mathcal{M}}}

\def\gP{{\mathcal{P}}}

\def\gS{{\mathcal{S}}}
\def\gT{{\mathcal{T}}}

\def\gX{{\mathcal{X}}}

\newcommand{\E}{\mathbb{E}}

\DeclareMathOperator*{\argmax}{argmax}
\DeclareMathOperator*{\argmin}{argmin}

%% file: sections/00-abs.tex
Causal dynamics learning has recently emerged as a promising approach to enhancing robustness in reinforcement learning (RL). Typically, the goal is to build a dynamics model that makes predictions based on the causal relationships among the entities. Despite the fact that causal connections often manifest only under certain contexts, existing approaches overlook such fine-grained relationships and lack a detailed understanding of the dynamics. In this work, we propose a novel dynamics model that infers fine-grained causal structures and employs them for prediction, leading to improved robustness in RL. The key idea is to jointly learn the dynamics model with a discrete latent variable that quantizes the state-action space into subgroups. This leads to recognizing meaningful context that displays sparse dependencies, where causal structures are learned for each subgroup throughout the training. Experimental results demonstrate the robustness of our method to unseen states and locally spurious correlations in downstream tasks where fine-grained causal reasoning is crucial. We further illustrate the effectiveness of our subgroup-based approach with quantization in discovering fine-grained causal relationships compared to prior methods.

%% file: sections/01-intro.tex
Model-based reinforcement learning (MBRL) has showcased its capability of solving various sequential decision making problems \citep{Kaiser2020Model,schrittwieser2020mastering}. Since learning an accurate and robust dynamics model is crucial in MBRL, recent works incorporate the causal relationships between the environmental variables, such as objects and the agent, into dynamics learning  \citep{wang2022causal,ding2022generalizing}. Unlike the traditional dense models that employ the whole state and action variables to predict the future state, causal dynamics models infer the causal structure of the transition dynamics and make predictions based on it. Consequently, they are more robust to unseen states by discarding spurious dependencies.

Our motivation stems from the observation that causal connections often manifest only under certain contexts in many practical scenarios. Consider autonomous driving, where recognizing the traffic signal is crucial for its safety (e.g., stops at red lights). However, in the presence of a pedestrian on the road, it must stop, even with a green light, ignoring the signal, i.e., the traffic signal becomes \textit{locally spurious}. Therefore, such fine-grained causal reasoning will be crucial to the robustness of MBRL for its real-world deployment.

\input{figure/fig1}

Fine-grained causal relationships can be understood with local independence between the variables, which holds under certain contexts but does not hold in general \citep{boutilier2013contextspecific}. Our goal is to incorporate them into dynamics modeling by capturing meaningful contexts that exhibit more sparse dependencies than the entire domain. Unfortunately, prior causal dynamics models examining global independence (\Cref{fig:dynamics_learning_comparison}-(a)) cannot harness them. On the other hand, existing methods for discovering fine-grained relationships have focused on examining sample-specific dependencies \citep{pitis2020counterfactual,hwang2023on}  (\Cref{fig:dynamics_learning_comparison}-(b)). However, it is unclear under which circumstances the inferred dependencies hold, making them hard to interpret and challenging to generalize to unseen states.

In this work, we propose a dynamics model that infers fine-grained causal structures and employs them for prediction, leading to improved robustness in MBRL. For this, we establish a principled way to examine fine-grained causal relationships based on the quantization of the state-action space. Importantly, this provides a clear understanding of meaningful contexts displaying sparse dependencies (\Cref{fig:dynamics_learning_comparison}-(c)). However, this involves the optimization of the regularized maximum likelihood score over the quantization which is generally intractable. To this end, we present a practical differentiable method that jointly learns the dynamics model and a discrete latent variable that decomposes the state-action space into subgroups by utilizing vector quantization \citep{van2017neural}. Theoretically, we show that joint optimization leads to identifying meaningful contexts and fine-grained causal structures.

We evaluate our method on both discrete and continuous control environments where fine-grained causal reasoning is crucial. Experimental results demonstrate the superior robustness of our approach to locally spurious correlations and unseen states in downstream tasks compared to prior causal/non-causal approaches. Finally, we illustrate that our method infers fine-grained relationships in a more effective and robust manner compared to sample-specific approaches.

Our contributions are summarized as follows. 
\begin{itemize}%
    \item We establish a principled way to examine fine-grained causal relationships based on the quantization of the state-action space which offers an identifiability guarantee and better interpretability.
    \item We present a theoretically grounded and practical approach to dynamics learning that infers fine-grained causal relationships by utilizing vector quantization.
    \item We empirically demonstrate that the agent capable of fine-grained causal reasoning is more robust to locally spurious correlations and generalizes well to unseen states compared to past causal/non-causal approaches. 
\end{itemize}

%% file: figure/fig1.tex
\begin{figure}[t!]
\centering
\includegraphics[width=\linewidth]{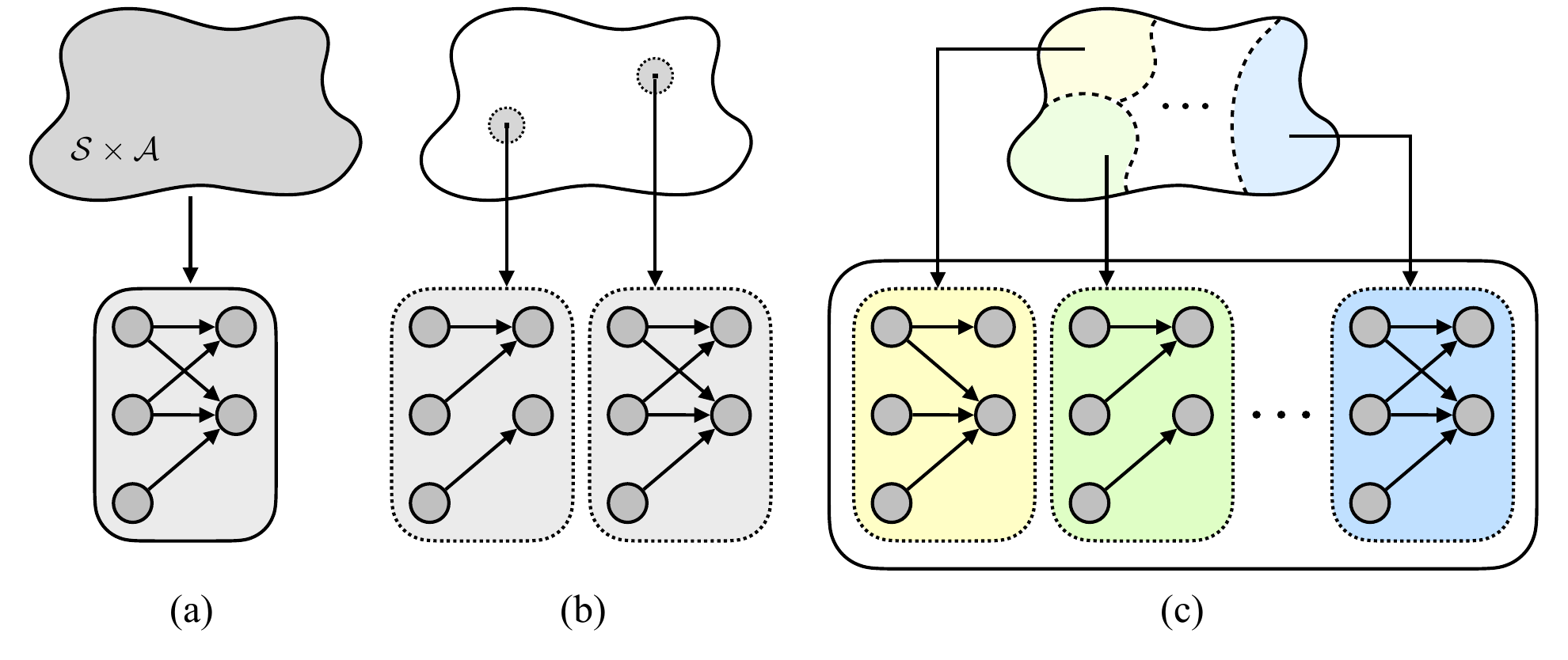}
\vspace{-20pt}
\caption{
(a) Previous causal dynamics models infer the global causal structure of the transition dynamics.
(b) Existing approaches to discovering fine-grained relationships examine individual samples.
(c) Our approach quantizes the state-action space into subgroups and infers causal relationships specific to each subgroup.
}
\label{fig:dynamics_learning_comparison}
\vspace{-10pt}
\end{figure}

%% file: sections/02-preliminary.tex
We first introduce the notations and terminologies. Then, we examine related works on causal dynamics learning for RL and fine-grained causal relationships.

\subsection{Background}

\paragrapht{Structural causal model.}
We adopt a framework of a structural causal model (SCM) \citep{pearl2009causality} to understand the relationship among variables. An SCM $\mathcal{M}$ is defined as a tuple $\left< \rmV, \mathbf{U}, \mathbf{F}, P(\mathbf{U}) \right>$, where $\rmV = \{X_1, \cdots, X_d\}$ is a set of endogenous variables and $\rmU$ is a set of exogenous variables. A set of functions $\mathbf{F} = \{f_1, \cdots, f_d\}$ determine how each variable is generated; $X_j = f_j(Pa(j), \*U_j)$ where $Pa(j) \subseteq \rmV \setminus \{X_j\}$ is parents of $X_j$ and $\*U_j \subseteq \*U$. An SCM $\gM$ induces a directed acyclic graph (DAG) $\gG=(V, E)$, i.e., a causal graph (CG) \citep{peters2017elements}, where $V=\{1, \cdots, d\}$ and $E \subseteq V \times V$ are the set of nodes and edges, respectively. Each edge denotes a direct causal relationship from $X_i$ to $X_j$. An SCM entails the conditional independence relationship of each variable (namely, local Markov property): $X_i \Perp ND(X_i) \mid Pa(X_i)$, where $ND(X_i)$ is a non-descendant of $X_i$, which can be read off from the corresponding causal graph.

\paragrapht{Factored Markov Decision Process.}
A Markov Decision Process (MDP) \citep{sutton2018reinforcement} is defined as a tuple $\left< \gS, \gA, T, r, \gamma \right>$ where $\gS$ is a state space, $\gA$ is an action space, $T: \gS \times \gA \rightarrow \gP(\gS)$ is a transition dynamics, $r$ is a reward function, and $\gamma$ is a discount factor. We consider a factored MDP \citep{kearns1999efficient} where the state and action spaces are factorized as $\gS = \gS_1 \times \cdots \times \gS_N$ and $\gA = \gA_1 \times \cdots \times \gA_M$. A transition dynamics is factorized as $p(s'\mid s, a) = \prod_j p(s'_j\mid s, a)$ where $s = (s_1, \cdots, s_N)$ and $a = (a_1, \cdots, a_M)$.

\paragrapht{Assumptions and notations.} 
We are concerned with an SCM associated with the transition dynamics in a factored MDP where the states are fully observable. To properly identify the causal relationships, we make assumptions standard in the field, namely, Markov property \citep{pearl2009causality}, faithfulness \citep{peters2017elements}, causal sufficiency \citep{spirtes2000causation}, and that causal connections only appear within consecutive time steps. With these assumptions, we consider a bipartite causal graph $\gG=(V, E)$ which consists of the set of nodes $V=\rmX\cup\rmY$ and the set of edges $E \subseteq \rmX \times \rmY$, where $\rmX = \{S_1, \cdots, S_N, A_1, \cdots, A_M\}$ and $\rmY = \{S'_1, \cdots, S'_N\}$. $Pa(j)$ denotes parent variables of $S'_j$. The conditional independence 
\begin{equation}\label{eq:ci}
S'_j \Perp \rmX \setminus {Pa(j)} \mid {Pa(j)},
\end{equation}
entailed by the causal graph $\gG$ represents the causal structure of the transition dynamics $p(s'\mid s, a) = \prod_j p(s'_j\mid Pa(j))$.

\paragrapht{Dynamics modeling.}
The traditional way is to use dense dependencies for dynamics modeling: $\prod_j p(s'_j\mid s, a)$. Causal dynamics models \citep{wang2021task,wang2022causal,ding2022generalizing} examine the causal structure $\gG$ to employ only relevant dependencies: $p(s'\mid s, a; \gG) = \prod_j p(s'_j\mid Pa(j))$ (\Cref{fig:dynamics_learning_comparison}-(a)). Consequently, they are more robust to spurious correlations and unseen states.

\subsection{Related Work}
\paragrapht{Causal dynamics models in RL.} There is a growing body of literature on the intersection of causality and RL \citep{de2019causal,buesing2018woulda,zhang2020invariant,sontakke2021causal,scholkopf2021toward,zholus2022factorized,zhang2020causal}. One focus is dynamics learning, which involves the causal structure of the transition dynamics \citep{li2020causal,yao2022learning,bongers2018causal,wang2022causal,ding2022generalizing,feng2022factored,huang2022action} (more broad literature on causal reasoning in RL is discussed in \Cref{appendix:preliminary-related_work}). Recent works proposed causal dynamics models that make robust predictions based on the causal dependencies (\Cref{fig:dynamics_learning_comparison}-(a)), utilizing conditional independence tests \citep{ding2022generalizing} or conditional mutual information \citep{wang2022causal} to infer the causal graph in a factored MDP. However, prior methods cannot harness fine-grained causal relationships that provide a more detailed understanding of the dynamics. In contrast, our work aims to discover and incorporate them into dynamics modeling, demonstrating that fine-grained causal reasoning leads to improved robustness in MBRL.

\paragrapht{Discovering fine-grained causal relationships.}
In the context of RL, a fine-grained structure of the environment dynamics has been leveraged in various ways, e.g., with data augmentation \citep{pitis2022mocoda}, efficient planning \citep{hoey1999spudd,chitnis2021camps}, or exploration \citep{seitzer2021causal}. For this, previous works often exploited domain knowledge \citep{pitis2022mocoda} or true dynamics model \citep{chitnis2021camps}. However, such prior knowledge is often unavailable in the context of dynamics learning. Existing methods for discovering fine-grained relationships examine the gradient \citep{wang2023elden} or attention score \citep{pitis2020counterfactual} of each sample (\Cref{fig:dynamics_learning_comparison}-(b)). However, such \textit{sample-specific} approaches lack an understanding of under which circumstances the inferred dependencies hold, and it is unclear whether they can generalize to unseen states.

In the field of causality, fine-grained causal relationships have been widely studied, e.g., context-specific independence \citep{boutilier2013contextspecific,zhang1999role,poole1998context,dal2018parallel,tikka2020identifying,jamshidi2024causal} (see \Cref{appendix:preliminary-local_independence} for the background). Recently, \citet{hwang2023on} proposed an auxiliary network that examines local independence for \textit{each sample}. However, it also does not explicitly capture the context where the local independence holds. In contrast to existing approaches relying on sample-specific inference \citep{lowe2022amortized,pitis2020counterfactual,hwang2023on}, we propose to examine causal dependencies at a subgroup level through quantization (\Cref{fig:dynamics_learning_comparison}-(c)), providing a more robust and principled way of discovering fine-grained causal relationships with a theoretical guarantee.

%% file: sections/03-method.tex
In this section, we first describe a brief background on local independence and intuition of our approach (\Cref{sec:method_preliminary}). We then provide a principled way to examine fine-grained causal relationships (\Cref{sec:method_problem_formulation}). Based on this, we propose a theoretically grounded and practical method for fine-grained causal dynamics modeling (\Cref{sec:method_overall}). Finally, we provide a theoretical analysis with discussions (\Cref{sec:method_theory}). All omitted proofs are provided in \Cref{appendix:theory}.

\subsection{Preliminary}\label{sec:method_preliminary}

Analogous to the conditional independence explaining the causal relationship between the variables (i.e., \Cref{eq:ci}), their fine-grained relationships can be understood with local independence \citep{hwang2023on}. This is written as:
\begin{equation}\label{eq:cssi}
S'_j \Perp \rmX \setminus  {Pa(j; \gD)} \mid {Pa(j; \gD)}, \gD,
\end{equation}
where $\gD \subseteq \gX = \gS \times \gA$ is a local subset of the joint state-action space, which we say \textit{context}, and ${Pa(j; \gD)}\subseteq \rmX$ is a set of state and action variables locally relevant for predicting $S'_j$ under $\gD$. We provide a formal definition and detailed background of local independence in \Cref{appendix:theory_preliminary}.

For example, consider a mobile home robot interacting with various objects ($Pa(j)$). Under the context of the \textit{door closed} ($\gD$), only objects within the same room ($Pa(j; \gD)$) become relevant. On the other hand, all objects remain relevant under the context of the \textit{door opened}. We say that a context is \textit{meaningful} if it displays sparse dependencies: $Pa(j; \gD) \subsetneq Pa(j)$, e.g., \textit{door closed}. We are concerned with the subgraph of the (global) causal graph $\gG$ as a graphical representation of such local dependencies.

\begin{restatable}[]{definition}{lcg}\label{def:lcg}
Local subgraph of the causal graph\footnote{For brevity, we will henceforth denote it as \textit{local causal graph}.} (LCG) on $\gD\subseteq\gX$ is 
$\gG_\gD = (V, E_\gD)$ where $E_\gD = \{(i, j) \mid i \in Pa(j; \gD) \}$.
\end{restatable}
LCG $\gG_\gD$ represents a causal structure of the transition dynamics specific to a certain context $\gD$. It is useful for our approach to fine-grained dynamics modeling, e.g., it is sufficient to consider only objects in the same room when the door is closed. In contrast, prior causal dynamics models consider all objects under any circumstances (\Cref{fig:dynamics_learning_comparison}-(a)).

\input{figure/fig2_overall}

Importantly, such information (e.g., $\gD$ and $\gG_{\gD}$) is not known in advance, and it is our goal to discover them. For this, existing sample-specific approaches have focused on inferring LCG directly from individual samples \citep{pitis2020counterfactual,hwang2023on} (\Cref{fig:dynamics_learning_comparison}-(b)). However, it is unclear under which context the inferred dependencies hold.

Our approach is to quantize the state-action space into subgroups and examine causal structures on \textit{each subgroup} (\Cref{fig:dynamics_learning_comparison}-(c)). This now makes it clear that each inferred LCG will represent fine-grained causal relationships under the corresponding subgroup. We now proceed to describe a principled way to discover LCGs based on quantization.

\subsection{Score for Decomposition and Graphs}\label{sec:method_problem_formulation}

Let us consider \textit{arbitrary} decomposition $\{\gE_z\}_{z=1}^K$ of the state-action space $\gX$, where $K$ is the degree of the quantization. The transition dynamics can be decomposed as:
\begin{align}\label{eq:decompose}
p(s'_j\mid s, a) 
&= \sum_z p(s'_j\mid s, a, z) p(z\mid s, a) \notag \\
&= \sum_z p(s'_j\mid {Pa(j; \gE_z)}, z) p(z\mid s, a),
\end{align}
where $p(z\mid s, a) = 1$ if $(s, a)\in \gE_z$. This illustrates our approach to fine-grained dynamics modeling, employing only {locally} relevant dependencies according to $\gG_{\gE_z}$ on each subgroup $\gE_z$. We now aim to learn each LCG $\gG_{\gE_z}$ based on \Cref{eq:decompose}. Specifically, we consider the regularized maximum likelihood score $\gS(\{\gG_z, \gE_z\}_{z=1}^K)$ of the graphs $\{\gG_z\}^K_{z=1}$ and decomposition $\{\gE_z\}^K_{z=1}$ which is defined as:
\begin{equation}
\label{eq:scorebasedmletwo}
\sup_\phi \mathbb{E}_{p(s, a, s')} \left[ \log \hat{p}(s'\mid s, a; \{\gG_z, \gE_z\}, \phi) - \lambda \lvert \gG_z \rvert  \right],
\end{equation}
where $\phi$ is the parameters of the dynamics model $\hat{p}$ which employs the graph $\gG_z$ for prediction on corresponding subgroup $\gE_z$. We now show that graphs that maximize the score faithfully represent causal dependencies on each subgroup.
\begin{restatable}[Identifiability of LCGs]{theorem}{fixedz}
\label{prop:identifiability-graph-fixed}
With \Cref{assumption:basic,assumption:localfaithfulness,assumption:capacity,assumption:finiteentropy}, let $\{\hat{\gG}_{z}\} \in \argmax \gS(\{\gG_z, \gE_z\}_{z=1}^K)$ for $\lambda>0$ small enough. Then, each $\hat{\gG}_z$ is true LCG on $\gE_z$, i.e., $\hat{\gG}_z = \gG_{\gE_z}$.
\end{restatable}
Given the subgroups, corresponding LCGs can be recovered by score maximization. Therefore, it provides a principled way to discover LCGs, which is valid for \textit{any} quantization. 

Unfortunately, not all quantization is useful for fine-grained dynamics modeling, e.g., by dividing into \textit{lights on} and \textit{lights off}, it still needs to consider all objects under both circumstances. Thus, it is crucial for quantization to capture \textit{meaningful} contexts displaying sparse dependencies. Such useful quantization will allow more sparse dynamics modeling, i.e., the higher score of \Cref{eq:scorebasedmletwo}. Therefore, the decomposition is now also a learning objective towards maximizing \Cref{eq:scorebasedmletwo}, i.e., 
$\{{\gG}^*_{z}, {\gE}^*_{z}\} \in \argmax \gS(\{\gG_z, \gE_z\}_{z=1}^K)$. However, a naive optimization with respect to decomposition is generally intractable. Thus, we devise a practical method allowing joint training with the dynamics model.

\subsection{Fine-Grained Causal Dynamics Learning with Quantization}
\label{sec:method_overall}

We propose a practical differentiable method that allows joint optimization of \Cref{eq:scorebasedmletwo} over dynamics model $\hat{p}$, decomposition $\{\gE_z\}$, and graphs $\{\gG_z\}$,  in an end-to-end manner. The key component is a discrete latent codebook $C=\{e_z\}$ where each code $e_z$ represents the pair of a subgroup $\gE_z$ and a graph $\gG_z$. The codebook learning is differentiable, and these pairs will be learned throughout the training with the dynamics model. The overall framework is shown in \Cref{fig:overall_framework}.

\paragrapht{Quantization.} 
The encoder $g_\texttt{enc}$ maps each sample $(s, a)$ into a latent embedding $h$, which is then quantized to the nearest prototype vector $e$ (i.e., code) in the codebook $C= \{e_1, \cdots, e_K\}$, following \citet{van2017neural}:
\begin{equation}\label{eq:quantize}
e = e_z, \quad \text{where} \,\,\,  z = \argmin_{j \in \left[K\right]} \|h - e_j\|_2.
\end{equation}
This entails the subgroups since each sample corresponds to exactly one of the codes, i.e., each code $e_z$ represents the subgroup $\gE_z = \{(s, a) \mid e = e_z\}$. Thus, this corresponds to the term $p(z\mid s, a)$ in \Cref{eq:decompose}. In other words, the codebook $C$ serves as a proxy for decomposition $\{\gE_z\}_{z=1}^K$.

\paragrapht{Local causal graphs.}
Quantized embedding $e$ is then decoded to an adjacency matrix $A\in \{0, 1\}^{(N+M)\times N}$. The output of the decoder $g_\texttt{dec}$ is the parameters of Bernoulli distributions from which the matrix is sampled: 
$A\sim g_\texttt{dec}(e)$. In other words, each code $e_z$ corresponds to the matrix $A_z$ that represents the graph $\gG_z$. To properly backpropagate gradients, we adopt Gumbel-Softmax reparametrization trick \citep{jang2016categorical,maddison2016concrete}.

\paragrapht{Dynamics learning.}
The dynamics model $\hat{p}$ employs the matrix $A$ for prediction: $\sum_j \log \hat{p}(s_j'\mid s, a; A^{(j)})$, where $A^{(j)}\in \{0, 1\}^{(N+M)}$ is the $j$-th column of $A$. Each entry of $A^{(j)}$ indicates whether the corresponding state or action variable will be used to predict the next state $s'_j$. This corresponds to the term $p(s'_j\mid Pa(j; \gE_z), z)$ in \Cref{eq:decompose}. For the implementation, we mask out the features of unused variables according to $A$. We found that this is more stable compared to the input masking \citep{brouillard2020differentiable}.

\paragrapht{Training objective.}
We employ a regularization loss $\lambda \cdot \|A\|_1$ to induce a sparse LCG, where $\lambda$ is a hyperparameter. To update the codebook, we use a quantization loss \citep{van2017neural}. The training objective is as follows:
\begin{align}\label{eq:total_loss}
\gL_\texttt{total}  
=&\,\, \underbrace{- \log \hat{p}(s'\mid s, a; A) + \lambda \cdot \|A\|_1}_{\gL_\texttt{pred}} \notag \\
&\,\,+ \underbrace{\| \text{sg}\left[h\right] - e\|_2^2 + \beta\cdot \| h - \text{sg}\left[e\right]\|_2^2}_{\gL_\texttt{quant}}.
\end{align}
Here, $\gL_\texttt{pred}$ is the masked prediction loss with regularization. $\gL_\texttt{quant}$ is the quantization loss where $\text{sg}\left[\cdot\right]$ is a stop-gradient operator and $\beta$ is a hyperparameter. Specifically, $\| \text{sg}\left[h\right] - e\|_2^2$ moves each code toward the center of the embeddings assigned to it and $\beta\cdot \| h - \text{sg}\left[e\right]\|_2^2$ encourages the encoder to output the embeddings close to the codes. This allows us to jointly train the dynamics model and the codebook in an end-to-end manner. Intuitively, vector quantization clusters the samples under a similar context and reconstructs the LCGs for each clustering. The rationale is that any error in the graph $\gG_z$ or clustering $\gE_z$ would lead to the prediction error of the dynamics model. We provide the details of our model in \Cref{appendix:implementation_details_ours}.

We note that prior works on learning a discrete latent codebook have mostly focused on the reconstruction of the observation \citep{van2017neural,ozair2021vector}. To the best of our knowledge, our work is the first to utilize vector quantization for discovering diverse causal structures.

\paragrapht{Discussion on the codebook collapsing.}
It is well known that training a discrete latent codebook with vector quantization often suffers from the codebook collapsing, where many codes learn the same output and converge to a trivial solution. For this, we employ exponential moving averages (EMA) to update the codebook, following \citet{van2017neural}. 
In practice, we found that the training was relatively stable for any choice of the codebook size $K>2$. In our experiments, we simply fixed it to 16 across all environments since they all performed comparably well, which we will demonstrate in \Cref{sec:experiments_results}.

\subsection{Theoretical Analysis and Discussions}
\label{sec:method_theory}

So far, we have described how our method learns the decomposition and LCGs through the discrete latent codebook $C$ as a proxy. Our method can be viewed as a practical approach towards the maximization of $\gS(\{\gG_z, \gE_z\}_{z=1}^K)$ since $\gL_\texttt{pred}$ corresponds to \Cref{eq:scorebasedmletwo} and $\gL_\texttt{quant}$ is a mean squared error in the latent space which can be minimized to $0$. In this section, we provide its implications and discussions.
\begin{restatable}[]{proposition}{propoptimality}
\label{prop:identifiability-graph-general}
Let $\{{\gG}^*_{z}, {\gE}^*_{z}\} \in \argmax \gS(\{\gG_z, \gE_z\}_{z=1}^K)$ for $\lambda>0$ small enough, with \Cref{assumption:basic,assumption:localfaithfulness,assumption:capacity,assumption:finiteentropy,assumption:inf}. Then, (i) each ${\gG}^*_z$ is true LCG on ${\gE}^*_z$, and (ii) $\E \big[ \lvert {\gG}^*_{z} \rvert \big] \leq \E \left[ \lvert \gG_{z} \rvert \right]$ where $\{{\gG}_z\}$ are LCGs on arbitrary decomposition $\{{\gE}_z\}^K_{z=1}$.
\end{restatable}
In other words, the decomposition that maximizes the score is optimal in terms of $\E \left[ \lvert \gG_{z} \rvert \right]$ = $\sum_z p(\gE_z) \lvert \gG_{z} \rvert$. This is an important property involving the contexts which are more likely (i.e., large $p(\gE)$) and more meaningful (i.e., sparse $\gG_\gE$). Therefore, \Cref{prop:identifiability-graph-general} implies that score maximization would lead to the fine-grained understanding of the dynamics \textit{at best} it can achieve given the quantization degree $K$.

We now illustrate how the optimal decomposition $\{{\gE}^*_z\}^K_{z=1}$ in \Cref{prop:identifiability-graph-general} with sufficient quantization degree identifies important context $\gD$ (e.g., \textit{door closed}) that displays fine-grained causal relationships. We say the context $\gD$ is \textit{canonical} if $\gG_\gF = \gG_\gD$ for any $\gF \subset \gD$.
\begin{restatable}[Identifiability of contexts]{theorem}{identifiabilityeventgeneral}
\label{prop:identifiability-event-general}
Let $\{{\gG}^*_{z}, {\gE}^*_{z}\} \in \argmax \gS(\{\gG_z, \gE_z\}_{z=1}^K)$  for $\lambda>0$ small enough, with \Cref{assumption:basic,assumption:localfaithfulness,assumption:capacity,assumption:finiteentropy,assumption:inf}. Suppose $\gX = \cup_{m\in[H]} \gD_m$ where $\gG_{\gD_m}$ is distinct for all $m\in[H]$, and $\gD_1, \cdots, \gD_H$ are disjoint and canonical. Suppose $K\geq H$. Then, for all $m\in[H]$, there exists $I_m \subset [K]$ such that $\gD_m = \bigcup_{z\in I_m} {\gE}^*_z$ almost surely.
\end{restatable}
In other words, the joint optimization of \Cref{eq:scorebasedmletwo} over the quantization and dynamics model with sufficient quantization degree perfectly captures meaningful contexts that exist in the system (\Cref{prop:identifiability-event-general}) and recovers corresponding LCGs (\Cref{prop:identifiability-graph-general}-(i)), thereby leading to a fine-grained understanding of the dynamics. Our method described in the previous section serves as a practical approach toward this goal.

\paragrapht{Discussion on the codebook size.}
\Cref{prop:identifiability-event-general} also implies that the identification of the meaningful contexts is agnostic to the quantization degree $K$, as long as $K\geq H$. In \Cref{sec:experiments_results}, we demonstrate that our method works reasonably well for various quantization degrees in practice. We note that determining a minimal and sufficient number of quantization $H$ is \textit{not} our primary focus. This is because over-parametrization of quantization incurs only a small memory cost for additional codebook vectors in practice. Note that even if $K<H$, \Cref{prop:identifiability-graph-general}-(ii) guarantees that it would still discover meaningful fine-grained causal relationships, optimal in terms of $\E \left[ \lvert \gG_{z} \rvert \right]$. 

\paragrapht{Relationship to past approaches.}
To better understand our approach, we draw connections to (i) prior causal dynamics models and (ii) sample-specific approaches to discovering fine-grained dependencies. First, our method with the quantization degree $K=1$ degenerates to prior causal dynamics models \citep{wang2022causal,ding2022generalizing}: it would discover (global) causal dependencies (i.e., special case of \Cref{prop:identifiability-graph-fixed}) but cannot harness fine-grained relationships. Second, our method without quantization reverts to sample-specific approaches ($K \to\infty$), e.g., the auxiliary network that infers local independence directly from each sample \citep{hwang2023on}. As described earlier, it is unclear under which context the inferred dependencies hold. In \Cref{sec:experiments_results}, we demonstrate that this makes their inferences often inconsistent within the same context and prone to overfitting, while our approach with quantization infers fine-grained causal relationships in a more effective and robust manner.

%% file: figure/fig2_overall.tex
\begin{figure*}[t!]%
\centering
\includegraphics[width=.825\textwidth]{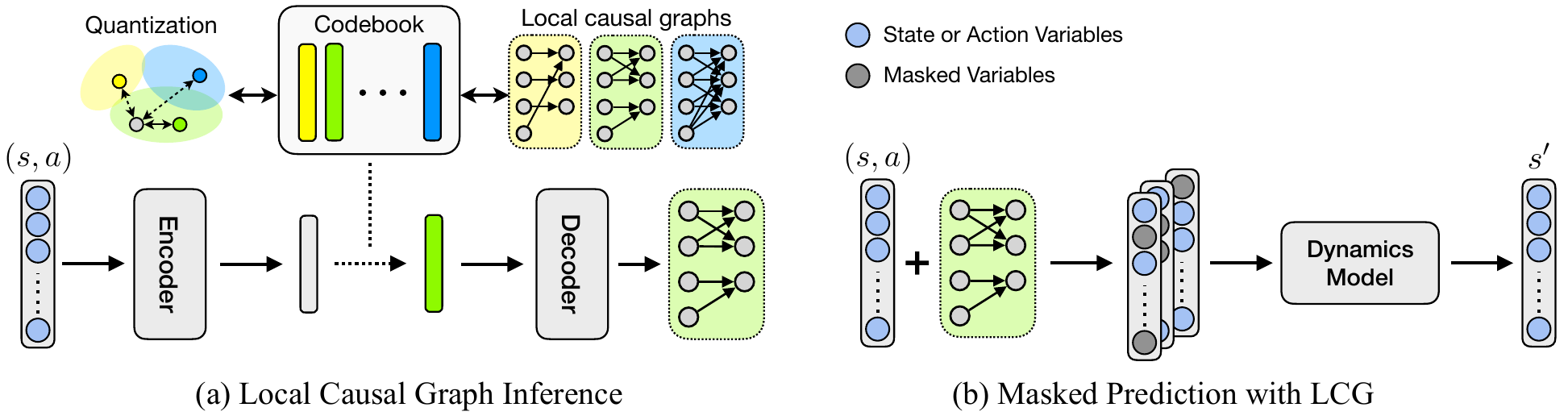}
\caption{Overall framework. 
(a) For each sample $(s, a)$, our method determines the subgroup to which the sample belongs through quantization and infers the local causal graph (LCG) that represents fine-grained causal relationships specific to the corresponding subgroup. (b) The dynamics model predicts the future state based on the inferred LCG. All components (e.g., dynamics model and codebook) are jointly learned throughout the training in an end-to-end manner.
}
\vspace{-5pt}
\label{fig:overall_framework}
\end{figure*}

%% file: sections/04-experiments.tex
In this section, we evaluate our method, coined Fine-Grained Causal Dynamics Learning (\textbf{FCDL}), to investigate the following questions:
(1) Does our method improve robustness in MBRL (\Cref{table:main_result,table:chemical_ood_accuracy})?
(2) Does our method discover fine-grained causal relationships and capture meaningful contexts (\Cref{fig:lcg_chemical_k04,fig:lcg_magnetic,fig:magnetic_shd})?
(3) Is our method more effective and robust compared to sample-specific approaches (\Cref{fig:lcg_magnetic,fig:magnetic_shd})?
(4) How does the degree of quantization affect performance (\Cref{fig:magnetic_shd,table:ablation_codebook})?

\subsection{Experimental Setup}\label{sec:experiments_setup}
The environments are designed to exhibit fine-grained causal relationships under a particular context $\gD$. The state variables (e.g., position, velocity) are fully observable, following prior works \citep{ding2022generalizing,wang2022causal,seitzer2021causal,pitis2020counterfactual,pitis2022mocoda}. Experimental details are provided in \Cref{appendix:experiment}.\footnote{Our code is publicly available at \url{https://github.com/iwhwang/Fine-Grained-Causal-RL}.}

\input{figure/fig_dataset}

\input{table/main_result}

\subsubsection{Environments}

\paragrapht{Chemical \textnormal{\citep{ke2021systematic}}.}
It is a widely used benchmark for systematic evaluation of causal reasoning in RL. There are 10 nodes, each colored with one of 5 colors. According to the underlying causal graph, an action changes the colors of the intervened node's descendants as depicted in \Cref{fig:dataset_chemical}. The task is to match the colors of each node to the given target. We designed two settings, named \textit{full-fork} and \textit{full-chain}. In both settings, the underlying CG is both \textit{full}. When the color of the root node is red ($\gD$), the colors change according to \textit{fork} or \textit{chain}, respectively ($\gG_{\gD}$).  For example, in \textit{full-fork}, all other parent nodes except the root become irrelevant under this context. Otherwise ($\gD^c$), the transition respects the graph \textit{full} (i.e., $\gG_{\gD^c}=\gG$). During the test, the root color is set to red, and LCG (\textit{fork} or \textit{chain}) is activated. Here, the agent receives a noisy observation for some nodes, and the task is to match the colors of other clean nodes, as depicted in \Cref{appendix:task_setup} (\Cref{fig:dataset_detail_chemical}). The agent capable of fine-grained causal reasoning would generalize well since corrupted nodes are locally spurious to predict other nodes.

\paragrapht{Magnetic.}
We designed a robot arm manipulation environment based on the Robosuite framework \citep{zhu2020robosuite}. There is a moving ball and a box on the table, colored red or black (\Cref{fig:dataset_magnetic}). Red color indicates that the object is \textit{magnetic}, and attracts the other magnetic object. For example, when both are red, magnetic force will be applied, and the ball will move toward the box. Otherwise, i.e., under the non-magnetic context, the box would have no influence on the ball. The color and position of the objects are randomly initialized for each episode, i.e., each episode is under the magnetic or non-magnetic context during training. The task is to reach the ball, predicting its trajectory. In this environment, non-magnetic context $\gD$ displays sparse dependencies ($\gG_\gD \subsetneq \gG$) because the box no longer influences the ball under this context.  In contrast, all causal dependencies remain the same under the magnetic context $\gD^c$, i.e., $\gG_{\gD^c} = \gG$. CG and LCGs are shown in \Cref{appendix:task_setup} (\Cref{fig:magnetic_gt}). During the test, one of the objects is black, and the box is located at an unseen position. Under this non-magnetic context, the box becomes locally spurious, and thus, the agent aware of fine-grained causal relationships would generalize well to unseen out-of-distribution (OOD) states.

\subsubsection{Experimental details}
\paragrapht{Baselines.}
We first consider dense models, i.e., a monolithic network implemented as MLP which learns $p(s'\mid s, a)$, and a modular network having a separate network for each variable: $\prod_j p(s'_j \mid s, a)$.
We also include a graph neural network (GNN) \citep{Kipf2020Contrastive}, which learns the relational information, and NPS \citep{goyal2021neural}, which learns sparse and modular dynamics. Causal models, including CDL \citep{wang2022causal} and  GRADER \citep{ding2022generalizing}, infer causal structure for dynamics learning: $\prod_j p(s'_j \mid Pa(j))$. We also consider an \textit{oracle} model, which leverages the ground truth (global) causal graph. Finally, we compare to NCD \citep{hwang2023on}, a sample-specific approach that examines local independence for each sample.

\paragrapht{Planning algorithm.} For all baselines and our method, we use a model predictive control \citep{camacho2013model} which selects the actions based on the prediction of the learned dynamics model.
Specifically, we use the cross-entropy method (CEM) \citep{rubinstein2004cross}, which iteratively generates and optimizes action sequences.

\paragrapht{Implementation.}
For our method, we set the hyperparameters $K=16, \lambda = 0.001$, and $\beta=0.25$ in all experiments. All methods have a similar model capacity for a fair comparison. For the evaluation, we ran 10 test episodes for every 40 training episodes. The results are averaged over eight different runs. All learning curves are shown in \Cref{appendix:additional_experiments}.

\input{table/chemical_ood_accuracy}

\input{figure/learning_curve}

\input{figure/lcg_chemical_k04}

\input{figure/lcg_magnetic}

\subsection{Results}
\label{sec:experiments_results}

\paragrapht{Downstream task performance {(\Cref{table:main_result}, \Cref{fig:learning_curve})}.}
All methods show similar performance on in-distribution (ID) states in training. However, dense models suffer from OOD states in the downstream tasks. Causal models are generally more robust compared to dense models, as they infer the causal graph and discard spurious dependencies. NCD, a sample-specific approach to infer fine-grained dependencies, performs better than causal models on a few downstream tasks, but not always. In contrast, our method consistently outperforms the baselines across all downstream tasks. This empirically validates our hypothesis that fine-grained causal reasoning leads to improved robustness in MBRL.

\paragrapht{Prediction accuracy (\Cref{table:chemical_ood_accuracy}).} 
To better understand the robustness of our method in downstream tasks, we investigate the prediction accuracy on ID and OOD states over the clean nodes in Chemical. As described earlier, noisy nodes are irrelevant for predicting the clean nodes under the LCG (i.e., \textit{fork} or \textit{chain}); thus, they are \textit{locally spurious} on OOD states in downstream tasks. While all methods perform reasonably well on ID states, dense models show a significant performance drop under the presence of noisy variables, merely above $20\%$ which is an expected accuracy of random guessing. As expected, causal dynamics models tend to be more robust compared to dense models, but they still suffer from OOD states. NCD is more robust than causal models when $n=2$, but eventually becomes similar to them as the number of noisy nodes increases. In contrast, our method outperforms baselines by a large margin across all downstream tasks, which demonstrates its effectiveness and robustness in fine-grained causal reasoning.

\paragrapht{Recognizing important contexts and fine-grained causal relationships (\Cref{fig:lcg_chemical_k04}).}
To illustrate the fine-grained causal reasoning of our method, we closely examine the behavior of our model with the quantization degree $K=4$ in {Chemical} (\textit{full-fork}, $n=2$). Recall each code corresponds to the pair of a subgroup and LCG, \Cref{fig:lcg_chemical_k04_a} shows how ID samples in the batch are allocated to one of the four codes. Interestingly, ID samples corresponding to LCG \textit{fork} are all allocated to the last code (\Cref{fig:lcg_chemical_k04_b}), i.e., the subgroup corresponding to the last code identifies this context. Furthermore, LCG decoded from this code (\Cref{fig:lcg_chemical_k04_e}) accurately captures the true \textit{fork} structure (\Cref{fig:lcg_chemical_k04_d}). This demonstrates that our method successfully recognizes meaningful context and fine-grained causal relationships. Notably, \Cref{fig:lcg_chemical_k04_c} shows that most of the OOD samples under \textit{fork} are correctly allocated to the last code. This illustrates the robustness of our method, i.e., its inference is consistent between ID and OOD states. Additional examples, including the visualization of all the learned LCGs from all codes, are provided in \Cref{appendix:additional_experiments}.

\input{table/ablation_and_shd}

\paragrapht{Inferred LCGs compared to sample-specific approach (\Cref{fig:lcg_magnetic}).}
We investigated the effectiveness and robustness of our method in fine-grained causal reasoning compared to the sample-specific approach. For this, we examine the inferred LCGs in Magnetic, where true LCGs and CG are shown in \Cref{appendix:task_setup} (\Cref{fig:magnetic_gt}). First, our method accurately learns LCG under the non-magnetic context (\Cref{fig:lcg_magnetic_b}). On the other hand, the LCG inferred by NCD is rather inaccurate (\Cref{fig:lcg_magnetic_c}), including some locally spurious dependencies (3 among 6 red boxes). Furthermore, its inference is inconsistent between ID and OOD states in the same non-magnetic context and completely fails on OOD states (\Cref{fig:lcg_magnetic_d}). This demonstrates that our approach is more effective and robust in discovering fine-grained causal relationships.

\paragrapht{Evaluation of local causal discovery (\Cref{fig:magnetic_shd}).}
We evaluate our method and NCD using structural hamming distance (SHD) in Magnetic. For each sample, we compare the inferred LCG with the true LCG based on the magnetic/non-magnetic context, and the SHD scores are averaged over the data samples in the evaluation batch. As expected, our method infers fine-grained relationships more accurately and maintains better performance on OOD states across various quantization degrees, which validates its effectiveness and robustness compared to NCD. Lastly, we note that our method with the quantization degree $K=1$ would learn only a single CG over the entire data domain, as shown in \Cref{fig:lcg_magnetic_a}. This explains its mean SHD score of 6 in non-magnetic samples in \Cref{fig:magnetic_shd}, since CG includes six redundant edges in non-magnetic context (i.e., red boxes in \Cref{fig:magnetic_gt}).

\paragrapht{Ablation on the quantization degree (\Cref{table:ablation_codebook}).}
Finally, we observe that our method works reasonably well across various quantization degrees on all downstream tasks in Chemical (\textit{full-fork}). Our method consistently outperforms the prior causal dynamics model (CDL) and sample-specific approach (NCD), which corroborates the results in \Cref{fig:magnetic_shd}. During our experiments, we found that the training was relatively stable for any quantization degree of $K>2$. We also found that instability often occurs under $K=2$, where the samples frequently fluctuate between two prototype vectors and result in the codebook collapsing. This is also shown in \Cref{table:ablation_codebook} where the performance of $K=2$ is worse compared to other choices of $K$. We speculate that over-parametrization of quantization could alleviate such fluctuation in general.

%% file: figure/fig_dataset.tex
\begin{figure}[t!]
\centering
\subfigure[{Chemical}]{
\centering
\includegraphics[clip,trim=5mm 5mm 5mm 5mm,height=0.4\linewidth]{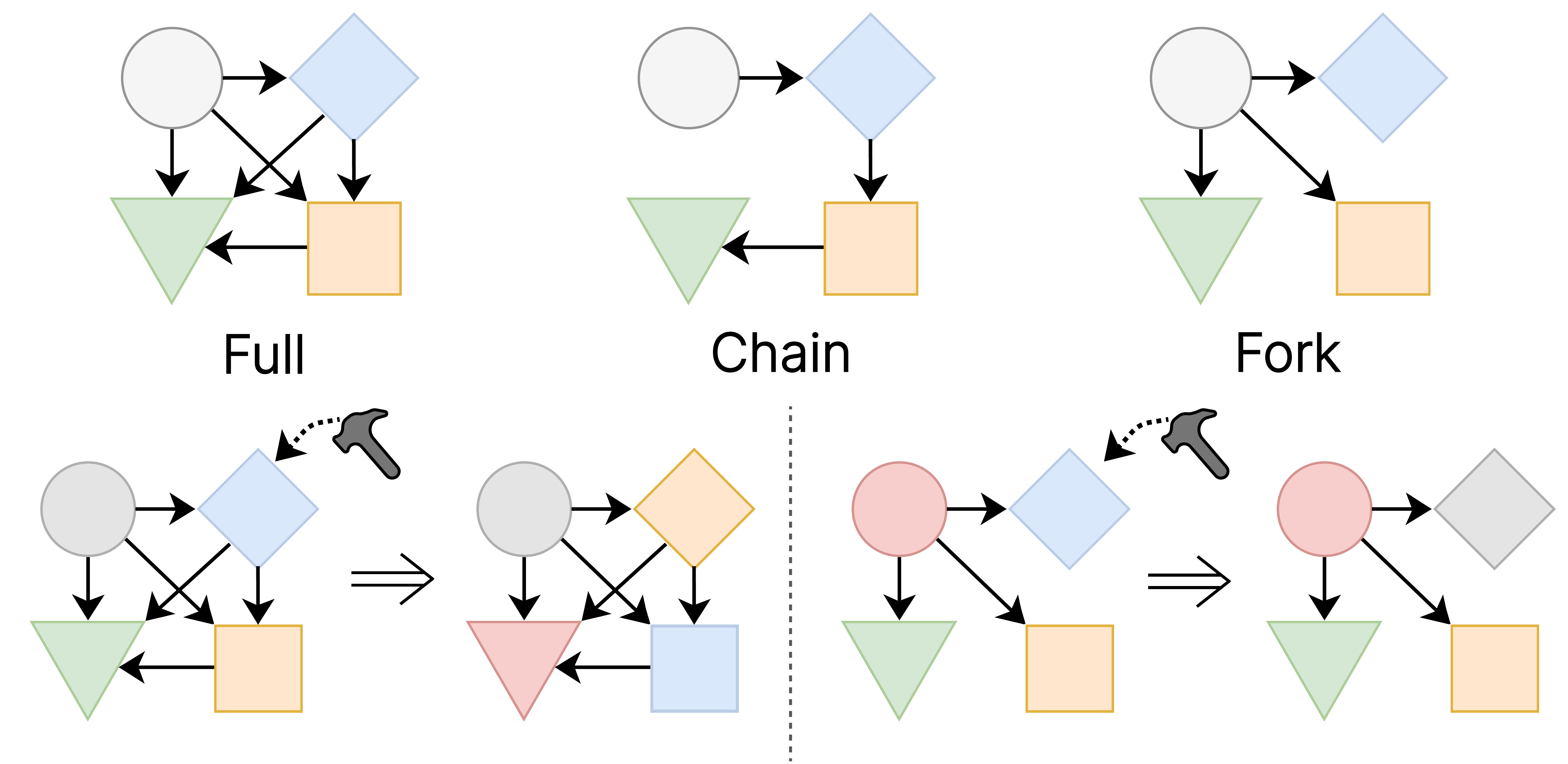}
\label{fig:dataset_chemical}
}%
\\
\vspace{-5pt}
\subfigure[{Magnetic}]{
\centering
\includegraphics[height=0.4\linewidth]{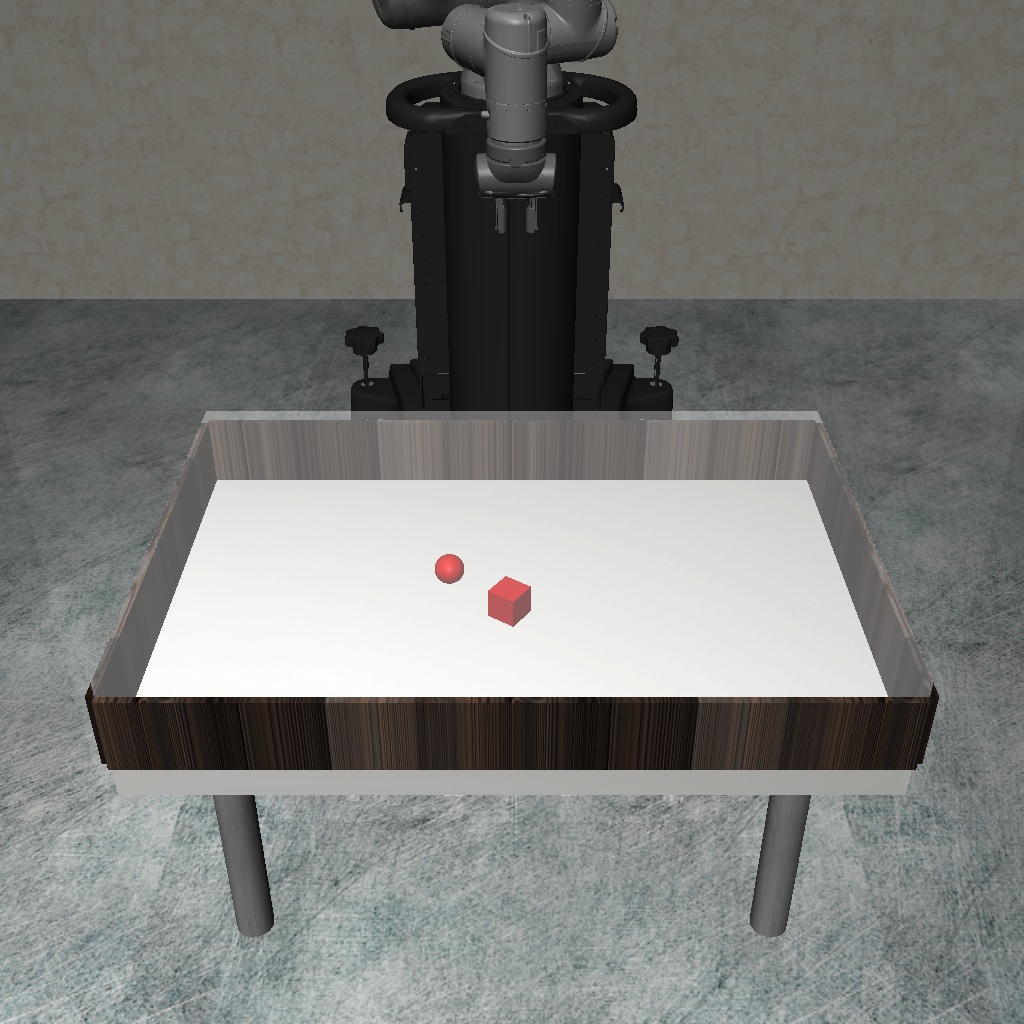}\hspace{1mm}%
\includegraphics[height=0.4\linewidth]{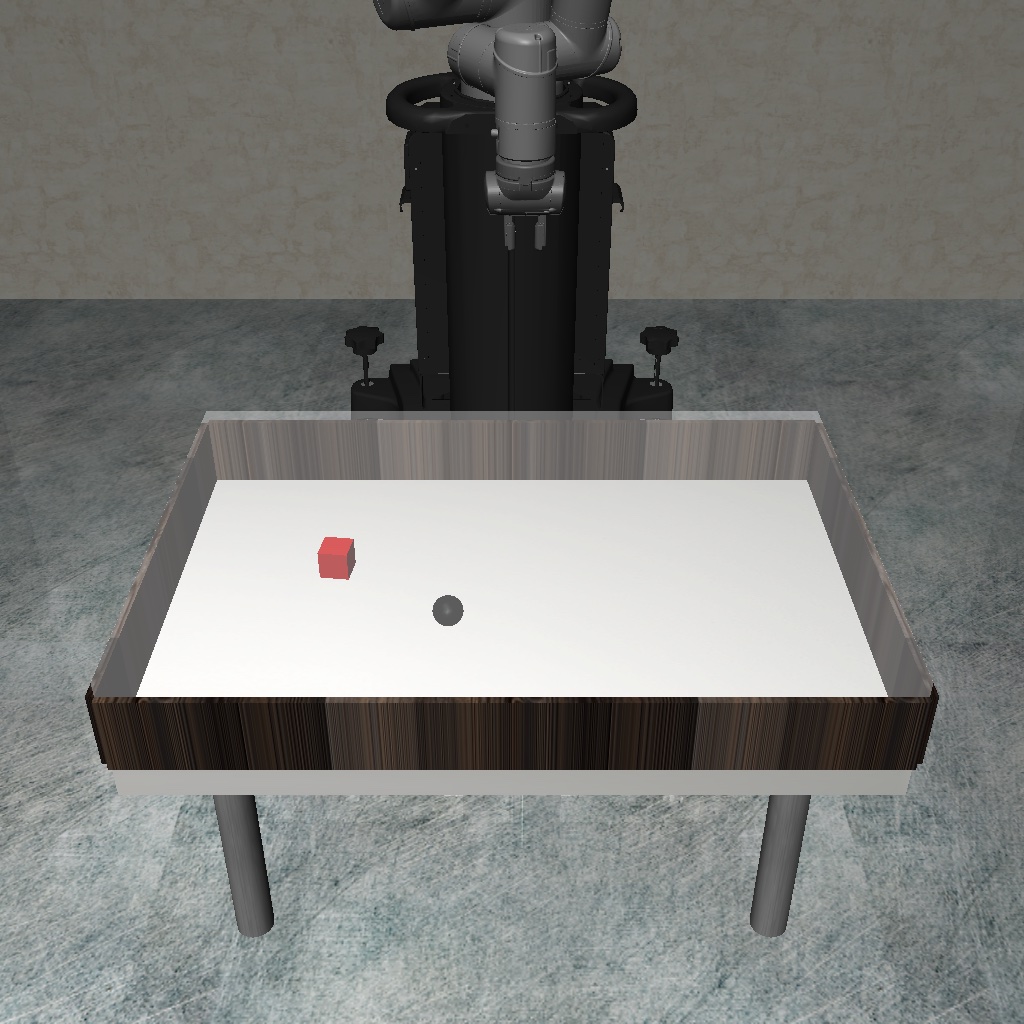}
\label{fig:dataset_magnetic}
}
\vspace{-5pt}
\caption{Illustrations for each environment. 
(a) In {Chemical}, colors change by the action according to the underlying causal graph. 
(b) In {Magnetic}, the red object exhibits magnetism.
}
\label{fig:dataset}
\vspace{-5pt}
\end{figure}

%% file: table/main_result.tex
\begin{table*}[t!]
\caption{Average episode reward on training and downstream tasks in each environment. In Chemical, $n$ denotes the number of noisy nodes in downstream tasks.
}
\label{table:main_result}
\centering
\begin{adjustbox}{width=\textwidth}%
\begin{tabular}{@{}lcccccccccccc@{}}
\toprule
& \multicolumn{4}{c}{\makecell{Chemical (\textit{full-fork})}} & \multicolumn{4}{c}{\makecell{Chemical (\textit{full-chain})}} & \multicolumn{2}{c}{\makecell{Magnetic}} \\
\cmidrule(lr){2-5} \cmidrule(lr){6-9} \cmidrule(lr){10-11}
Methods & \makecell[c]{Train\\ ($n=0$)}
& \makecell[c]{Test\\ ($n=2$)}
& \makecell[c]{Test\\ ($n=4$)}
& \makecell[c]{Test\\ ($n=6$)}
& \makecell[c]{Train\\ ($n=0$)}
& \makecell[c]{Test\\ ($n=2$)}
& \makecell[c]{Test\\ ($n=4$)}
& \makecell[c]{Test\\ ($n=6$)}
& Train
& Test
\\
\midrule
MLP & 	{19.00}\stdv{0.83} &	{6.49}\stdv{0.48} &	{5.93}\stdv{0.71} &	{6.84}\stdv{1.17} &	{17.91}\stdv{0.87} &	{7.39}\stdv{0.65} &	{6.63}\stdv{0.58} &	{6.78}\stdv{0.93}	& {8.37}\stdv{0.74}	& {0.86}\stdv{0.45}	\\
Modular & 	{18.55}\stdv{1.00} &	{6.05}\stdv{0.70} &	{5.65}\stdv{0.50} &	{6.43}\stdv{1.00} &	{17.37}\stdv{1.63} &	{6.61}\stdv{0.63} &	{7.01}\stdv{0.55} &	{7.04}\stdv{1.07}	& {8.45}\stdv{0.80}	& {0.88}\stdv{0.52}	\\
GNN \citep{Kipf2020Contrastive} & 	{18.60}\stdv{1.19} &	{6.61}\stdv{0.92} &	{6.15}\stdv{0.74} &	{6.95}\stdv{0.78} &	{16.97}\stdv{1.85} &	{6.89}\stdv{0.28} &	{6.38}\stdv{0.28} &	{6.56}\stdv{0.53}	& {8.53}\stdv{0.83}	& {0.92}\stdv{0.51}	\\
NPS \citep{goyal2021neural}	& {7.71}\stdv{1.22}	& {5.82}\stdv{0.83}	& {5.75}\stdv{0.57}	& {5.54}\stdv{0.80}	& {8.20}\stdv{0.54}	& {6.92}\stdv{1.03}	& {6.88}\stdv{0.79}	& {6.80}\stdv{0.39}	& {3.13}\stdv{1.00}	& {0.91}\stdv{0.69}	\\
CDL \citep{wang2022causal} & 	{18.95}\stdv{1.40} &	{9.37}\stdv{1.33} &	{8.23}\stdv{0.40} &	{9.50}\stdv{1.18} &	{17.95}\stdv{0.83} &	{8.71}\stdv{0.55} &	{8.65}\stdv{0.38} &	{10.23}\stdv{0.50}	& \textbf{8.75}\stdv{0.69}	& {1.10}\stdv{0.67}	\\
GRADER \citep{ding2022generalizing}	& {18.65}\stdv{0.98}	& {9.27}\stdv{1.31}	& {8.79}\stdv{0.65}	& {10.61}\stdv{1.31}	& {17.71}\stdv{0.54}	& {8.69}\stdv{0.56}	& {8.75}\stdv{0.80}	& {10.14}\stdv{0.33}	& -	& -	\\
Oracle	& \textbf{19.64}\stdv{1.18}	& {7.83}\stdv{0.87}	& {8.04}\stdv{0.62}	& {9.66}\stdv{0.21}	& {17.79}\stdv{0.76}	& {8.47}\stdv{0.69}	& {8.85}\stdv{0.78}	& {10.29}\stdv{0.37}	& {8.42}\stdv{0.86}	& {0.95}\stdv{0.55}	\\
NCD \citep{hwang2023on}	& {19.30}\stdv{0.95}	& {10.95}\stdv{1.63}	& {9.11}\stdv{0.63}	& {10.32}\stdv{0.93}	& \textbf{18.27}\stdv{0.27}	& {9.60}\stdv{1.52}	& {8.86}\stdv{0.23}	& {10.32}\stdv{0.37}	& {8.48}\stdv{0.70}	& {1.31}\stdv{0.77}	\\
FCDL (Ours)	& {19.28}\stdv{0.87}	& \textbf{15.27}\stdv{2.53}	& \textbf{14.73}\stdv{1.68}	& \textbf{13.62}\stdv{2.56} & 	{17.22}\stdv{0.61} &	\textbf{13.36}\stdv{3.60} &	\textbf{12.35}\stdv{3.23} &	\textbf{12.00}\stdv{1.21}	& {8.52}\stdv{0.74}	& \textbf{4.81}\stdv{3.01}	\\
\bottomrule
\end{tabular}
\end{adjustbox}
\vspace{-3mm}
\end{table*}

%% file: table/chemical_ood_accuracy.tex
\begin{table*}[t!]
\caption{Prediction accuracy on ID ($n=0$) and OOD ($n=2, 4, 6$) states in {Chemical} environment. 
}
\label{table:chemical_ood_accuracy}
\centering
\begin{adjustbox}{width=0.95\linewidth}\setlength{\tabcolsep}{7pt}
\begin{tabular}{@{}clccccccccc@{}}
\toprule
\multicolumn{2}{c}{\makecell{Setting / $n$}}
& MLP 
& Modular
& GNN %
& NPS %
& CDL %
& GRADER %
& Oracle
& NCD
& FCDL (Ours) %
\\
\midrule
\multirow{4}{*}{\makecell{\textit{full-fork}}}
& 	($n=0$)	& {88.31}\stdv{1.58}	& {89.24}\stdv{1.52}	& {88.81}\stdv{1.44}	& {58.34}\stdv{2.08}	& {89.22}\stdv{1.67}	& {87.75}\stdv{1.64}	& {89.63}\stdv{1.62}	& \textbf{90.07}\stdv{1.22}	& {89.46}\stdv{1.40}	\\
& 	($n=2$)	& {31.11}\stdv{1.69}	& {26.53}\stdv{3.45}	& {36.29}\stdv{3.45}	& {40.56}\stdv{4.61}	& {35.59}\stdv{1.85}	& {37.93}\stdv{1.06}	& {33.87}\stdv{1.34}	& {41.60}\stdv{5.08}	& \textbf{66.44}\stdv{12.22}	\\
& 	($n=4$)	& {30.44}\stdv{2.28}	& {24.73}\stdv{5.61}	& {25.80}\stdv{3.48}	& {26.81}\stdv{4.37}	& {35.82}\stdv{1.40}	& {38.94}\stdv{1.63}	& {36.48}\stdv{1.80}	& {37.47}\stdv{2.13}	& \textbf{58.49}\stdv{10.20}	\\
& 	($n=6$)	& {32.39}\stdv{1.76}	& {26.73}\stdv{8.31}	& {21.58}\stdv{3.44}	& {23.02}\stdv{4.27}	& {42.22}\stdv{1.39}	& {45.74}\stdv{2.25}	& {42.47}\stdv{0.75}	& {42.27}\stdv{1.82}	& \textbf{49.09}\stdv{4.77}	\\
\midrule
\multirow{4}{*}{\makecell{\textit{full-chain}}} 
& 	($n=0$)	& {84.38}\stdv{1.31}	& {85.92}\stdv{1.15}	& {85.41}\stdv{1.84}	& {58.48}\stdv{2.81}	& \textbf{86.85}\stdv{1.47}	& {84.24}\stdv{1.22}	& {85.76}\stdv{1.56}	& {85.63}\stdv{1.01}	& {86.07}\stdv{1.62}	\\
& 	($n=2$)	& {28.66}\stdv{3.65}	& {25.24}\stdv{4.68}	& {29.22}\stdv{3.39}	& {38.73}\stdv{2.63}	& {34.90}\stdv{1.59}	& {36.82}\stdv{3.12}	& {34.63}\stdv{1.78}	& {40.04}\stdv{6.21}	& \textbf{60.34}\stdv{12.10}	\\
& 	($n=4$)	& {26.52}\stdv{4.26}	& {24.94}\stdv{4.81}	& {23.28}\stdv{4.98}	& {27.69}\stdv{4.28}	& {36.52}\stdv{1.72}	& {37.41}\stdv{2.84}	& {38.31}\stdv{2.48}	& {37.47}\stdv{2.98}	& \textbf{56.64}\stdv{9.40}	\\
& 	($n=6$)	& {24.15}\stdv{4.17}	& {25.09}\stdv{5.91}	& {20.53}\stdv{6.96}	& {24.45}\stdv{3.84}	& {42.06}\stdv{1.29}	& {43.48}\stdv{4.14}	& {42.87}\stdv{2.08}	& {41.19}\stdv{1.66}	& \textbf{53.29}\stdv{6.63}	\\
\bottomrule
\end{tabular}
\end{adjustbox}
\vspace{-3mm}
\end{table*}

%% file: figure/learning_curve.tex
\begin{figure}[t]
\centering
\includegraphics[clip,trim=0 0mm 0 0,width=\linewidth]{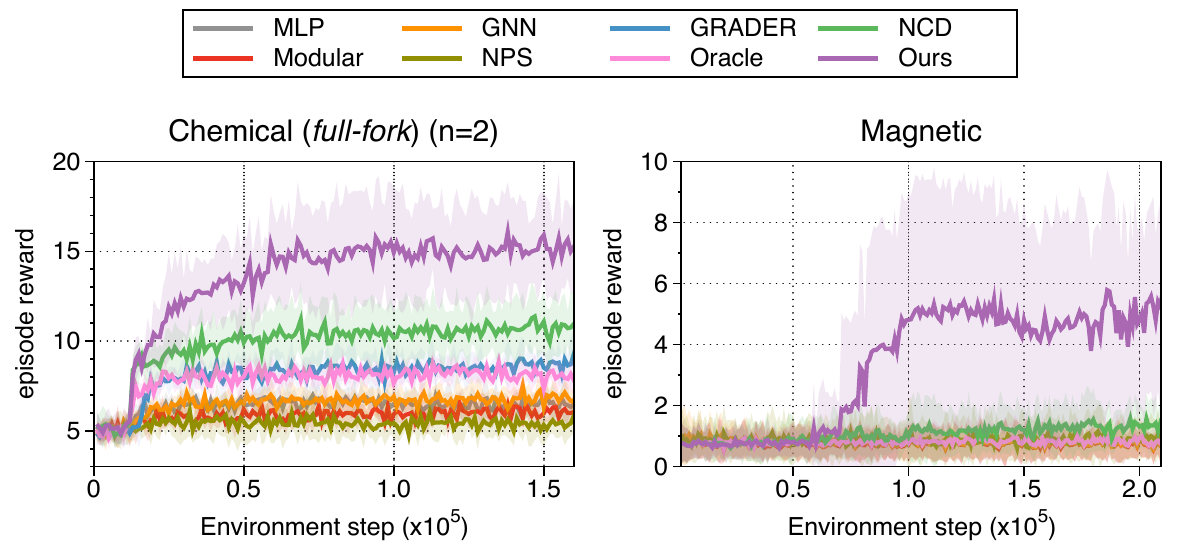}
\vspace{-7mm}
\caption{Learning curves on downstream tasks as measured on the average episode reward. Lines and shaded areas represent the mean and standard deviation, respectively.}
\vspace{-10pt}
\label{fig:learning_curve}
\end{figure}

%% file: figure/lcg_chemical_k04.tex
\begin{figure}[t]
    \centering
    \subfigure[ID (all)]{%
    \includegraphics[clip,trim=4mm 0mm 4mm 0mm, width=0.27\linewidth,height=0.25\linewidth]{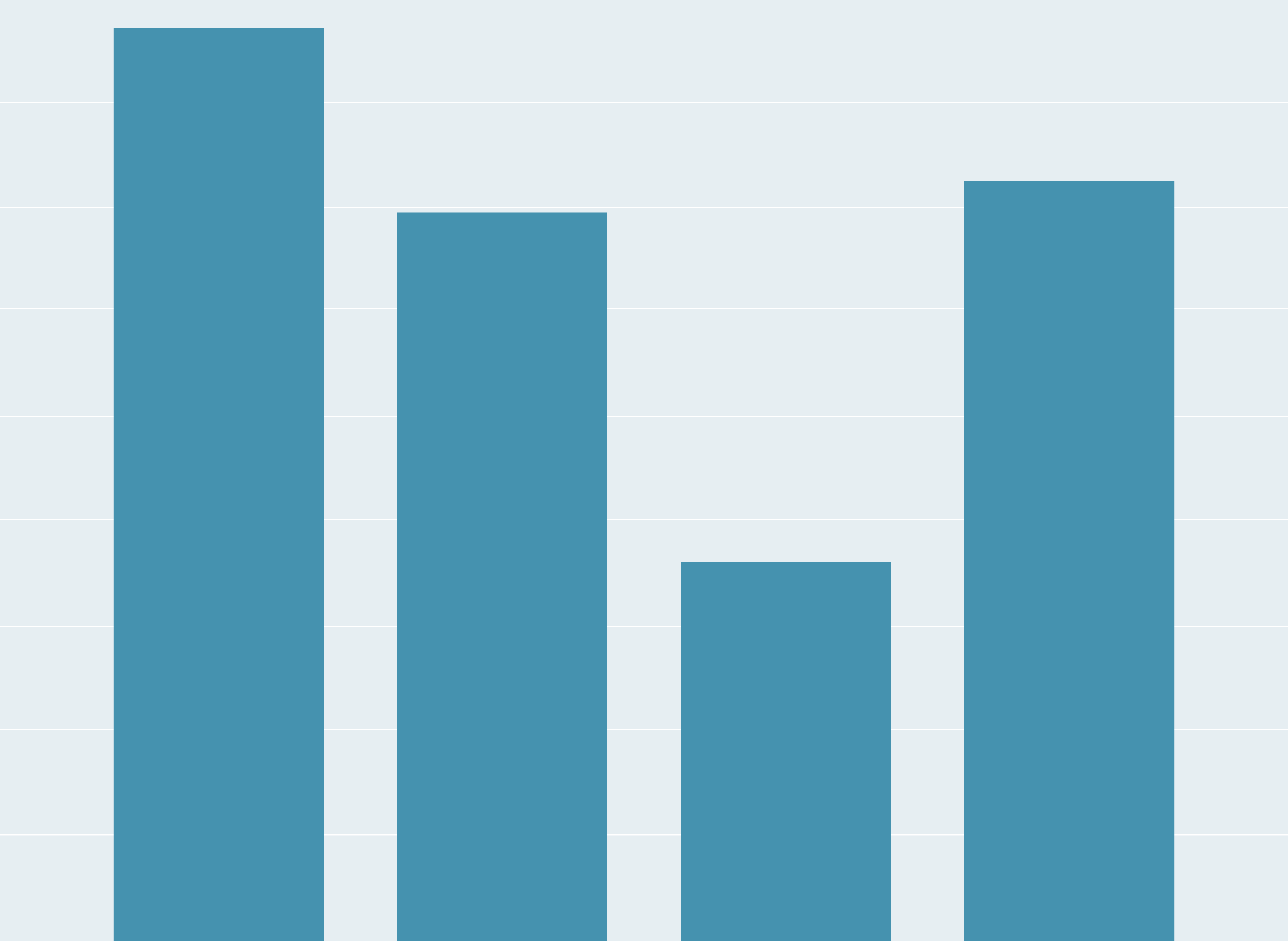}%
    \label{fig:lcg_chemical_k04_a}%
    }\hfil
    \subfigure[ID (\textit{fork})]{%
    \includegraphics[clip,trim=4mm 0mm 4mm 0, width=0.27\linewidth,height=0.25\linewidth]{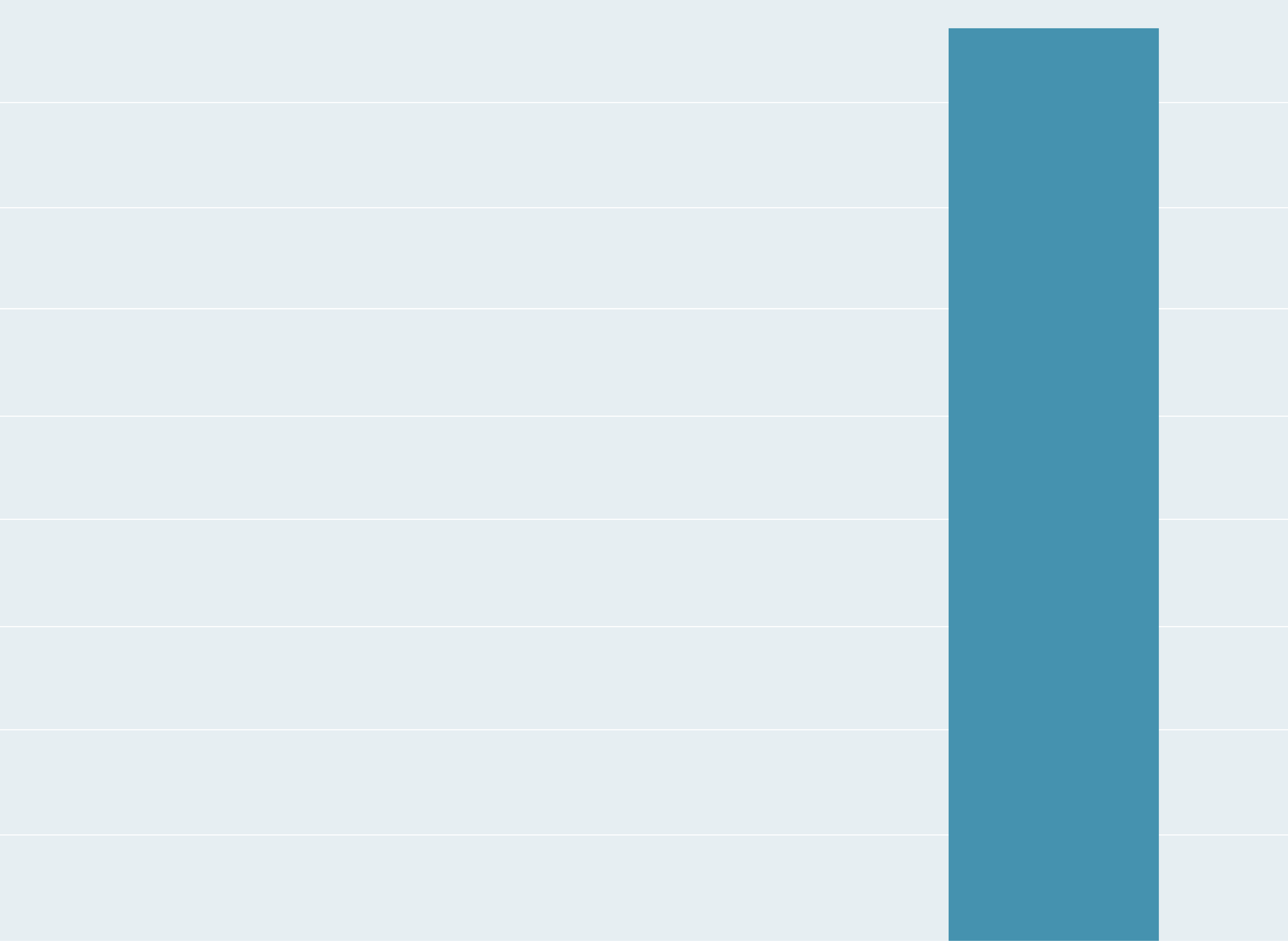}%
    \label{fig:lcg_chemical_k04_b}%
    }\hfil
    \subfigure[OOD (\textit{fork})]{%
    \includegraphics[clip,trim=4mm 0mm 4mm 0, width=0.27\linewidth,height=0.25\linewidth]{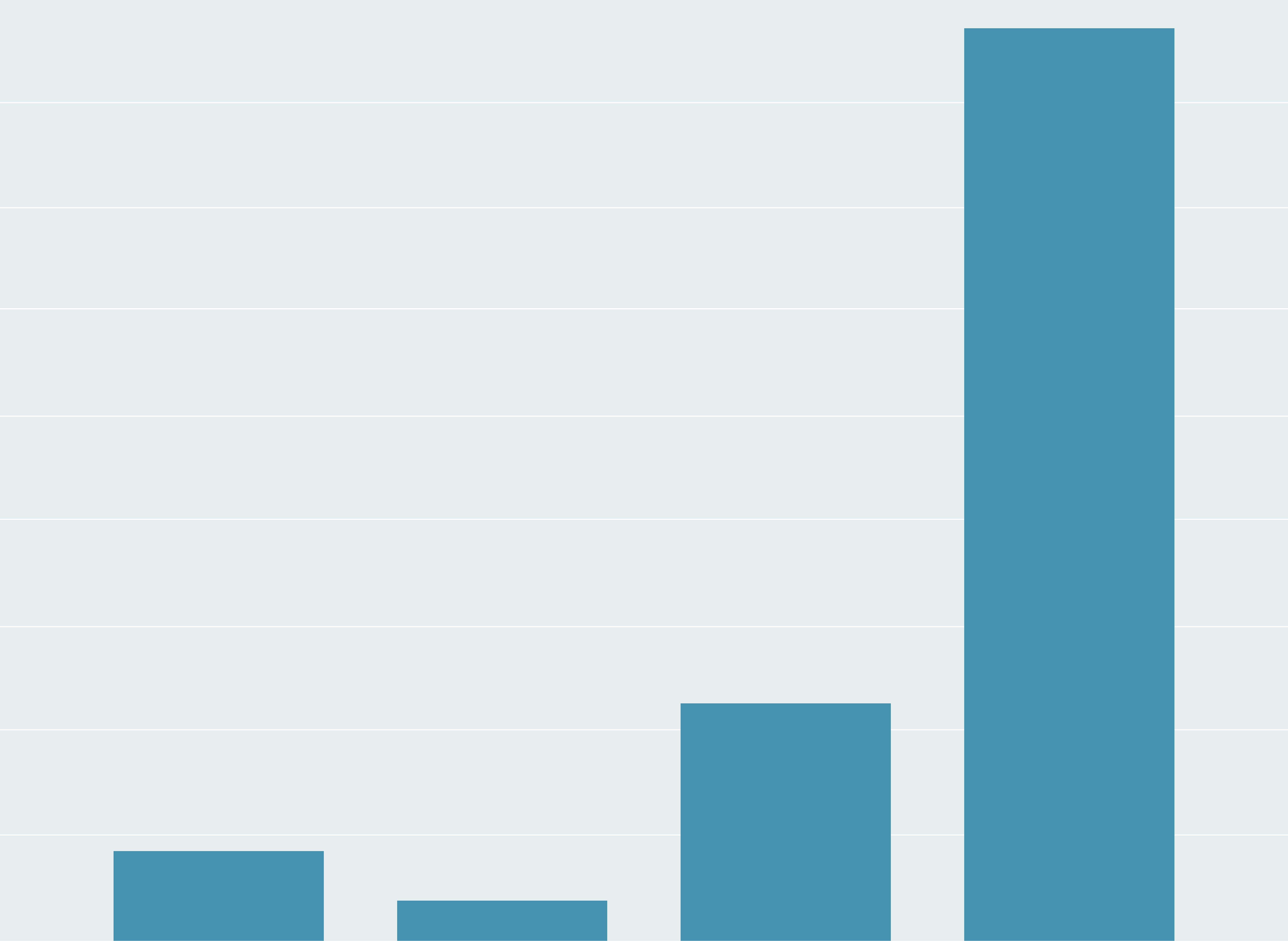}%
    \label{fig:lcg_chemical_k04_c}%
    }
    \\
    \subfigure[True LCG (\textit{fork})]{%
    \includegraphics[clip,trim=130mm 20mm 140mm 4mm, height=0.38\linewidth]{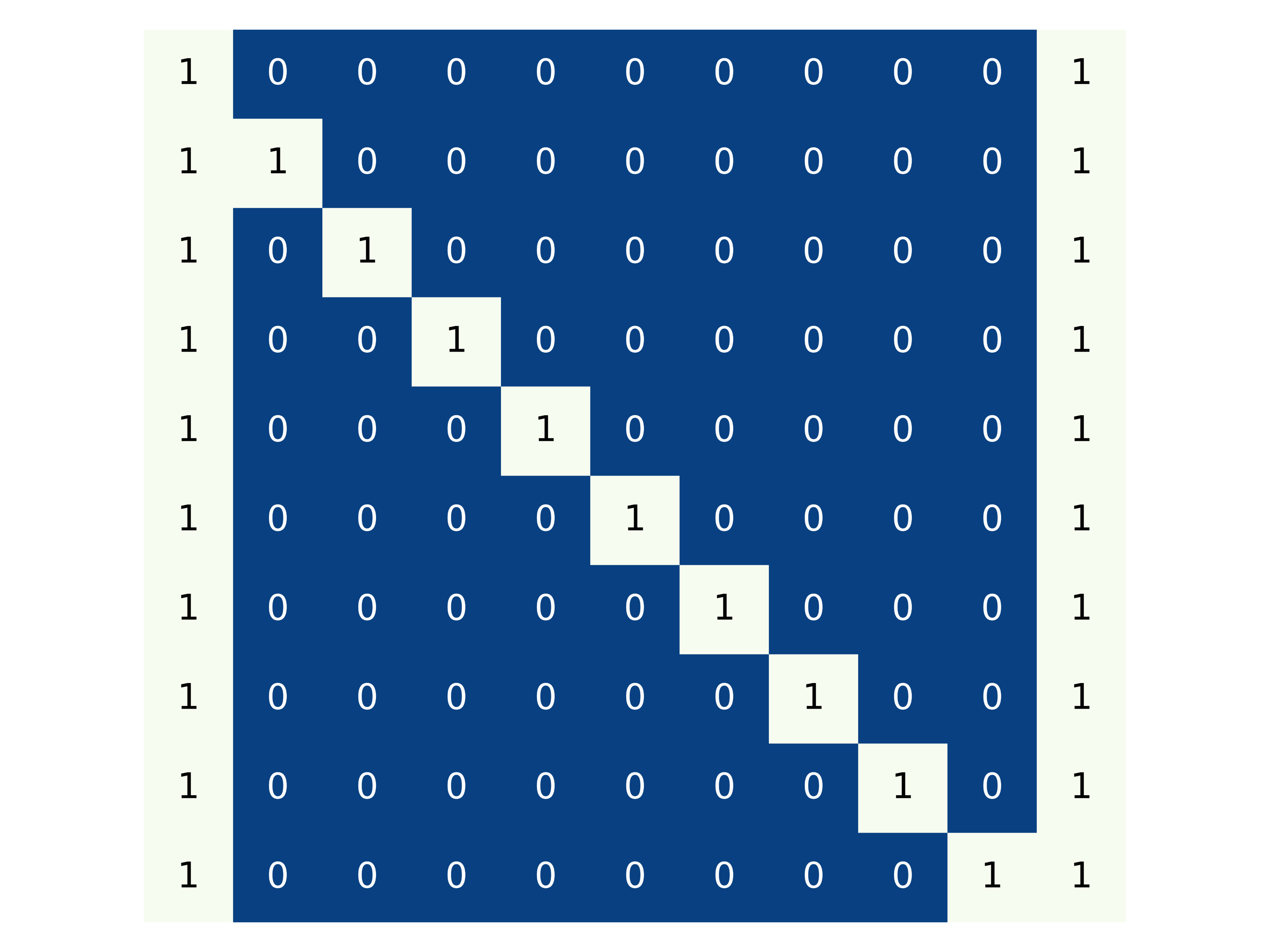}%
    \label{fig:lcg_chemical_k04_d}%
    }\hfil
    \subfigure[Learned LCG (\textit{fork})]{%
    \includegraphics[clip,trim=40mm 20mm 70mm 10mm, height=0.38\linewidth]{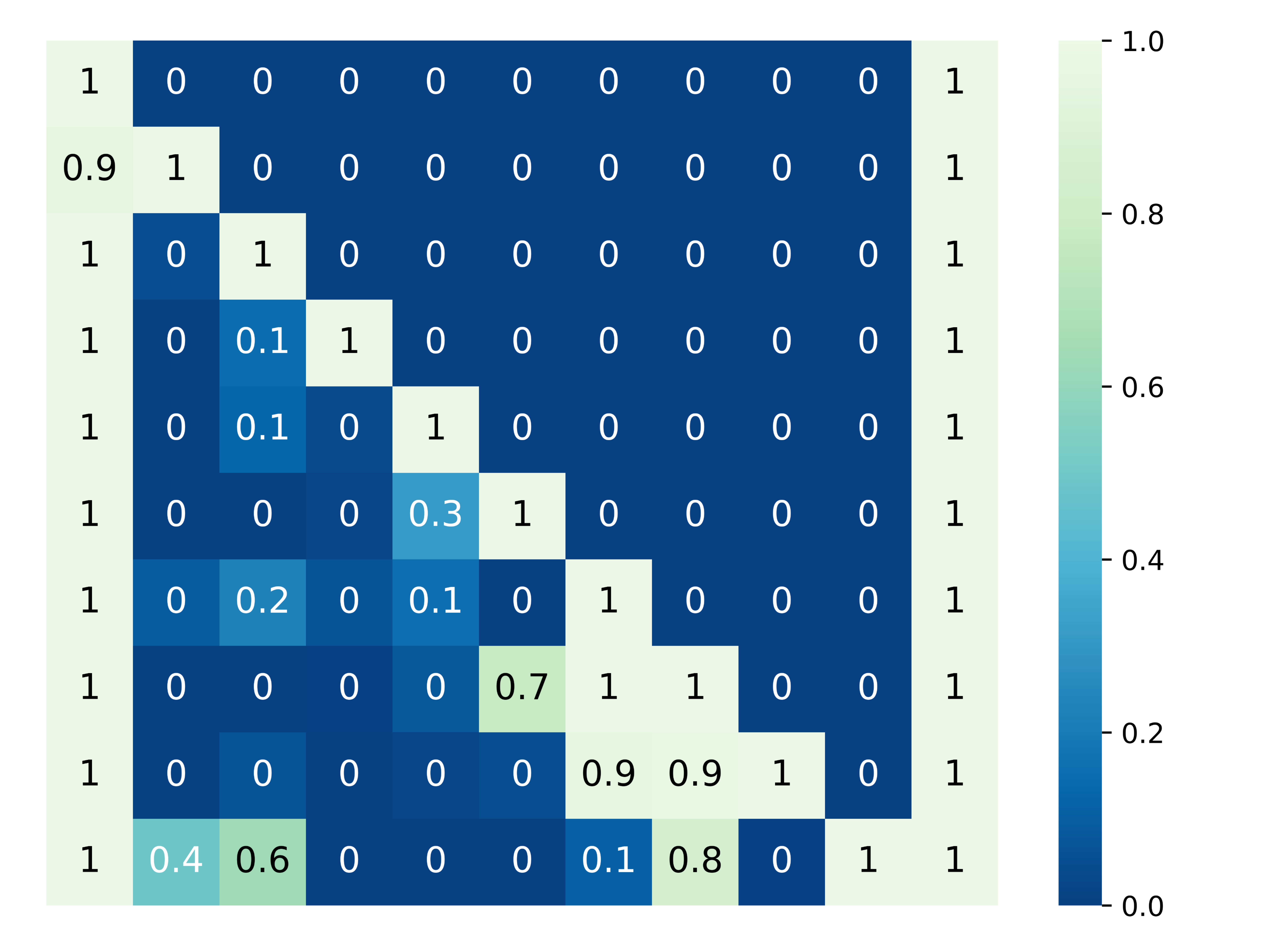}%
    \label{fig:lcg_chemical_k04_e}%
    }
    \vspace{-5pt}
    \caption{
    (\textbf{Top}) Codebook histogram of the sample allocations to each of the codes on (a) all ID states, (b) ID states under \textit{fork}, and (c) OOD states under \textit{fork}.
    (\textbf{Bottom}) 
    (d) True LCG (\textit{fork}).
    (e) Learned LCG corresponding to the most frequently allocated code in (b) and (c).}
    \label{fig:lcg_chemical_k04}
    \vspace{-10pt}
\end{figure}

%% file: figure/lcg_magnetic.tex
\begin{figure*}[t!]
\hfil
\subfigure[FCDL (CG)]{
\includegraphics[clip,trim=130mm 20mm 140mm 4mm, height=0.18\textwidth]{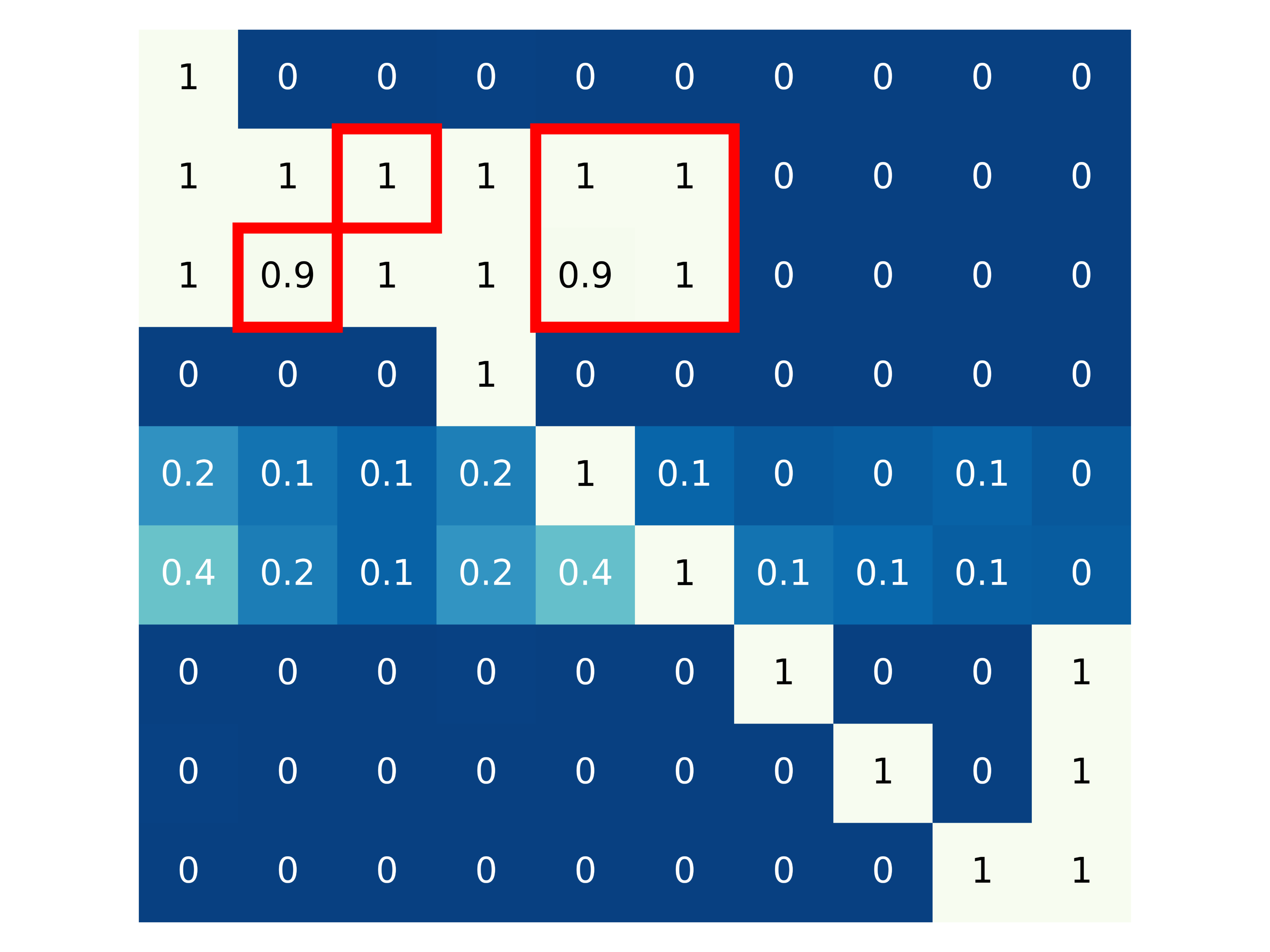}
\label{fig:lcg_magnetic_a}
}\hfil
\subfigure[FCDL (LCG)]{
\includegraphics[clip,trim=130mm 20mm 140mm 4mm, height=0.18\textwidth]{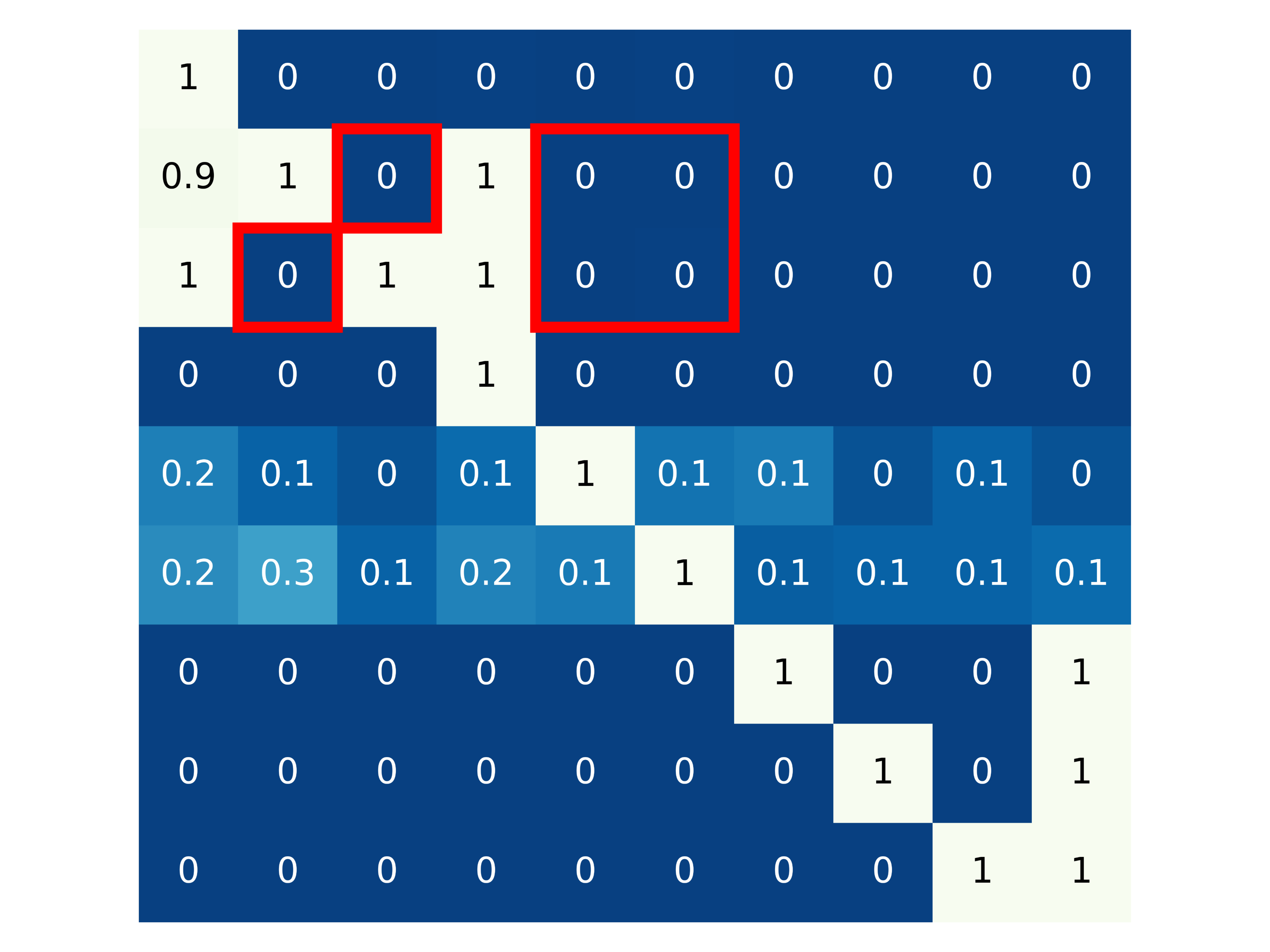}
\label{fig:lcg_magnetic_b}
}\hfil
\subfigure[NCD (LCG, ID)]{
\includegraphics[clip,trim=130mm 20mm 140mm 4mm, height=0.18\textwidth]{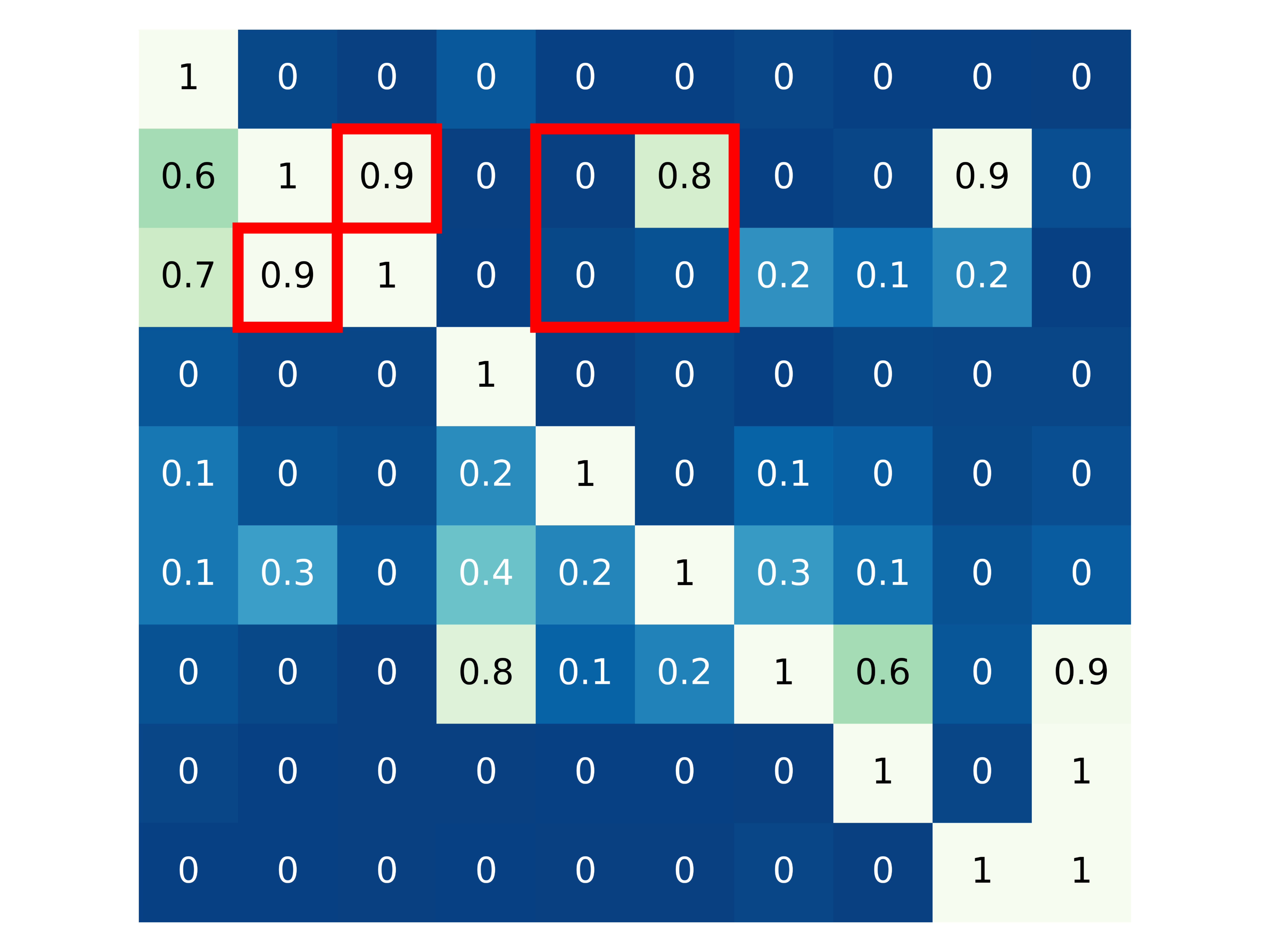}
\label{fig:lcg_magnetic_c}
}\hfil
\subfigure[NCD (LCG, OOD)]{
\includegraphics[clip,trim=40mm 20mm 70mm 10mm, height=0.18\textwidth]{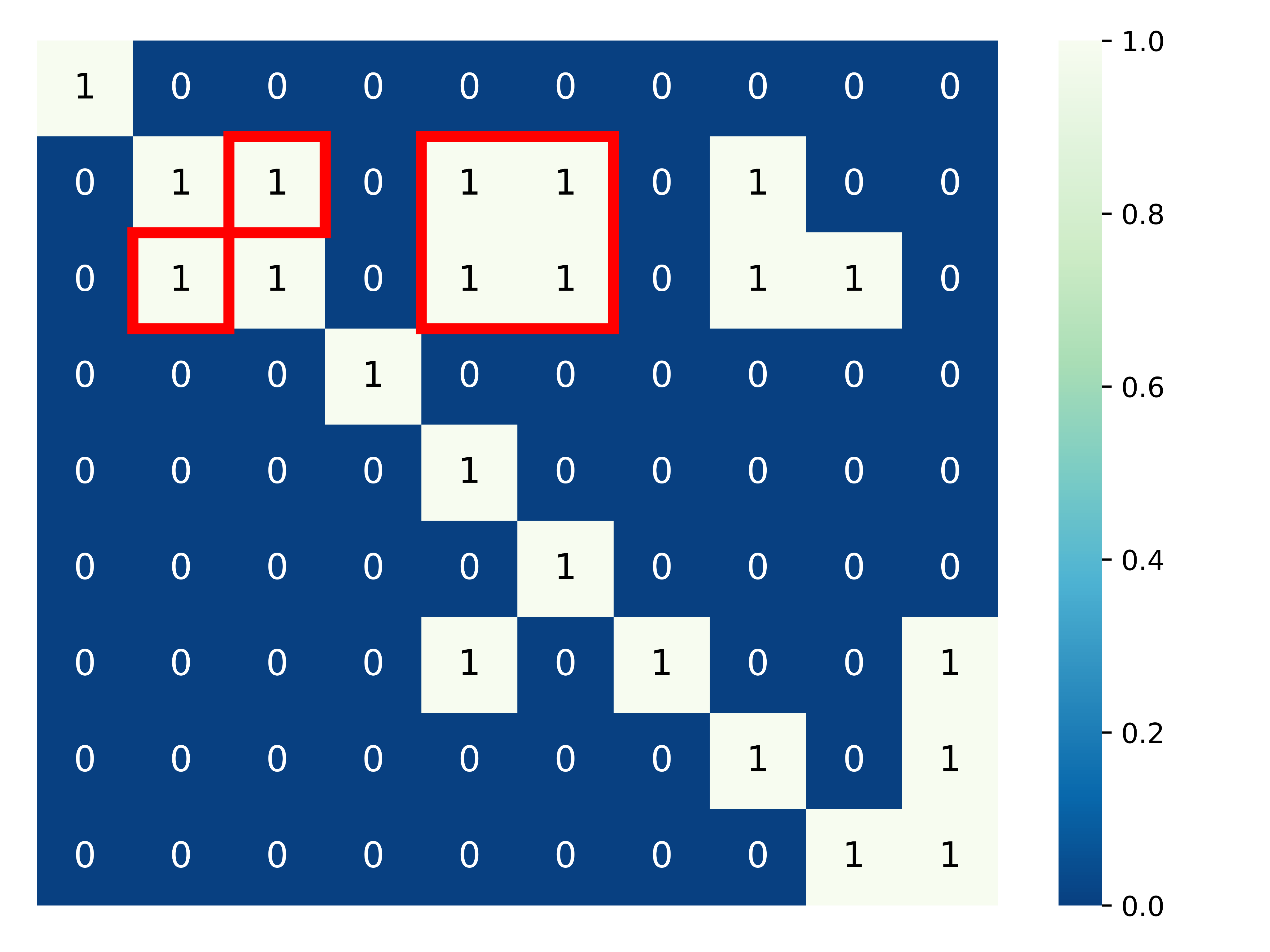}
\label{fig:lcg_magnetic_d}
}
\vspace{-5pt}
\caption{
Red boxes indicate edges included in global CG, but not in LCG under the non-magnetic context. 
(a) CG inferred by our method. 
(b-d) LCG under the non-magnetic context inferred by (b) our method, and NCD on (c) ID and (d) OOD state.
}
\label{fig:lcg_magnetic}
\end{figure*}

%% file: table/ablation_and_shd.tex
\begin{figure*}[t!] \centering
\begin{minipage}[t]{0.63\textwidth} \centering
    \vspace{0pt}
    \includegraphics[width=\linewidth]{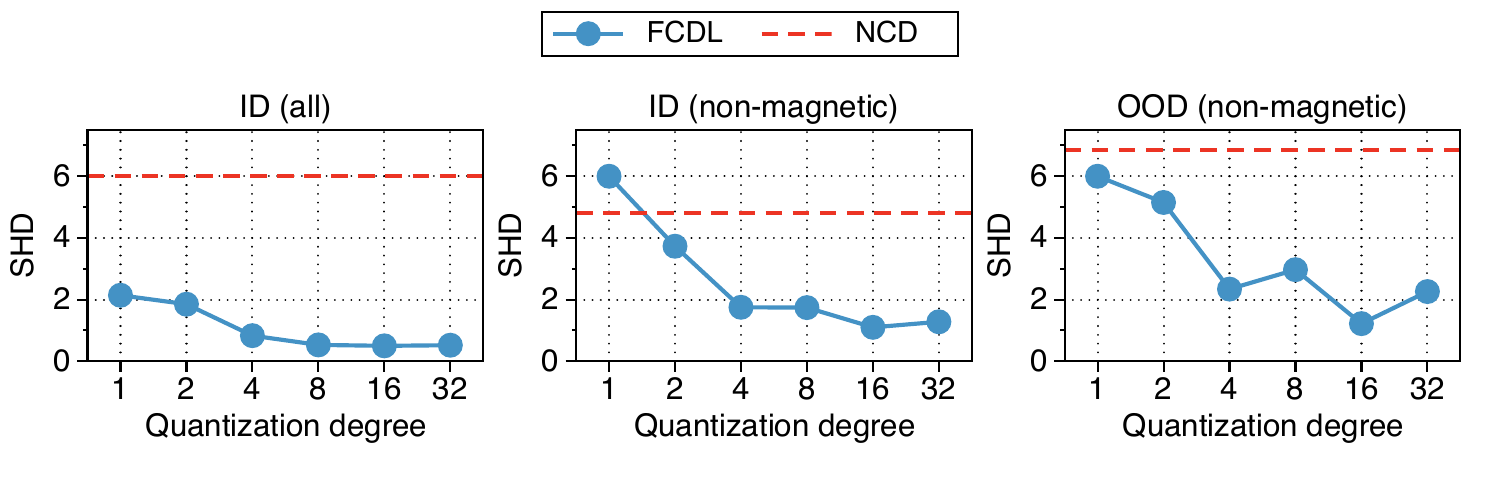}
    \vspace{-25pt}
    \caption{Evaluation of local causal discovery in {Magnetic} environment.}
    \label{fig:magnetic_shd}
\end{minipage}
\hfill
\begin{minipage}[t]{0.33\textwidth} \centering
    \vspace{0pt}
    \vspace{-5pt}
    \captionof{table}{Ablation on the quantization degree.}
    \label{table:ablation_codebook}
    \begin{adjustbox}{width=\linewidth}\setlength{\tabcolsep}{4pt}
    \begin{tabular}{@{}lccc@{}}
    \toprule
    & \multicolumn{3}{c}{\makecell{Chemical (\textit{full-fork})}} \\
    \cmidrule(lr){2-4}
    Methods 
    & \makecell[c]{($n=2$)}
    & \makecell[c]{($n=4$)}
    & \makecell[c]{($n=6$)}
    \\
    \midrule					
    CDL	& {9.37}\stdv{1.33}	& {8.23}\stdv{0.40}	& {9.50}\stdv{1.18}	\\
    NCD	& {10.95}\stdv{1.63}	& {9.11}\stdv{0.63}	& {10.32}\stdv{0.93}	\\
    FCDL ($K=2$)	& {13.44}\stdv{5.41}	& {12.86}\stdv{5.58}	& {12.99}\stdv{5.27}	\\
    FCDL ($K=4$)	& {15.73}\stdv{4.13}	& {16.50}\stdv{3.40}	& {12.40}\stdv{2.81}	\\
    FCDL ($K=8$)	& {14.95}\stdv{1.16}	& {15.03}\stdv{2.61}	& {13.42}\stdv{2.67}	\\
    FCDL ($K=16$)	& {15.27}\stdv{2.53}	& {14.73}\stdv{1.68}	& {13.62}\stdv{2.56}	\\
    FCDL ($K=32$)	& {16.12}\stdv{1.43}	& {14.35}\stdv{1.37}	& {14.79}\stdv{2.13}	\\
    \bottomrule
    \end{tabular}
    \end{adjustbox}
\end{minipage}
\hfil
\end{figure*}

%% file: sections/05-discussions.tex
\paragrapht{High-dimensional observation.} The factorization of the state space is natural in many real-world domains (e.g., healthcare, recommender system, social science, economics) where discovering causal relationships is an important problem. Extending our framework to the image would require extracting causal factors from pixels \citep{scholkopf2021toward}, which is orthogonal to ours and could be combined with.

\paragrapht{Scalability and stability in training.}
Vector quantization (VQ) is a well-established component in generative models where the quantization degree is usually very high (e.g., $K=512, 1024$), yet effectively captures diverse visual features. Its scalability is further showcased in complex large-scale datasets \citep{razavi2019generating}. In this sense, we believe our framework could extend to complex real-world environments. For training stability, techniques have been recently proposed to prevent codebook collapsing, such as codebook reset \citep{williams2020hierarchical} and stochastic quantization \citep{takida2022sq}. We consider that such techniques and tricks could be incorporated into our framework. 

\paragrapht{Conditional independence test (CIT).} A CIT is an effective tool for understanding causal relationships, although often computation-costly. Our method may utilize it to further calibrate the learned LCGs, e.g., applying CIT on each subgroup after the training, which we defer to future work.

\paragrapht{Domain knowledge.}
Our method could leverage prior information on important contexts displaying sparse dependencies, if available. While our method does not rely on such domain knowledge, it would still be useful for discovering fine-grained relationships more efficiently (e.g., \Cref{prop:identifiability-graph-fixed}).

\paragrapht{Implications to real-world scenarios.}
We believe our work has potential implications in many practical applications since context-dependent causal relationships are prevalent in real-world scenarios. For example, in healthcare, a dynamic treatment regime is a task of determining a sequence of decision rules (e.g., treatment type, drug dosage) based on the patient’s health status where it is known that many pathological factors involve fine-grained causal relationships \citep{barash2001context,edwards1985fast}. Our experiments illustrate that existing causal/non-causal RL approaches could suffer from locally spurious correlations and fail to generalize in downstream tasks. We believe our work serves as a stepping stone for further investigation into fine-grained causal reasoning of RL systems and their robustness in real-world deployment.

%% file: sections/06-conclusion.tex
We present a novel approach to dynamics learning that infers fine-grained causal relationships, leading to improved robustness of MBRL. We provide a principled way to examine fine-grained dependencies under certain contexts. As a practical approach, our method learns a discrete latent variable that represents the pairs of a subgroup and local causal graphs (LCGs), allowing joint optimization with the dynamics model. Consequently, our method infers fine-grained causal structures in a more effective and robust manner compared to prior approaches. As one of the first steps towards fine-grained causal reasoning in sequential decision-making systems, we hope our work stimulates future research toward this goal.

%% file: sections/90-appendix_A.tex
\section{Appendix for Preliminary}
\label{appendix:preliminary}
\subsection{Extended Related Work}
\label{appendix:preliminary-related_work}

Recently, incorporating causal reasoning into RL has gained much attention in the community in various aspects. For example, causality has been shown to improve off-policy evaluation \citep{buesing2018woulda,oberst2019counterfactual}, goal-directed tasks \citep{nair2019causal}, credit assignment \citep{mesnard2021counterfactual}, robustness \citep{lyle2021resolving,volodin2020resolving}, policy transfer \citep{killian2022counterfactually}, explainability \citep{madumal2020explainable}, and policy learning with counterfactual data augmentation \citep{lu2020sample,pitis2020counterfactual,pitis2022mocoda}. Causality has also been integrated with bandits \citep{bareinboim2015bandits,lee2018structural,lee2020characterizing}, curriculum learning \citep{li2024causally} or imitation learning \citep{bica2021invariant,de2019causal,zhang2020causal,kumor2021sequential,jamshidi2024causal} to handle the unobserved confounders and learn generalizable policies. Another line of work focused on causal reasoning over the high-dimensional visual observation \citep{lu2018deconfounding,rezende2020causally,feng2022factored,feng2023learning}, e.g., learning sparse and modular dynamics \citep{goyal2021recurrent,goyal2021factorizing,goyal2021neural}, where the representation learning is crucial \citep{zhang2019learning,sontakke2021causal,tomar2021model,scholkopf2021toward,zadaianchuk2021selfsupervised,yoon2023investigation}.

Our work falls into the category of incorporating causality into dynamics learning in RL \citep{mutti2023provably}, where recent works have focused on conditional independences between the variables and their global causal relationships \citep{wang2021task,wang2022causal,ding2022generalizing}. On the contrary, our work incorporates fine-grained causal relationships into dynamics learning, which is underexplored in prior works.

\subsection{Background on Local Independence Relationship}
\label{appendix:preliminary-local_independence}
In this subsection, we provide the background on the local independence relationship. We first describe context-specific independence (CSI) \citep{boutilier2013contextspecific}, which denotes a variable being conditionally independent of others given a particular context, not the full set of parents in the graph. 
\begin{definition}[Context-Specific Independence (CSI) \citep{boutilier2013contextspecific}, reproduced from \citet{hwang2023on}]
\label[definition]{def:CSI}
$Y$ is said to be {\bf contextually independent} of $\rmX_{B}$ given the context $\rmX_{A}=\rvx_{A}$ if $P\left(y \mid \rvx_{A}, \rvx_{B}\right)=P\left(y \mid \rvx_{A}\right)$, holds for all $y\in \mathcal{Y}$ and $\rvx_{B} \in \mathcal{X}_{B}$ whenever $P\left(\rvx_{A}, \rvx_{B}\right)>0$. This will be denoted by $Y \Perp \rmX_{B} \mid \rmX_{A} = \rvx_{A}$.
\end{definition}
CSI has been widely studied especially for discrete variables with low cardinality, e.g., binary variables.
Context-set specific independence (CSSI) generalizes the notion of CSI allowing continuous variables.
\begin{definition}[Context-Set Specific Independence (CSSI) \citep{hwang2023on}]
\label[definition]{def:CSSI}
Let $\*{X}=\{X_1, \cdots, X_d\}$ be a non-empty set of the parents of $Y$ in a causal graph, and $\gE\subseteq \gX$ be an event with a positive probability. $\gE$ is said to be a {\bf context set} which induces {\bf context-set specific independence (CSSI)} of $\*{X}_{A^c}$ from $Y$ if $p\left(y \mid \rvx_{A^c}, \rvx_{A}\right)=p\left(y \mid \rvx'_{A^c}, \rvx_{A}\right)$ holds for every $\left(\rvx_{A^c}, \rvx_{A}\right), \left(\rvx'_{A^c}, \rvx_{A}\right) \in \gE$. This will be denoted by $Y \Perp \*{X}_{A^c} \mid \*{X}_{A}, \gE$.
\end{definition}
Intuitively, it denotes that the conditional distribution $p(y\mid x) = p(y\mid {x}_{A^c}, {x}_{A})$ is the same for different values of $\rvx_{A^c}$, for all $x = \left(\rvx_{A^c}, \rvx_{A}\right) \in \gE$. In other words, only a subset of the parent variables is sufficient for modeling $p(y\mid x)$ on $\gE$.

\subsection{Fine-Grained Causal Relationships in Factored MDP}
\label{appendix:theory_preliminary}

As mentioned in \Cref{sec:preliminary}, we consider factored MDP where the causal graph is directed bipartite and make standard assumptions in the field to properly identify the causal relationships in MBRL \citep{ding2022generalizing,wang2021task,wang2022causal,seitzer2021causal,pitis2020counterfactual,pitis2022mocoda}. 
\begin{assumption}
\label{assumption:basic}
We assume Markov property \citep{pearl2009causality}, faithfulness \citep{peters2017elements}, and causal sufficiency \citep{spirtes2000causation}.
\end{assumption}
Recall that $\rmX = \{S_1, \cdots, S_N, A_1, \cdots, A_M\}$, $\rmY = \{S'_1, \cdots, S'_N\}$, and $Pa(j)$ is parent variables of $S'_j$. Now, we formally define local independence by adapting CSSI to our setting. 
\begin{definition}[Local Independence]
\label{def:local_independence}
Let $\rmT \subseteq Pa(j)$ and $\gE\subseteq \gX$ with $p(\gE)>0$. We say the local independence $S'_j \Perp \rmX \setminus  \rmT \mid \rmT, \gE$ holds on $\gE$ if $p(s'_j\mid \rvx_{T^c}, \rvx_T) = p(s'_j\mid \rvx'_{T^c}, \rvx_T) $ holds for every $\left(\rvx_{T^c}, \rvx_{T}\right), \left(\rvx'_{T^c}, \rvx_{T}\right) \in \gE$.\footnote{$T$ denotes an index set of $\rmT$.}
\end{definition}
It implies that only a subset of the parent variables ($\rvx_T$) is locally relevant on $\gE$, and any other remaining variables ($\rvx_{T^c}$) are locally irrelevant, i.e., $p(s'_j\mid \rvx)$ is a function of $\rvx_T$ on $\gE$. Local independence generalizes conditional independence in the sense that if $S'_j \Perp \rmX \setminus  \rmT \mid \rmT$ holds, then $S'_j \Perp \rmX \setminus  \rmT \mid \rmT, \gE$ holds for any $\gE\subseteq \gX$.
Throughout the paper, we are concerned with the events with the positive probability, i.e., $p(\gE)>0$.
\begin{definition}
${Pa(j; \gE)}$ is a subset of $Pa(j)$ such that $S'_j \Perp \rmX \setminus  {Pa(j; \gE)} \mid {Pa(j; \gE)}, \gE$ holds and $S'_j \not\Perp \rmX \setminus  \rmT \mid \rmT, \gE$ for any $\rmT\subsetneq {Pa(j; \gE)}$.
\end{definition}

In other words, ${Pa(j; \gE)}$ is a minimal subset of $Pa(j)$ in which the local independence on $\gE$ holds. Clearly, ${Pa(j; \gX)} = Pa(j)$, i.e., local independence on $\gX$ is equivalent to the (global) conditional independence.

LCG (\Cref{def:lcg}) describes fine-grained causal relationships specific to $\gE$.  LCG is always a subgraph of the (global) causal graph, i.e., $\gG_\gD \subseteq \gG$, because if a dependency (i.e., edge) does not exist under the whole domain, it cannot exist under any context. Note that $\gG_\gX = \gG$, i.e., local independence and LCG under $\gX$ are equivalent to conditional independence and CG, respectively. 

Analogous to the faithfulness assumption \citep{peters2017elements} that no conditional independences other than ones entailed by CG are present, we introduce a similar assumption for LCG and local independence.
\begin{assumption}[$\gE$-Faithfulness]
\label{assumption:localfaithfulness}
For any $\gE$, no local independences on $\gE$ other than the ones entailed by $\gG_\gE$ are present, i.e., for any $j$, there does not exists any $\rmT$ such that ${Pa(j; \gE)}\setminus \rmT \neq \emptyset $ and $S'_j \Perp \rmX \setminus  \rmT \mid \rmT, \gE$.
\end{assumption}
Regardless of $\mathcal{E}$-faithfulness assumption, LCG always exists because $Pa(j; \mathcal{E})$ always exists. However, such LCG may not be unique without this (see \citet[Example.~2]{hwang2023on} for this example).  \Cref{assumption:localfaithfulness} implies the uniqueness of ${Pa(j; \gE)}$ and $\gG_\gE$, and thus it is required to properly identify fine-grained causal relationships between the variables.

Such fine-grained causal relationships are prevalent in the real world. \textit{Physical law}; To move a static object, a force exceeding frictional resistance must be exerted. Otherwise, the object would not move. \textit{Logic}; Consider $A\vee B\vee C$. When $A$ is true, any changes of $B$ or $C$ no longer affect the outcome. \textit{Biology}; In general, smoking has a causal effect on blood pressure. However, one’s blood pressure becomes independent of smoking if a ratio of alpha and beta lipoproteins is larger than a certain threshold \citep{edwards1985fast}.

%% file: sections/91-appendix_B.tex
\section{Appendix for Method and Theoretical Analysis}
\label{appendix:theory}

\subsection{Fine-Grained Dynamics Modeling}\label{appendix:theory-method}
With the arbitrary decomposition $\set{\gE_z}^K_{z=1}$, true transition dynamics $p(s'\mid s, a)$ can be written as:
\begin{align}
\label{eq:decompose-p}
p(s'\mid s, a) = \sum_z p(s'\mid s, a, z) p(z\mid s, a)
&= \sum_z \prod_j p(s'_j\mid {Pa(j; \gE_z)}, z) \mathbbm{1}_{\{(s, a)\in \gE_z\}},    
\end{align}
where $p(z\mid s, a) = 1$ if $(s, a)\in \gE_z$. This illustrates our approach to dynamics modeling based on fine-grained causal dependencies: $p(s'_j\mid s, a)$ is a function of $Pa(j, \gE_z)$ on $\gE_z$, and our dynamics model employs locally relevant dependencies $Pa(j, \gE_z)$ for predicting $S'_j$. Our dynamics modeling with some graphs $\{\gG_z\}_{z=1}^K$ is: 
\begin{align}\label{eq:decompose-p-hat}
\hat{p}(s'\mid s, a; \{\gG_z, \gE_z\}, \phi) = 
\sum_z \hat{p}(s'\mid s, a; \gG_z, \phi_z) \mathbbm{1}_{\{(s, a)\in \gE_z\}}
&= \sum_z \prod_j \hat{p}_j(s'_j\mid Pa^{\gG_z}(j); \phi^{(j)}_z) \mathbbm{1}_{\{(s, a)\in \gE_z\}},    
\end{align}
where $\phi^{(j)}_z$ takes $Pa^{\gG_z}(j)$ as an input and outputs the parameters of the density function $\hat{p}_j$ and $\phi \coloneq \{\phi^{(j)}_z\}$. We denote $\hat{p}_{\{\gG_z, \gE_z\}, \phi}\coloneq \hat{p}(s'\mid s, a; \{\gG_z, \gE_z\}, \phi)$ and $\hat{p}_{\gG_z, \phi_z}\coloneq \hat{p}(s'\mid s, a; \gG_z, \phi_z)$. In other words, $\hat{p}_{\{\gG_z, \gE_z\}, \phi}(s'\mid s, a) = \hat{p}_{\gG_z, \phi_z}(s'\mid s, a)$  if $(s, a)\in \gE_z$.

Now, we revisit the score function in \Cref{eq:scorebasedmletwo}:
\begin{align}
\gS(\{\gG_z, \gE_z\}_{z=1}^K) 
&\coloneq  \sup_\phi \mathbb{E}_{p(s, a, s')} \left[ \log \hat{p}(s'\mid s, a; \{\gG_z, \gE_z\}, \phi) - \lambda \lvert \gG_z \rvert  \right], \\
&= \sup_\phi \mathbb{E}_{p(s, a)} \mathbb{E}_{p(s'\mid s, a)} \left[ \log \hat{p}(s'\mid s, a; \{\gG_z, \gE_z\}, \phi) - \lambda \lvert \gG_z \rvert  \right], \\
&= \sup_\phi \sum_z \int_{(s, a)\in \gE_z} p(s, a) \left( \mathbb{E}_{p(s'\mid s, a)}  \log \hat{p}(s'\mid s, a; \gG_z, \phi_z) - \lambda \lvert \gG_z \rvert  \right), \\
&= \sup_\phi \sum_z \left[ \int_{(s, a)\in \gE_z} p(s, a) \left( \mathbb{E}_{p(s'\mid s, a)}  \log \hat{p}(s'\mid s, a; \gG_z, \phi_z) \right) - \lambda \cdot p(\gE_z)\cdot \lvert \gG_z \rvert \right],
\end{align}
where $\hat{p}(s'\mid s, a; \gG_z, \phi_z) = \prod_j \hat{p}_j(s'_j\mid Pa^{\gG_z}(j); \phi^{(j)}_z)$.

\subsection{Proof of \Cref{prop:identifiability-graph-fixed}}
\label{appendix:theory-proof-prop-1}

Due to the nature of factored MDP where the causal graph is directed bipartite, each Markov equivalence class (MEC) constrained under temporal precedence contains a \textit{unique} causal graph (i.e., a skeleton determines a unique causal graph since temporal precedence fully orients the edges). Given this background, it is known that the causal graph is \textit{uniquely identifiable} with oracle conditional independence test \citep{ding2022generalizing} or score-based method \citep{brouillard2020differentiable}.

We will now show that LCG is also uniquely identifiable via score maximization.  Our proof techniques are built upon \citet{brouillard2020differentiable}. It is worth noting that they provide the identifiability of (global) CG up to $\gI$-MEC \citep{yang2018characterizing} by utilizing observational and interventional data. In contrast, our analysis is on the identifiability of \textbf{LCGs} by utilizing only observational data. We start by adopting some assumptions from \citet{brouillard2020differentiable}.

\begin{assumption}\label{assumption:capacity}
The ground truth density $p(s'\mid s, a)\in \gH(\{\gG^*_z, \gE_z\})$ for any decomposition $\{\gE_z\}$ with corresponding true LCGs $\{\gG^*_z\}$, where $\gH(\{\gG^*_z, \gE_z\})\coloneq \{p\mid \exists \phi, p =\hat{p}_{\{\gG^*_z, \gE_z\}, \phi} \}$. We assume the density $\hat{p}_{\{\gG^*_z, \gE_z\}, \phi}$ is strictly positive for all $\phi$.
\end{assumption}
\begin{definition}
For a graph $\gG$ and $\gE\subset \gX$, let $\gF_\gE(\gG)$ be a set of conditional densities $f$ such that $f(s'\mid s, a) = \prod_j f_j(s'_j\mid Pa^\gG(j))$ for all $(s, a)\in \gE$ where each $f_j$ is a conditional density.
\end{definition}
\begin{assumption}\label{assumption:finiteentropy}
$\lvert \mathbb{E}_{p(s, a, s')} \log p(s'\mid s, a)\rvert < \infty$.
\end{assumption}
\Cref{assumption:capacity} states that the model parametrized by neural network has sufficient capacity to represent the ground truth density. \Cref{assumption:finiteentropy} is a technical tool for handling the score differences as we will see later.
\begin{lemma}\label{lemma:maxscore}
Let $\gG^*_z$ be a true LCG on $\gE_z$ for all $z$. Then, $\gS(\{\gG^*_z, {\gE}_z\}_{z=1}^K) = \mathbb{E}_{p(s, a, s')} \log p(s'\mid s, a) - \lambda \cdot \mathbb{E}\left[ \lvert \gG^*_z \rvert \right]$.
\end{lemma}
\begin{proof}
First, 
\begin{equation}
0 \leq D_{KL} (p \parallel \hat{p}_{\{\gG^*_z, \gE_z\}, \phi}) = \mathbb{E}_{p(s, a, s')} \log p(s'\mid s, a) - 
\mathbb{E}_{p(s, a, s')} \log \hat{p}(s'\mid s, a; \{\gG^*_z, \gE_z\}, \phi),
\end{equation}
where the equality holds because $\mathbb{E}_{p(s, a, s')} \log p(s'\mid s, a) < \infty$ by \Cref{assumption:finiteentropy}. Therefore, 
\begin{equation}\label{eq:sup_leq}
\sup_\phi \mathbb{E}_{p(s, a, s')} \log \hat{p}(s'\mid s, a; \{\gG^*_z, \gE_z\}, \phi) \leq \mathbb{E}_{p(s, a, s')} \log p(s'\mid s, a).
\end{equation}
On the other hand, by \Cref{assumption:capacity}, there exists $\phi^*$ such that $p=\hat{p}_{\{\gG^*_z, \gE_z\}, \phi^*}$. Hence, 
\begin{equation}\label{eq:sup_geq}
\sup_\phi \mathbb{E}_{p(s, a, s')} \log \hat{p}(s'\mid s, a; \{\gG^*_z, \gE_z\}, \phi) 
\geq \mathbb{E}_{p(s, a, s')} \log \hat{p}(s'\mid s, a; \{\gG^*_z, \gE_z\}, \phi^*) = \mathbb{E}_{p(s, a, s')} \log p(s'\mid s, a).
\end{equation}
By \Cref{eq:sup_leq,eq:sup_geq}, we have $\sup_\phi \mathbb{E}_{p(s, a, s')} \log \hat{p}(s'\mid s, a; \{\gG^*_z, \gE_z\}, \phi) = \mathbb{E}_{p(s, a, s')} \log p(s'\mid s, a)$. Therefore, we have $\gS(\{\gG^*_z, {\gE}_z\}_{z=1}^K) = \mathbb{E}_{p(s, a, s')} \log p(s'\mid s, a) - \lambda \cdot \mathbb{E}\left[ \lvert \gG^*_z \rvert \right]$.
\end{proof}
\begin{corollary}\label{lemma:finitescore}
Let $\gG^*_z$ be a true LCG on $\gE_z$ for all $z$. Then, $\lvert \gS(\{\gG^*_z, {\gE}_z\}_{z=1}^K) \rvert < \infty$.
\end{corollary}
\begin{proof}
By \Cref{lemma:maxscore}, $\gS(\{\gG^*_z, {\gE}_z\}_{z=1}^K) = \mathbb{E}_{p(s, a, s')} \log p(s'\mid s, a) - \lambda \cdot \mathbb{E}\left[ \lvert \gG^*_z \rvert \right]$. Since $\lvert \mathbb{E}_{p(s, a, s')} \log p(s'\mid s, a)\rvert < \infty$ by \Cref{assumption:finiteentropy} and $\lvert \gG^*_z \rvert \leq N(N+M)$, this concludes that $\lvert \gS(\{\gG^*_z, {\gE}_z\}_{z=1}^K) \rvert < \infty$.
\end{proof}
\begin{lemma}\label{lemma:scoredifference}
Let $\gG^*_z$ be a true LCG on $\gE_z$ for all $z$. Then,
\begin{equation}
\gS(\{\gG^*_z, {\gE}_z\}_{z=1}^K) - \gS(\{\gG_z, \gE_z\}_{z=1}^K)
=\inf_\phi D_{KL} (p \parallel \hat{p}_{\{\gG_z, \gE_z\}, \phi}) + \lambda \sum_z p(\gE_z) (\lvert \gG_z \rvert - \lvert \gG^*_z \rvert).
\end{equation}
\end{lemma}
\begin{proof}
First, we can rewrite the score $\gS(\{\gG_z, \gE_z\}_{z=1}^K)$ as:
\begin{align}
\gS(\{\gG_z, \gE_z\}_{z=1}^K) 
& = \sup_\phi \mathbb{E}_{p(s, a, s')} \log \hat{p}(s'\mid s, a; \{\gG_z, \gE_z\}, \phi) - \lambda \cdot \mathbb{E} \left[ \lvert \gG_z \rvert \right]  \\
& = - \inf_\phi - \mathbb{E}_{p(s, a, s')} \log \hat{p}(s'\mid s, a; \{\gG_z, \gE_z\}, \phi) - \lambda \cdot \mathbb{E} \left[ \lvert \gG_z \rvert \right] \\
& = - \inf_\phi D_{KL} (p \parallel \hat{p}_{\{\gG_z, \gE_z\}, \phi}) 
+ \mathbb{E}_{p(s, a, s')} \log p(s'\mid s, a)
- \lambda \cdot \mathbb{E} \left[ \lvert \gG_z \rvert \right]
\label{eq:generalscore}
\end{align}
The last equality holds by \Cref{assumption:finiteentropy}. Subtracting \Cref{eq:generalscore} from \Cref{lemma:maxscore}, we obtain:
\begin{equation*}
\gS(\{\gG^*_z, {\gE}_z\}_{z=1}^K) - \gS(\{\gG_z, \gE_z\}_{z=1}^K)
= \inf_\phi D_{KL} (p \parallel \hat{p}_{\{\gG_z, \gE_z\}, \phi}) + \lambda \sum_z p(\gE_z) (\lvert \gG_z \rvert - \lvert \gG^*_z \rvert).
\end{equation*}
Note that $\lvert \gS(\{\gG^*_z, {\gE}_z\}_{z=1}^K) \rvert < \infty$ by  \Cref{lemma:finitescore}, and thus, this score difference is well defined. 
\end{proof}
\begin{lemma}[Modified from \citet{brouillard2020differentiable}, Lemma 16]
\label{lemma:infpos}
If $p\notin \gF_{\gE_z}(\gG_z)$, then 
\begin{equation}
\inf_{\phi_z} \int {p_z(s, a)} D_{KL} (p(\cdot \mid s, a) \parallel \hat{p}_{\gG_z, \phi_z}(\cdot \mid s, a)) > 0,
\end{equation}
where $p_z(s, a)\coloneq p(s, a\mid z) = p(s, a)/p(\gE_z)$ for all $(s, a)\in \gE_z$, i.e., density function of the distribution $P_{S\times A \mid \gE_z}$.
\end{lemma}
\begin{proof}
First, since $\hat{p}_{\gG_z, \phi_z} \in \gF_{\gE_z}(\gG_z)$ for all $\phi_z$, 
\begin{equation}
\label{eq:infpos-eq1}
\inf_{\phi_z} \int {p_z(s, a)} D_{KL} (p(\cdot \mid s, a) \parallel \hat{p}_{\gG_z, \phi_z}(\cdot \mid s, a))
\geq 
\inf_{f\in \gF_{\gE_z}(\gG_z)} \int {p_z(s, a)} D_{KL} (p(\cdot \mid s, a) \parallel f(\cdot \mid s, a)).
\end{equation}
Now, let $\hat{f}(s'\mid s, a)\coloneq  \prod_j p_z(s'_j \mid Pa^{\gG_z}(j))$ for all $(s, a)\in \gE_z$. Then, for any $f\in \gF_{\gE_z}(\gG_z)$, 
\begin{align}
\label{eq:infpos-eq2}
\int {p_z(s, a)} \int p(s'\mid s, a) \log \frac{\hat{f}(s' \mid s, a)}{f(s' \mid s, a)} 
&= \int p_z(s, a, s') \sum_j \log \frac{p_z(s'_j \mid Pa^{\gG_z}(j))}{f_j(s'_j \mid Pa^{\gG_z}(j))} \notag \\
&= \sum_j  \int p_z(s, a, s') \log \frac{p_z(s'_j \mid Pa^{\gG_z}(j))}{f_j(s'_j \mid Pa^{\gG_z}(j))} \notag \\
&= \sum_j \int p_z(Pa^{\gG_z}(j)) \int p_z(s'_j \mid Pa^{\gG_z}(j)) \log \frac{p_z(s'_j \mid Pa^{\gG_z}(j))}{f_j(s'_j \mid Pa^{\gG_z}(j))} \geq 0.
\end{align}
Therefore, for any $f\in \gF_{\gE_z}(\gG_z)$, 
\begin{align*}
\int {p_z(s, a)} D_{KL} (p(\cdot \mid s, a) \parallel f(\cdot \mid s, a))
&= \int {p_z(s, a)} \int p(s'\mid s, a) \log \frac{p(s' \mid s, a)}{\hat{f}(s' \mid s, a)} \frac{\hat{f}(s' \mid s, a)}{f(s' \mid s, a)} \\
&= \int {p_z(s, a)} D_{KL} (p\parallel \hat{f}) + \int {p_z(s, a)} \int p(s'\mid s, a) \log \frac{\hat{f}(s' \mid s, a)}{f(s' \mid s, a)} \\
&\geq \int {p_z(s, a)} D_{KL} (p\parallel \hat{f}).
\end{align*}
Here, the last inequality holds by \Cref{eq:infpos-eq2}. Therefore, 
\begin{equation}
\label{eq:infpos-eq3}
\inf_{f\in \gF_{\gE_z}(\gG_z)} \int {p_z(s, a)} D_{KL} (p(\cdot \mid s, a) \parallel f(\cdot \mid s, a))
= \int {p_z(s, a)} D_{KL} (p(\cdot \mid s, a)\parallel \hat{f}(\cdot \mid s, a)) > 0.
\end{equation}
Here, the last inequality holds because $\hat{f}\in \gF_{\gE_z}(\gG_z)$ and $p\notin \gF_{\gE_z}(\gG_z)$ and thus $p\neq \hat{f}$. By \Cref{eq:infpos-eq1,eq:infpos-eq3}, the proof is complete.
\end{proof}

\fixedz*
\begin{proof}
To simplify the notation, let $\gG^*_z$ be a true LCG on $\gE_z$ for all $z$, i.e., $\gG^*_z \coloneq  \gG_{\gE_z}$ for brevity. It is enough to show that $\gS(\{\gG^*_z, {\gE}_z\}_{z=1}^K) > \gS(\{\gG_z, \gE_z\}_{z=1}^K)$ if $\gG^*_z \neq \gG_z$ for some $z$. 

Now, by \Cref{lemma:scoredifference},
\begin{align}
&\gS(\{\gG^*_z, {\gE}_z\}_{z=1}^K) - \gS(\{\gG_z, \gE_z\}_{z=1}^K) \\
= &\inf_\phi D_{KL} (p \parallel \hat{p}_{\{\gG_z, \gE_z\}, \phi}) + \lambda \sum_z p(\gE_z) (\lvert \gG_z \rvert - \lvert \gG^*_z \rvert)  \\
= &\inf_\phi \int {p(s, a)} D_{KL} (p(\cdot \mid s, a) \parallel \hat{p}_{\{\gG_z, \gE_z\}, \phi}(\cdot \mid s, a)) + \lambda \sum_z p(\gE_z) (\lvert \gG_z \rvert - \lvert \gG^*_z \rvert) \\
= &\inf_\phi \sum_z \int_{(s, a)\in \gE_z} {p(s, a)} D_{KL} (p(\cdot \mid s, a) \parallel \hat{p}_{\gG_z, \phi_z}(\cdot \mid s, a)) + \lambda \sum_z p(\gE_z) (\lvert \gG_z \rvert - \lvert \gG^*_z \rvert) \\
= &\sum_z \inf_{\phi_z} p(\gE_z) \int {p_z(s, a)} D_{KL} (p(\cdot \mid s, a) \parallel \hat{p}_{\gG_z, \phi_z}(\cdot \mid s, a)) + \lambda \sum_z p(\gE_z) (\lvert \gG_z \rvert - \lvert \gG^*_z \rvert) \\
= &\sum_z p(\gE_z) \inf_{\phi_z} D_{KL} (p_z \parallel \hat{p}_{\gG_z, \phi_z}) + \lambda \sum_z p(\gE_z) (\lvert \gG_z \rvert - \lvert \gG^*_z \rvert) \\
= &\sum_z p(\gE_z) \left[ \inf_{\phi_z} D_{KL} (p_z \parallel \hat{p}_{\gG_z, \phi_z}) + \lambda  (\lvert \gG_z \rvert - \lvert \gG^*_z \rvert) \right]
= \sum_z p(\gE_z) \cdot A_z.
\end{align}
For brevity, we denote $D_{KL} (p_z \parallel \hat{p}_{\gG_z, \phi_z}) \coloneq  \int {p_z(s, a)} D_{KL} (p(\cdot \mid s, a) \parallel \hat{p}_{\gG_z, \phi_z}(\cdot \mid s, a))$ and $A_z \coloneq  \inf_{\phi_z} D_{KL} (p_z \parallel \hat{p}_{\gG_z, \phi_z}) + \lambda  (\lvert \gG_z \rvert - \lvert \gG^*_z \rvert)$. Now, we will show that for all $z\in [K]$, $A_z > 0$ if and only if $\gG^*_z \neq \gG_z$.

\textbf{Case 0:} $\gG^*_z = \gG_z$. Clearly, $A_z = 0$ in this case. 

\textbf{Case 1:} $\gG^*_z \subsetneq \gG_z$. Then, $\lvert \gG_z \rvert > \lvert \gG^*_z \rvert$ and thus $A_z > 0$ since $\lambda  (\lvert \gG_z \rvert - \lvert \gG^*_z \rvert) >0$. 

\textbf{Case 2:} $\gG^*_z \not\subseteq \gG_z$. 
In this case, there exists $(i\rightarrow j)\in \gG^*_z$ such that $(i\rightarrow j) \notin \gG_z$.
Thus, $S'_j\Perp_{\!\!\!\gG_z} X_i \mid \rmX\setminus \{X_i\}$ and $S'_j\not\Perp_{\!\!\!\gG^*_z} X_i \mid \rmX\setminus \{X_i\}$. Therefore, $S'_j\not\Perp_{\!\!\!p} X_i \mid \rmX\setminus \{X_i\}, \gE_z$ by \Cref{assumption:localfaithfulness}. Thus, $p\notin \gF_{\gE_z}(\gG_z)$ and we have $\inf_{\phi_z} D_{KL} (p_z \parallel \hat{p}_{\gG_z, \phi_z})>0$ by \Cref{lemma:infpos}.

Now, we consider two subcases: (i) $\gG_z \in \mathbb{G}^+_z \coloneq  \{\gG' \mid \gG^*_z \not\subseteq \gG', \lvert \gG' \rvert \geq \lvert \gG^*_z \rvert \}$,  and (ii)  $\gG_z \in \mathbb{G}^-_z \coloneq  \{\gG' \mid \gG^*_z \not\subseteq \gG', \lvert \gG' \rvert < \lvert \gG^*_z \rvert \}$. Clearly, if $\gG_z \in \mathbb{G}^+_z$ then $A_z > 0$. Suppose $\gG_z \in \mathbb{G}^-_z$. Then, 
\begin{align}
& \lambda \leq \eta_z \coloneq \frac{1}{N(N+M)+1} \min_{\gG' \in \mathbb{G}^-_z} \inf_{\phi_z} D_{KL} (p_z \parallel \hat{p}_{\gG', \phi_z}) \\
\Longrightarrow \quad & 
\lambda 
\leq \frac{\inf_{\phi_z} D_{KL} (p_z \parallel \hat{p}_{\gG', \phi_z}) }{N(N+M)+1} 
< \frac{\inf_{\phi_z} D_{KL} (p_z \parallel \hat{p}_{\gG', \phi_z}) }{\lvert \gG^*_z \rvert - \lvert \gG' \rvert} 
\quad \text{for } \forall \gG' \in \mathbb{G}^-_z \\
\Longrightarrow \quad & 
\inf_{\phi_z} D_{KL} (p_z \parallel \hat{p}_{\gG', \phi_z}) + \lambda  (\lvert \gG_z \rvert - \lvert \gG^*_z \rvert) > 0
\quad \text{for } \forall \gG' \in \mathbb{G}^-_z.
\end{align}
Here, we use the fact that $\lvert \gG^*_z \rvert - \lvert \gG' \rvert \leq \lvert \gG^*_z \rvert < N(N+M)+1$. Therefore, for $0<\forall \lambda \leq \eta_z$, we have $A_z > 0$ if $\gG^*_z \neq \gG_z$. Here, we note that $\eta_z > 0$ for all $z$, since $\mathbb{G}^-_z$ is finite and $\inf_{\phi_z} D_{KL} (p_z \parallel \hat{p}_{\gG', \phi_z}) > 0$ for any $\gG'\in\mathbb{G}^-_z$ by \Cref{lemma:infpos}.

Consequently, for $0<\lambda \leq \eta(\{\gE_z\}) \coloneq \min_z \eta_z$, we have $\gS(\{\gG^*_z, {\gE}_z\}_{z=1}^K) - \gS(\{\gG_z, \gE_z\}_{z=1}^K) > 0$ if $\gG^*_z \neq \gG_z$ for some $z$. We also note that $\eta(\{\gE_z\}) > 0$ since $\eta_z > 0$ for all $z$. 
\end{proof}

\subsection{Proof of \Cref{prop:identifiability-graph-general}}
\label{appendix:theory-proof-prop-2}

\begin{definition}
Let $\gT \coloneq \{\{\gE_z\}^K_{z=1}\}$, i.e., a set of all decompositions of size $K$. 
\end{definition}

\begin{definition}
Let $\gT_\lambda  \coloneq  \{\{\gE_z\}^K_{z=1} \mid \eta(\{\gE_z\}) \geq \lambda \}$. %
\end{definition}

\begin{remark}
$\gT_\lambda \rightarrow \gT(=\gT_0)$ as $\lambda \rightarrow 0$.
\end{remark}

Recall that \Cref{prop:identifiability-graph-fixed} holds for $0<\lambda \leq \eta(\{\gE_z\})$. Here, $\eta(\{\gE_z\})$ is the value corresponding to the specific decomposition $\{\gE_z\}$. For the arguments henceforth, we consider the arbitrary decomposition and thus introduce the following assumption.

\begin{assumption}
\label{assumption:inf}
$\inf_{\{\gE_z\}\in \gT} \eta(\{\gE_z\}) > 0$.
\end{assumption}

Note that $\eta(\{\gE_z\}) > 0$ for any $\{\gE_z\}$, and thus $\inf_{\{\gE_z\}\in \gT} \eta(\{\gE_z\}) \geq 0$. We now take $0 < \lambda \leq \inf_{\{\gE_z\}\in \gT} \eta(\{\gE_z\})$ with \Cref{assumption:inf}, which allows \Cref{prop:identifiability-graph-fixed} to hold on any arbitrary decomposition. It is worth noting that this assumption is purely technical because for a small fixed $\lambda > 0$, the arguments henceforth hold for all $\{\gE_z\}\in \gT_\lambda$, where $\gT_\lambda \rightarrow \gT$ as $\lambda \rightarrow 0$.

\propoptimality*
\begin{proof}
Let $0 < \lambda \leq \inf_{\{\gE_z\}\in \gT} \eta(\{\gE_z\})$. (i) First, $\{{\gG}^*_{z}, {\gE}^*_{z}\}_{z=1}^K \in \argmax _{\{\gG_z, \gE_z\}} \gS(\{\gG_z, \gE_z\}_{z=1}^K)$ implies that $\{{\gG}^*_{z}\}_{z=1}^K$ also maximizes the score on the fixed $\{{\gE}^*_{z}\}_{z=1}^K$, i.e., $\{{\gG}^*_{z}\}_{z=1}^K \in \argmax _{\{\gG_z\}} \gS(\{\gG_z, {\gE}^*_{z}\}_{z=1}^K)$. Thus, each ${\gG}^*_{z}$ is true LCG on $\gE^*_z$ by \Cref{prop:identifiability-graph-fixed}, i.e., ${\gG}^*_{z} = \gG_{\gE^*_z}$.

(ii) Also, since $\{\gE_z\}_{z=1}^K$ is the arbitrary decomposition, $\gS(\{{\gG}^*_z, {\gE}^*_z\}) \geq \gS(\{\gG_z, \gE_z\})$ holds. Since $\{\gG_z\}$ is the true LCGs on each $\gE_z$, i.e., $\gG_z = \gG_{\gE_z}$, by \Cref{lemma:maxscore}, 
\begin{equation}
\gS(\{\gG_z, {\gE}_z\}_{z=1}^K) = \mathbb{E}_{p(s, a, s')} \log p(s'\mid s, a) - \lambda \sum_z p(\gE_z) \cdot \lvert \gG_z \rvert.
\end{equation} 
Similarly, since $\{{\gG}^*_z\}$ is the true LCGs on each ${\gE}^*_z$,
\begin{equation}
\gS(\{{\gG}^*_z, {\gE}^*_z\}_{z=1}^K) = \mathbb{E}_{p(s, a, s')} \log p(s'\mid s, a) - \lambda \sum_z p({\gE}^*_z) \cdot \lvert {\gG}^*_z \rvert.
\end{equation}
Therefore, $0 \leq \gS(\{{\gG}^*_z, {\gE}^*_z\}) - \gS(\{\gG_z, \gE_z\}) =  \mathbb{E} \left[ \lvert \gG_{z} \rvert \right] - \mathbb{E} \left[ \lvert {\gG}^*_{z} \rvert \right]$ holds, and thus $\mathbb{E} \left[ \lvert {\gG}^*_{z} \rvert \right] \leq \mathbb{E} \left[ \lvert \gG_{z} \rvert \right]$.
\end{proof}

\subsection{Proof of \Cref{prop:identifiability-event-general}}
\label{appendix:theory-proof-prop-3}

We first provide some useful lemma.

\begin{lemma}[\citet{hwang2023on}, Prop.~4]
\label{lemma:hwang}
$S'_j \Perp \rmX \setminus  {Pa(j; \gE)} \mid {Pa(j; \gE)}, \gF$ holds for any $\gF \subseteq \gE$.
\end{lemma}

\begin{restatable}[Monotonicity]{lemma}{propmonotonicity}
\label{prop:monotonicity}
Let $\gF \subseteq \gE$. Then, $\gG_\gF \subseteq \gG_\gE$.
\end{restatable}
\begin{proof}
Since $S'_j \Perp \rmX \setminus  {Pa(j; \gE)} \mid {Pa(j; \gE)}, \gF$ holds by \Cref{lemma:hwang}, ${Pa(j; \gF)} \subseteq {Pa(j; \gE)}$ holds by definition; otherwise, ${Pa(j; \gF)} \setminus {Pa(j; \gE)} \neq \emptyset$ which leads to contradiction.
Therefore, $Pa(j; \gF) \subseteq Pa(j; \gE)$ for all $j$ and thus $\gG_\gF \subseteq \gG_\gE$.
\end{proof}

Now, we provide a proof of \Cref{prop:identifiability-event-general}.

\begin{definition}\label{def:canonical}
The context $\gD\subset\gX$ is canonical if $\gG_\gF = \gG_\gD$ for any $\gF \subset \gD$.
\end{definition}

\identifiabilityeventgeneral*
\begin{proof}
Let $\{\gF_z\}_{z=1}^K$ be the decomposition such that for all $m\in[H]$, $\bigcup_{z\in J_m} {\gF}_z = \gD_m$ for some $J_m \subset [K]$. Note that such decomposition exists since $K\geq H$. Let $\{\gG_z\}_{z=1}^K$ be the true LCGs corresponding to each $\gF_z$, i.e., $\gG_z = \gG_{\gF_z}$. Recall that $\mathbb{E} \left[ \lvert {\gG}^*_{z} \rvert \right] \leq \mathbb{E} \left[ \lvert \gG_{z} \rvert \right]$ holds by \Cref{prop:identifiability-graph-general}, we have 
\begin{align}
0\leq \mathbb{E} \left[ \lvert \gG_{z} \rvert \right] - \mathbb{E} \left[ \lvert {\gG}^*_{z} \rvert \right]
&= \sum_i p(\gF_i) \lvert \gG_{i} \rvert - \sum_j p({\gE}^*_j) \lvert {\gG}^*_{j} \rvert \notag \\
&= \sum_{i, j} p(\gF_i \cap {\gE}^*_j) (\lvert \gG_{i} \rvert - \lvert {\gG}^*_{j} \rvert). \label{eq:identifiability-event-general-1}
\end{align}
Suppose $p(\gF_i \cap {\gE}^*_j) > 0$ for some $i, j$. Let $\gC_{ij} \coloneq  \gF_i \cap {\gE}^*_j$. Since $\gF_i\subset \gD_m$ for some $m$ and $\gD_m$ is canonical, $\gF_i$ is also canonical. Therefore, $\gG_{i} = \gG_{\gC_{ij}}$ since $\gC_{ij} \subset \gF_i$. Since $\gC_{ij} \subset {\gE}^*_j$, we have $\gG_{\gC_{ij}} \subseteq {\gG}^*_{j}$ by \Cref{prop:monotonicity}. Therefore, we have $\gG_{i} \subseteq {\gG}^*_{j}$. Therefore, $\lvert \gG_{i} \rvert - \lvert {\gG}^*_{j} \rvert \leq 0$ for any $i, j$ such that $p(\gF_i \cap {\gE}^*_j) > 0$. Thus, by \Cref{eq:identifiability-event-general-1}, $\lvert \gG_{i} \rvert = \lvert {\gG}^*_{j} \rvert$ if $p(\gF_i \cap {\gE}^*_j) > 0$. Since $\gG_{i} \subseteq {\gG}^*_{j}$ if $p(\gF_i \cap {\gE}^*_j) > 0$, we conclude that 
\begin{equation}
\label{eq:identifiability-event-general-2}
\gG_{i} = {\gG}^*_{j} \quad \text{if} \quad p(\gF_i \cap {\gE}^*_j) > 0.
\end{equation}
Now, for arbitrary ${\gE}^*_j$, suppose there exist $s \neq t$ such that $p(\gD_s \cap {\gE}^*_j) > 0$ and $p(\gD_t \cap {\gE}^*_j) > 0$. Then, there exist some $\gF_i\subset \gD_s$ and $\gF_k\subset \gD_t$ such that $p(\gF_i \cap {\gE}^*_j) > 0$ and $p(\gF_k \cap {\gE}^*_j) > 0$. By \Cref{eq:identifiability-event-general-2}, we have $\gG_i = {\gG^*_j} = \gG_k$. Also, $\gG_{\gD_s} = \gG_i$ and $\gG_{\gD_t} = \gG_k$ since $\gD_s, \gD_t$ are canonical. Therefore, we have $\gG_{\gD_s} = \gG_{\gD_t}$, which contradicts that $\gG_{\gD_m}$ is distinct for all $m$. Therefore, for any ${\gE}^*_j$, there exists a unique $\gD_m$ such that $p(\gD_m \cap \gE^*_j) > 0$, which leads $p(\gE^*_j \setminus \gD_m) = 0$ since $\{\gD_m\}_{m\in[H]}$ is a decomposition of $\gX$. Let $I_m = \{j\in [K] \mid p(\gD_m \cap \gE^*_j) > 0\}$. Here, we have 
\begin{equation}
\label{eq:identifiability-event-general-3}
p\left(\bigcup_{z\in I_m} \gE^*_z\setminus \gD_m \right) = \sum_{z\in I_m} p\big(\gE^*_z\setminus \gD_m \big)=0.
\end{equation}
Also, by the definition of $I_m$ and because $\{\gE^*_z\}_{z\in[K]}$ is a decomposition of $\gX$, we have 
\begin{equation}
\label{eq:identifiability-event-general-4}
p\left(\gD_m \setminus \bigcup_{z\in I_m} \gE^*_z\right)=0.
\end{equation}
Therefore, by \Cref{eq:identifiability-event-general-3,eq:identifiability-event-general-4}, we have $\gD_m = \bigcup_{z\in I_m} \gE^*_z$ almost surely for all $m\in [H]$.
\end{proof}

%% file: sections/92-appendix_C.tex
\section{Appendix for Experiments}
\label{appendix:experiment}

\subsection{Environment Details}
\label{appendix:task_setup}

\input{table/param_env_cem}

\input{figure/dataset_detail_chemical}

\subsubsection{Chemical}
Here, we describe two settings, namely \textit{full-fork} and \textit{full-chain}, modified from \citet{ke2021systematic}. In both settings, there are 10 state variables representing the color of corresponding nodes, with each color represented as a one-hot encoding. The action variable is a 50-dimensional categorical variable that changes the color of a specific node to a new color (e.g., changing the color of the third node to blue). According to the underlying causal graph and pre-defined conditional probability distributions, implemented with randomly initialized neural networks, an action changes the colors of the intervened object's descendants as depicted in \Cref{fig:dataset_detail_chemical}. As shown in \Cref{fig:dataset_chemical}, the (global) causal graph is \textit{full} in both settings, and the LCG is \textit{fork} and \textit{chain}, respectively. For example in \textit{full-fork}, the LCG \textit{fork} is activated according to the particular color of the root node, as shown in \Cref{fig:dataset_detail_chemical}.

In both settings, the task is to match the colors of each node to the given target. The reward function is defined as: 
\begin{equation}
\label{eq:reward_chemical}
r = \frac{1}{\lvert O \rvert}\sum_{i\in O} \mathbbm{1}\left[ s_i = g_i\right],
\end{equation}
where $O$ is a set of the indices of observable nodes, $s_i$ is the current color of the $i$-th node, and $g_i$ is the target color of the $i$-th node in this episode. Success is determined if all colors of observable nodes are the same as the target. During training, all 10 nodes are observable, i.e., $O=\{0, \cdots, 9\}$. In downstream tasks, the root color is set to induce the LCG, and the agent receives noisy observations for a subset of nodes, aiming to match the colors of the rest of the observable nodes. As shown in \Cref{fig:dataset_detail_chemical}, noisy nodes are spurious for predicting the colors of other nodes under the LCG. Thus, the agent capable of reasoning the fine-grained causal relationships would generalize well in downstream tasks. Note that the transition dynamics of the environment is the same in training and downstream tasks. To create noisy observations, we use a noise sampled from $\mathcal{N}(0, \sigma^2)$, similar to \citet{wang2022causal}, where the noise is multiplied to the one-hot encoding representing color during the test. In our experiments, we use $\sigma=100$.

As the root color determines the local causal graph in both settings, the root node is always observable to the agent during the test. The root colors of the initial state and the goal state are the same, inducing the local causal graph. As the root color can be changed by the action during the test, this may pose a challenge in evaluating the agent's reasoning of local causal relationships. This can be addressed by modifying the initial distribution of CEM to exclude the action on the root node and only act on the other nodes during the test. Nevertheless, we observe that restricting the action on the root during the test has little impact on the behavior of any model, and we find that this is because the agent rarely changes the root color as it already matches the goal color in the initial state.

\input{figure/magnetic_gt_gcg_lcg}

\subsubsection{Magnetic}
In this environment, there are two objects on a table, a moving ball and a box, colored either red or black, as shown in \Cref{fig:dataset_magnetic}. The red color indicates that the object is \textit{magnetic}. In other words, when they are both colored red, magnetic force will be applied and the ball will move toward the box. If one of the objects is colored black, the ball would not move since the box has no influence on the ball.

The state consists of the color, $x, y$ position of each object, and $x, y, z$ position of the end-effector of the robot arm, where the color is given as the 2-dimensional one-hot encoding. The action is a 3-dimensional vector that moves the robot arm. The causal graph of the Magnetic environment is shown in \Cref{fig:magnetic_gt_gcg}. LCGs under magnetic and non-magnetic context are shown in \Cref{fig:magnetic_gt_lcg1,fig:magnetic_gt_lcg2}, respectively. The table in our setup has a width of 0.9 and a length of 0.6, with the y-axis defined by the width and the x-axis defined by the length. For each episode, the initial positions of a moving ball and a box are randomly sampled within the range of the table. 

The task is to move the robot arm to reach the moving ball. Thus, accurately predicting the trajectory of the ball is crucial. The reward function is defined as:
\begin{equation}
\label{eq:reward_magnetic}
r = 1 - \operatorname{tanh}(5\cdot \|eef - g\|_1),
\end{equation}
where the $eef\in\mathbb{R}^3$ is the current position of the end-effector, $g= (b_x, b_y, 0.8) \in \mathbb{R}^3$, and $(b_x, b_y)$ is the current position of the moving ball. Success is determined if the distance is smaller than 0.05. During the test, the color of one of the objects is black and the box is located at the position unseen during the training. Specifically, the box position is sampled from $\mathcal{N}(0, \sigma^2)$ during the test.  Note that the box can be located outside of the table, which never happens during the training. In our experiments, we use $\sigma=100$.

\subsection{Experimental Details}

To assess the performance of different dynamics models of the baselines and our method, we use a model predictive control (MPC) \citep{camacho2013model} which selects the actions based on the prediction of the learned dynamics model, following prior works \citep{ding2022generalizing,wang2022causal}. Specifically, we use a cross-entropy method (CEM) \citep{rubinstein2004cross}, which iteratively generates and refines action sequences through a process of sampling from a probability distribution that is updated based on the performance of these sampled sequences, with a known reward function. We use a random policy for the initial data collection. Environmental configurations and CEM parameters are shown in \Cref{table:param_env,table:param_cem}, respectively. Most of the experiments were processed using a single NVIDIA RTX 3090. For \Cref{fig:magnetic_shd}, we use structural hamming distance (SHD) for evaluation, which is a metric used to quantify the dissimilarity between two graphs based on the number of edge additions or deletions needed to make the graphs identical \citep{acid2003searching,ramsey2006adjacency}.

\subsection{Implementation of Baselines}
\label{appendix:implementation_details_baselines}

For all methods, the dynamics model outputs the parameters of categorical distribution for discrete variables, and the mean and standard deviation of normal distribution for continuous variables. All methods have a similar number of model parameters for a fair comparison. Detailed parameters of each model are shown in \Cref{table:param_model}.

\paragrapht{MLP and Modular.} MLP models the transition dynamics as $p(s'\mid s, a)$. Modular has a separate network for each state variable, i.e., $\prod_j p(s'_j \mid s, a)$, where each network is implemented as an MLP.

\paragrapht{GNN, NPS, and CDL.} We employ publicly available source codes.\footnote{\url{https://github.com/wangzizhao/CausalDynamicsLearning}} For NPS \citep{goyal2021neural}, we search the number of rules $N \in \{4, 15, 20\}$. CDL \citep{wang2022causal} infers the causal structure by estimating conditional mutual information (CMI) and models the dynamics as $\prod_j p(s'_j \mid Pa(j))$. For CDL, we search the initial CMI threshold $\epsilon \in \{0.001, 0.002, 0.005, 0.01, 0.02\}$ and exponential moving average (EMA) coefficient $\tau \in \{0.9, 0.95, 0.99, 0.999\}$. As CDL is a two-stage method, we only report their final performance.

\paragrapht{GRADER.}
We implement GRADER \citep{ding2022generalizing} based on the code provided by the authors.\footnote{\url{https://github.com/GilgameshD/GRADER}} GRADER relies on the conditional independence test (CIT) to discover the causal structure. In Chemical, we ran the CIT for every 10 episodes, following their default setting. We only report its performance in Chemical due to the poor scalability of the conditional independence test in Magnetic environment, which took about 30 minutes for each test.

\paragrapht{Oracle and NCD.}
For a fair comparison, we employ the same architecture for the dynamic models of Oracle, NCD, and our method, as their main difference lies in the inference of local causal graphs (LCG). As illustrated in \Cref{fig:comparison_ncd_ours}, the key difference is that NCD \citep{hwang2023on} performs direct inference of the LCG from each individual sample (referred to as \textit{sample-specific} inference), while our method decomposes the data domain and infers the LCGs for each subgroup through quantization. We provide an implementation details of our method in the next subsection.

\input{figure/comparison_ncd_ours}

\subsection{Implementation of FCDL}
\label{appendix:implementation_details_ours}

For our method, we use MLPs for the implementation of $g_\texttt{enc}, g_\texttt{dec}$, and $\hat{p}$, with configurations provided in \Cref{table:param_model}. The quantization encoder $g_\texttt{enc}$ of our method or the auxiliary network of NCD shares the initial feature extraction layer with the dynamics model $\hat{p}$ as we found that it yields better performance compared to full decoupling of them.

\subsubsection{Dynamics Model} 
Recall our dynamics modeling in \Cref{eq:decompose-p-hat} that $\hat{p}(s'_j\mid Pa^{\gG_z}(j); \phi^{(j)}_z)$  if $(s, a)\in \gE_z$, which corresponds to $p(s'_j\mid s, a)=p(s'_j\mid Pa(j; \mathcal{E}_z), z)$ in \Cref{eq:decompose}. Here, each $\phi^{(j)}_z$ is a neural network that takes $Pa^{\gG_z}(j)$ as an input and predicts $s'_j$ under $\gE_z$. In general, this separate network for each subgroup would allow it to effectively adapt to environments with complex dynamics and learn transition functions separately for each subgroup. However, this requires a total of $K\times N$ separate networks, which could incur a computational burden. Instead, we employ an efficient parameter-sharing mechanism to simplify the model implementation: we let the dynamics model consist of separate networks for each state variable, i.e., $\phi = \{\phi^{(j)}\}$ and each $\phi^{(j)}$ takes $(Pa^{\gG_z}(j), z)$ as an input, instead of using separate networks $\phi^{(j)}_z$ for each $\gE_z$, which is analogous to $p(s'_j\mid Pa(j; \mathcal{E}_z), z)$. This requires a total of $N$ separate networks, one for each state variable. There are different implementation design choices for $z$ in $(Pa^{\gG_z}(j), z)$. We consider two cases: (i) concatenation of $Pa^{\gG_z}(j)$ and $e_z$ (i.e., code), and (ii) concatenation of $Pa^{\gG_z}(j)$ and one-hot encoding of $z$ (dimension of $K$). We opt for a simpler choice of the latter. This allows us to model (possibly) different transition functions for each subgroup with a single dynamics model for each state variable. Note that if the subgroups having the same LCG share the same transition function, such labeling of $z$ could be further omitted.

For the implementation of taking $Pa^{\gG_z}(j)$ as input for $\hat{p}(s'_j\mid Pa^{\gG_z}(j); \phi^{(j)}_z)$, we simply mask out the features of unused variables, but other design choices such as Gated Recurrent Unit \citep{chung2014empirical,ding2022generalizing} are also possible. As architectural design is not the primary focus of this work, we leave the exploration of different architectures to future work. Note that all baselines except MLP (e.g., GNN and causal dynamics models) use separate networks for each state variable, and we made sure that all methods have a similar number of model parameters for a fair comparison.

\subsubsection{Backpropagation} 
We now describe how each component of our method is updated by the training objective in \Cref{eq:total_loss}. First, $\mathcal{L}_{\texttt{pred}}$ updates the encoder $g_{\texttt{enc}}(s,a)$, decoder $g_{\texttt{dec}}(e)$, and the dynamics model $\hat{p}$. Recall that $A\sim g_{\texttt{dec}}(e)$, backpropagation from $A$ in $\mathcal{L}_{\texttt{pred}}$ updates the quantization decoder $g_{\texttt{dec}}$ through $e$. During the backward path in \Cref{eq:quantize}, gradients are copied from $e$ (= input of $g_{\texttt{dec}}$) to $h$ (= output of $g_{\texttt{enc}}$), following VQ-VAE \citep{van2017neural}. By doing so, $\mathcal{L}_{\texttt{pred}}$ also updates the quantization encoder $g_{\texttt{enc}}$ and $h$. Second, $\mathcal{L}_{\texttt{quant}}$ updates $g_{\texttt{enc}}$ and the codebook $C$. We note that $\gL_\texttt{pred}$ also affects the learning of the codebook $C$ since $h$ is updated with $\gL_\texttt{pred}$. The rationale behind this trick of VQ-VAE is that the gradient $\nabla_e \mathcal{L}_{\texttt{pred}}$ could guide the encoder $g_{\texttt{enc}}$ to change its output $h=g_{\texttt{enc}}(s, a)$ to lower the prediction loss $\mathcal{L}_{\texttt{pred}}$, altering the quantization (i.e., assignment of the cluster) in the next forward pass. A larger prediction loss (which implies that this sample $(s, a)$ is assigned to the wrong cluster) induces a bigger change on $h$, and consequently, it would be more likely to cause a re-assignment of the cluster.

\subsubsection{Hyperparameters}
For all experiments, we fix the codebook size $K=16$, regularization coefficient  $\lambda=0.001$, and commitment coefficient $\beta=0.25$, as we found that the performance did not vary much for any $K>2$, $\lambda \in \{10^{-4}, 10^{-3}, 10^{-2}\}$ and $\beta \in \{0.1, 0.25\}$.

\subsection{Additional Experimental Results}
\label{appendix:additional_experiments}

\input{figure/lcg_chemical_k16}

\input{figure/lcg_chemical_analysis_k04_s1}
\input{figure/lcg_chemical_analysis_k04_s4}

\subsubsection{Detailed analysis of learned LCGs}
\label{appendix:detaild_analysis_lcg}
LCGs learned by our method with a quantization degree of 4 in Chemical are shown in \Cref{fig:lcg_chemical_analysis_k04_s1,fig:lcg_chemical_analysis_k04_s4}.
Among the 4 codes, one (\Cref{fig:lcg_chemical_analysis_k04_s1_b}) or two (\Cref{fig:lcg_chemical_analysis_k04_s4_b}) represent the local causal structure \textit{fork}.
Our method successfully infers the proper code for most of the OOD samples (\Cref{fig:lcg_chemical_analysis_k04_s1_c,fig:lcg_chemical_analysis_k04_s4_c}).
Two sample runs of our method with a quantization degree of 4 in Magnetic are shown in \Cref{fig:lcg_magnetic_analysis_k04_s1,fig:lcg_magnetic_analysis_k04_s4}.
Our method successfully learns LCGs correspond to a non-magnetic context (\Cref{fig:lcg_magnetic_analysis_k04_s1_d,fig:lcg_magnetic_analysis_k04_s1_g,fig:lcg_magnetic_analysis_k04_s4_d,fig:lcg_magnetic_analysis_k04_s4_f}) and magnetic context (\Cref{fig:lcg_magnetic_analysis_k04_s1_e,fig:lcg_magnetic_analysis_k04_s1_f,fig:lcg_magnetic_analysis_k04_s4_e,fig:lcg_magnetic_analysis_k04_s4_g}).

We also observe that our method discovers more fine-grained relationships. Recall that the non-magnetic context is determined when one of the objects is black, the box would have no influence on the ball regardless of the color of the box when the ball is black, and vice versa. As shown in \Cref{fig:lcg_magnetic_analysis_k16_s4}, our method discovers the context where the ball is black (\Cref{fig:lcg_magnetic_analysis_k16_s4_b}), and the context where the box is black (\Cref{fig:lcg_magnetic_analysis_k16_s4_a}).

We observe that the training of latent codebook with vector quantization is often unstable when $K=2$. We demonstrate the success (\Cref{fig:lcg_chemical_analysis_k02_s1}) and failure (\Cref{fig:lcg_chemical_analysis_k02_s2}) cases of our method with a quantization degree of 2. In a failure case, we observe that the embeddings frequently fluctuate between the two codes, resulting in both codes corresponding to the global causal graph and failing to capture the LCG, as shown in \Cref{fig:lcg_chemical_analysis_k02_s2}.

\input{figure/lcg_magnetic_analysis_k04_s1}
\input{figure/lcg_magnetic_analysis_k04_s4}
\input{figure/lcg_magnetic_analysis_k16_s4}

\input{figure/lcg_chemical_analysis_k02_s1}
\input{figure/lcg_chemical_analysis_k02_s2}

\subsubsection{Learning curves on all downstream tasks}
\Cref{fig:learning_curve_train} shows the learning curves on training in all environments.
\Cref{fig:learning_curve_fork,fig:learning_curve_chain,fig:learning_curve} shows the learning curves on all downstream tasks.\footnote{As CDL is a two-stage method that requires searching the best threshold after the first stage training, we only report their final performance.} 

\section{Additional Discussions}

\subsection{Difference from Sample-based Inference}
Sample-based inference methods, e.g., NCD \citep{hwang2023on} for LCG or ACD \citep{lowe2022amortized} for CG, can be seen as learning causal graphs with gated edges. They learn a function that maps each sample to the adjacency matrix where each entry is the binary variable indicating whether the corresponding edge is on or off under the current state. The critical difference from ours is that LCGs learned from sample-based inference methods are \textit{unbounded} and \textit{blackbox}.

Specifically, it is hard to understand which local structures and contexts are identified since they can only be examined by observing the inference outcome from all samples (i.e., \textit{blackbox}). Also, there is no (practical or theoretical) guarantee that it outputs the same graph from the states within the same context, since the output of the function is \textit{unbounded}. In contrast, our method learns a finite set of LCGs where the contexts are explicitly identified by latent clustering. In other words, the outcome is bounded (infers one of the $K$ graphs) and the contexts are more interpretable.

For the robustness of the model and principled understanding of the fine-grained structures, the practical or theoretical guarantee and interpretability are crucial, and we demonstrate the improved robustness of our method compared to prior sample-based inference methods. On the other hand, sample-based inference or local edge switch methods have strength in their simple design and efficiency, and it is known that the signals from such local edge switch enhance exploration in RL \citep{seitzer2021causal,wang2023elden}. For the practitioners, the choice would depend on their purpose, e.g., whether their primary interest is on the robustness and principled understanding of the fine-grained structures.

\subsection{Limitations and Future Works}
Insufficient or biased data may lead to inaccurate learning of causal relationships, including both CG and LCG. Our work explored the potential of utilizing LCGs to deal with (locally) spurious correlations arising from insufficient or biased data in the context of MBRL. While we assumed causal sufficiency, unobserved variables may also influence the causal relationships. These assumptions are commonly adopted in the field, yet we consider that relaxing these assumptions would be a promising future direction. Another promising future direction is to explore an inherent structure to the quantization that can efficiently handle a large number of contexts. 

\input{table/param_model}

\clearpage
\newpage
\input{figure/curve_train}
\input{figure/curve_chemical_fork}
\input{figure/curve_chemical_chain}

%% file: table/param_env_cem.tex
\begin{table}[ht]
    \hfil
    \begin{minipage}[t]{0.4\textwidth}
        \caption{Environment configurations.}
        \label{table:param_env}
        \centering
        \begin{adjustbox}{width=\textwidth}
        \begin{tabular}{@{}lccc@{}}
        \toprule
        & \multicolumn{2}{c}{\makecell{Chemical}} 
        & \multirow{2}{*}{\makecell{Magnetic}}
        \\
        \cmidrule(lr){2-3}
        Parameters
        & \makecell[c]{\textit{full-fork}}
        & \makecell[c]{\textit{full-chain}}
        &
        \\
        \midrule				
        Training step	& $1.5 \times 10^5$	& $1.5 \times 10^5$	& $2 \times 10^5$	\\
        Optimizer	& Adam	& Adam	& Adam	\\
        Learning rate	& 1e-4	& 1e-4	& 1e-4	\\
        Batch size	& 256	& 256	& 256	\\
        Initial step	& 1000	& 1000	& 2000	\\
        Max episode length	& 25	& 25	& 25	\\
        Action type	& Discrete	& Discrete	& Continuous	\\
        \bottomrule
        \end{tabular}
        \end{adjustbox}
    \end{minipage}
    \hfil
    \begin{minipage}[t]{0.4\textwidth}
        \caption{CEM parameters.}
        \label{table:param_cem}
        \centering
        \begin{adjustbox}{width=\textwidth}
        \begin{tabular}{@{}lccc@{}}
        \toprule
        & \multicolumn{2}{c}{\makecell{Chemical}} 
        & \multirow{2}{*}{\makecell{Magnetic}}
        \\
        \cmidrule(lr){2-3}
        CEM parameters
        & \makecell[c]{\textit{full-fork}}
        & \makecell[c]{\textit{full-chain}}
        &
        \\
        \midrule				
        Planning length	& 3	& 3	& 1	\\
        Number of candidates	& 64	& 64	& 64	\\
        Number of top candidates	& 32	& 32	& 32	\\
        Number of iterations	& 5	& 5	& 5	\\
        Exploration noise	& N/A 	& N/A 	& 1e-4	\\
        Exploration probability	& 0.05	& 0.05	& N/A 	\\
        \bottomrule
        \end{tabular}
        \end{adjustbox}
    \end{minipage}
    \hfil  
\end{table}

%% file: figure/dataset_detail_chemical.tex
\begin{figure*}[t]
\centering
\includegraphics[clip,trim=5mm 5mm 5mm 5mm,height=0.3\textwidth]{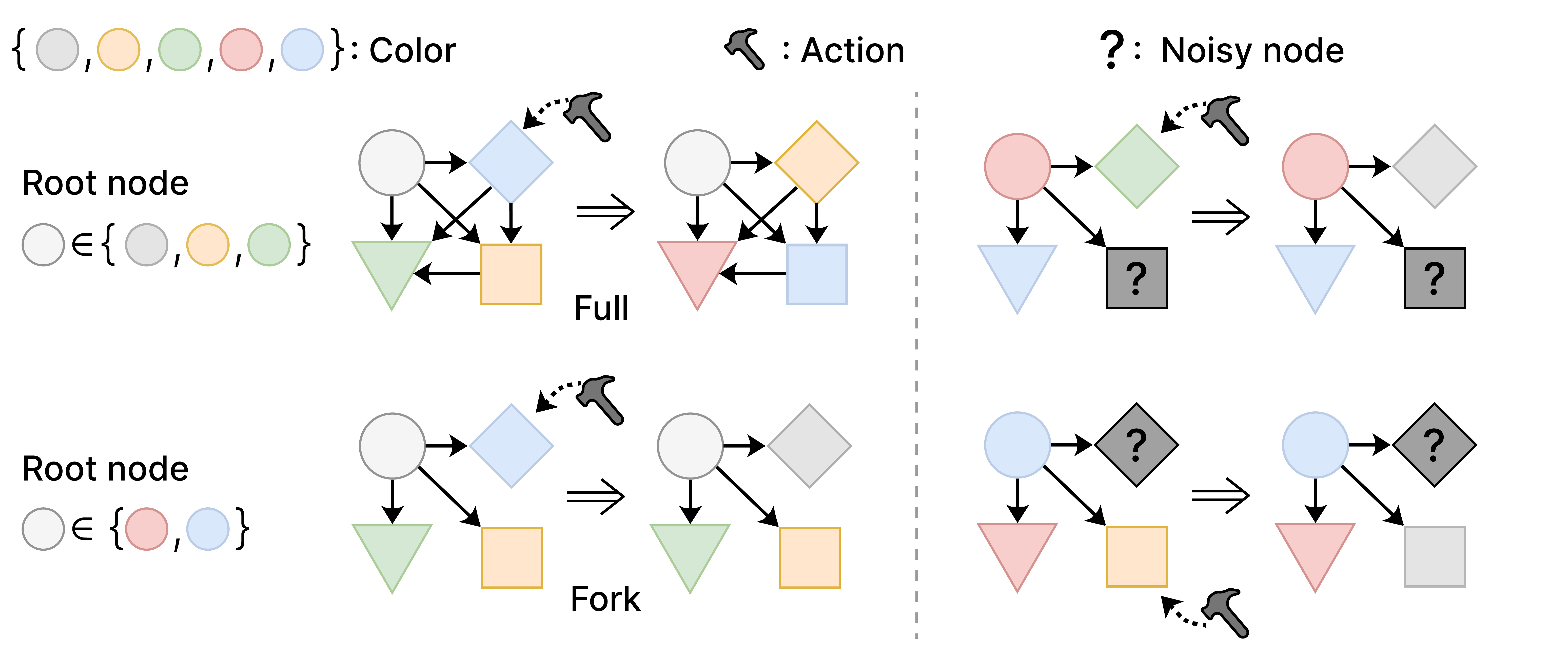}
\label{fig:dataset_detail_chemical_fullfork}
\caption{Illustration of \textsc{Chemical} (\textit{full-fork}) environment with 4 nodes.
(\textbf{Left}) the color of the root node determines the activation of local causal graph \textit{fork}.
(\textbf{Right}) the noisy nodes are redundant for predicting the colors of other nodes under the local causal graph.}
\label{fig:dataset_detail_chemical}
\vspace{-1mm}
\end{figure*}

%% file: figure/magnetic_gt_gcg_lcg.tex
\begin{figure*}[t!]
\subfigure[]{
\includegraphics[clip,trim=90mm 50mm 130mm 20mm, width=.3\linewidth]{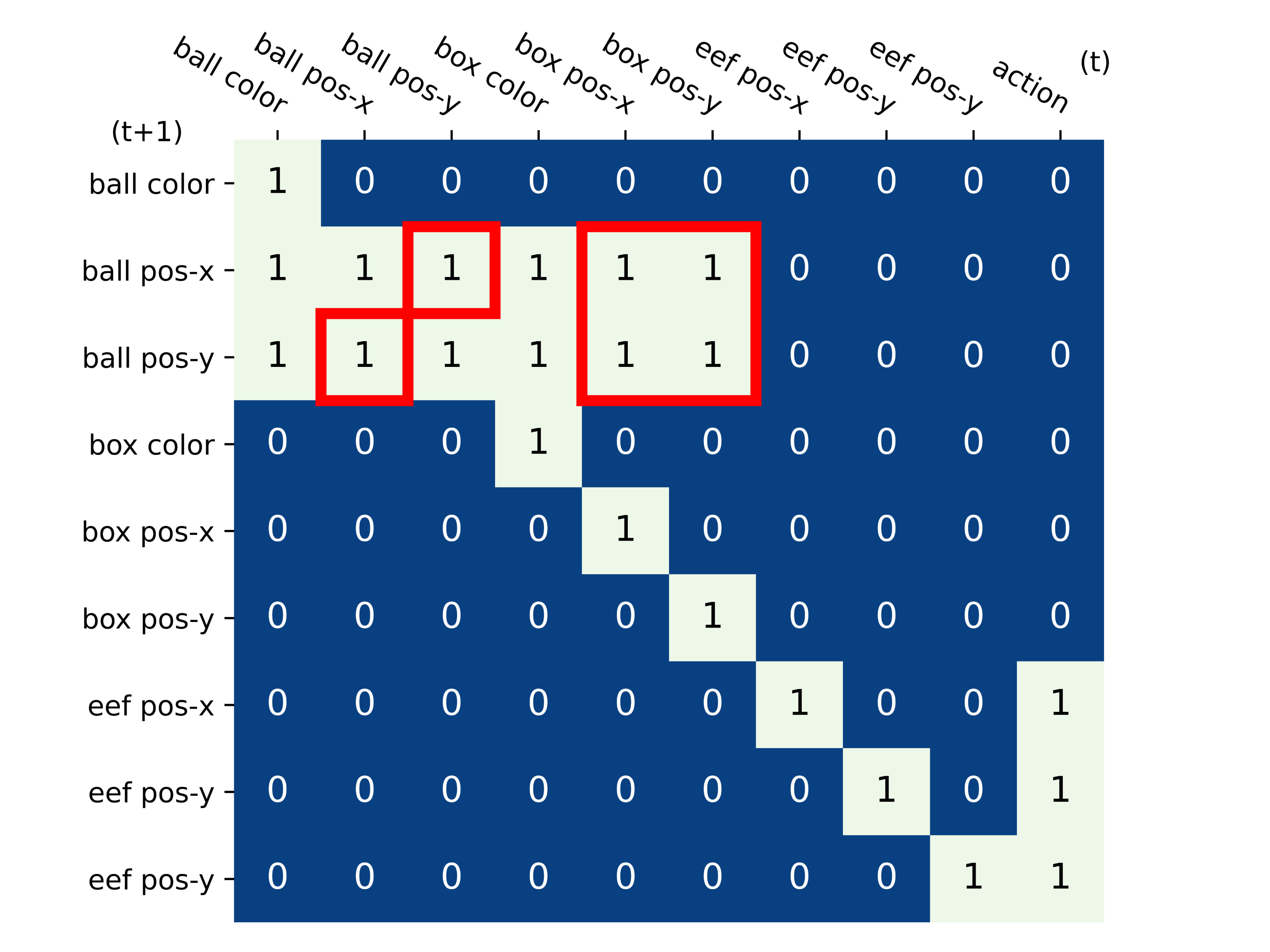}
\vspace{-15pt}
\label{fig:magnetic_gt_gcg}
}\hfill
\subfigure[]{
\includegraphics[clip,trim=90mm 50mm 130mm 20mm, width=.3\linewidth]{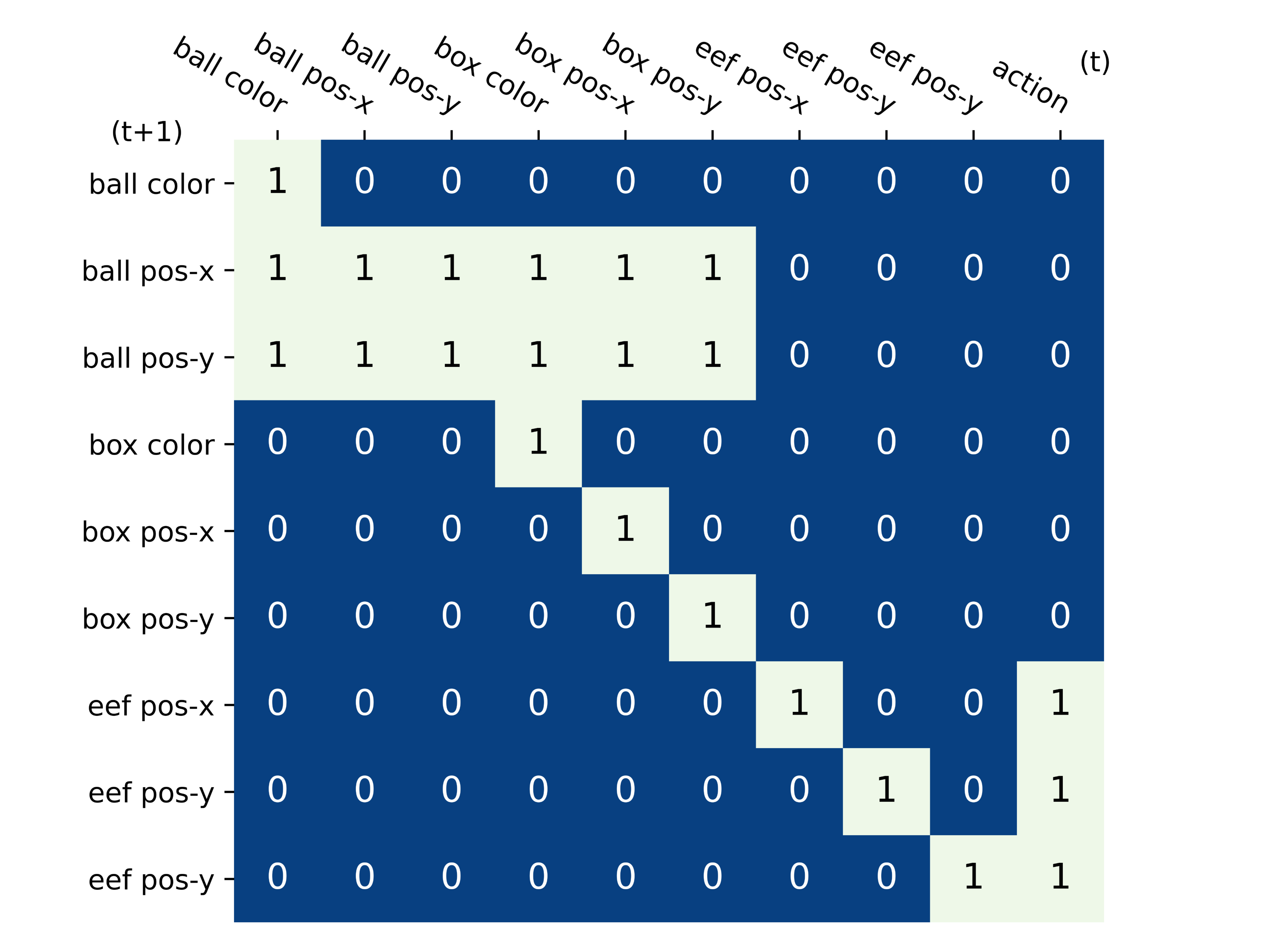}
\vspace{-15pt}
\label{fig:magnetic_gt_lcg1}
}\hfill
\subfigure[]{
\includegraphics[clip,trim=90mm 50mm 130mm 20mm, width=.3\linewidth]{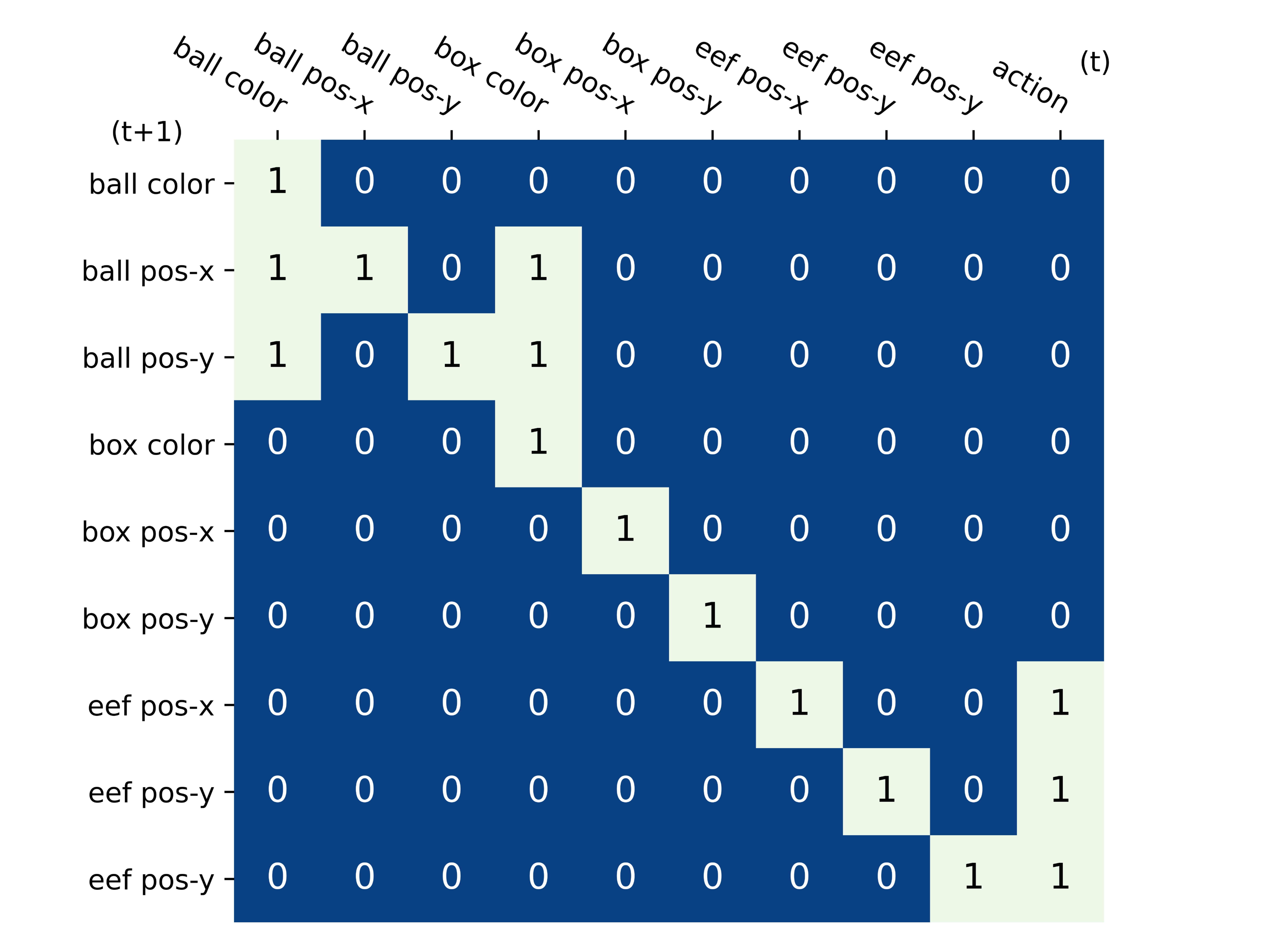}
\vspace{-15pt}
\label{fig:magnetic_gt_lcg2}
}
\vspace{-3mm}
\caption{(a) Causal graph of {Magnetic} environment.
Red boxes indicate redundant edges under the non-magnetic context. (b) LCG under the magnetic context, which is the same as global CG. (c) LCG under the non-magnetic context. }
\label{fig:magnetic_gt}
\vspace{-3mm}
\end{figure*}

%% file: figure/comparison_ncd_ours.tex
\begin{figure}[t!]%
\centering
\includegraphics[clip, trim=0mm 5mm 0mm 0mm, width=0.95\textwidth]{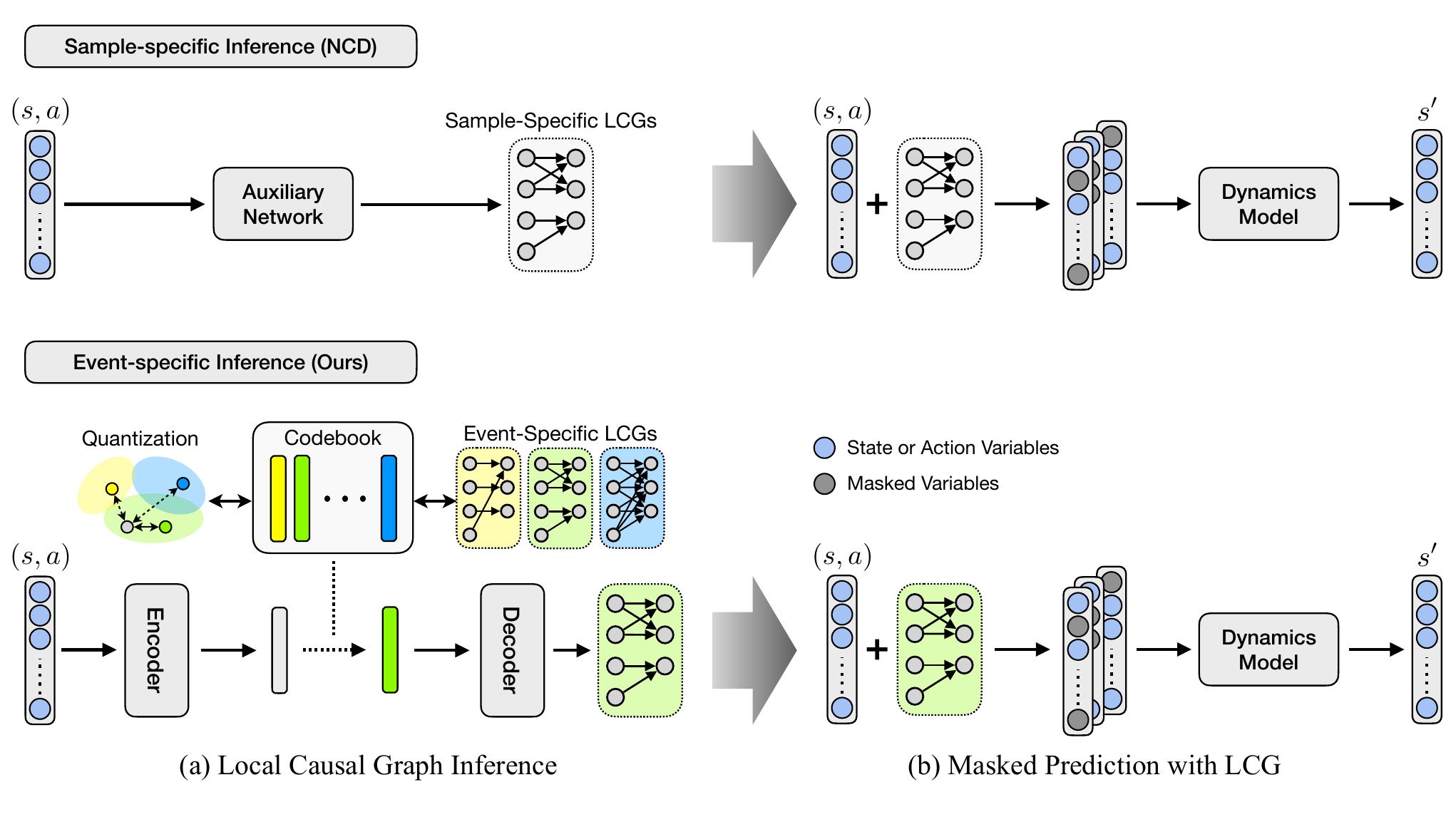}
\vspace{-3mm}
\caption{Comparison of  the sample-specific inference of NCD \textbf{(top)} and quantization-based inference of our method \textbf{(bottom)}.}
\vspace{-3mm}
\label{fig:comparison_ncd_ours}
\end{figure}

%% file: figure/lcg_chemical_k16.tex
\begin{figure*}[t!]
\centering
\subfigure[]{%
\includegraphics[clip,trim=4mm 0mm 4mm 0, height=0.18\textwidth]{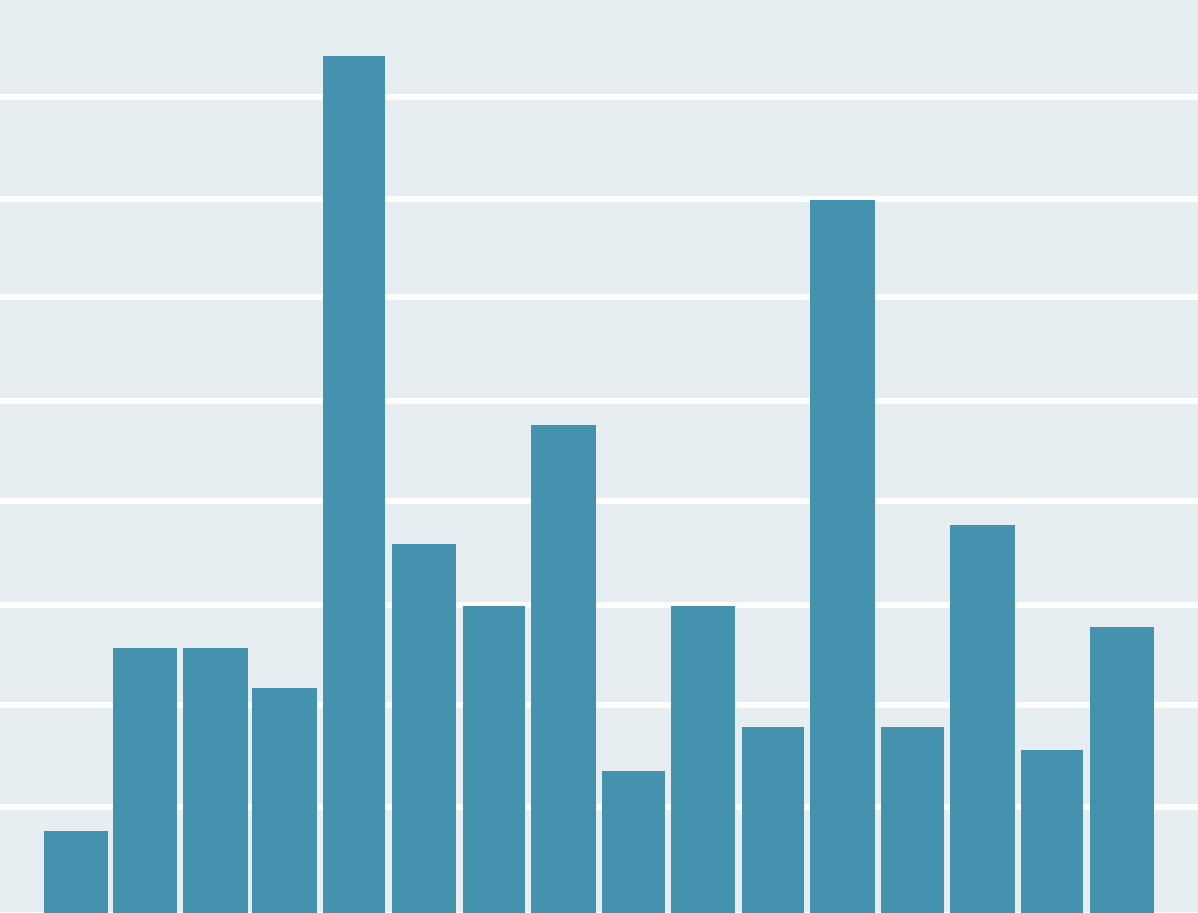}%
\label{fig:lcg_chemical_a}%
}\hspace{5mm}%
\subfigure[]{%
\includegraphics[clip,trim=4mm 0mm 4mm 0,height=0.18\textwidth]{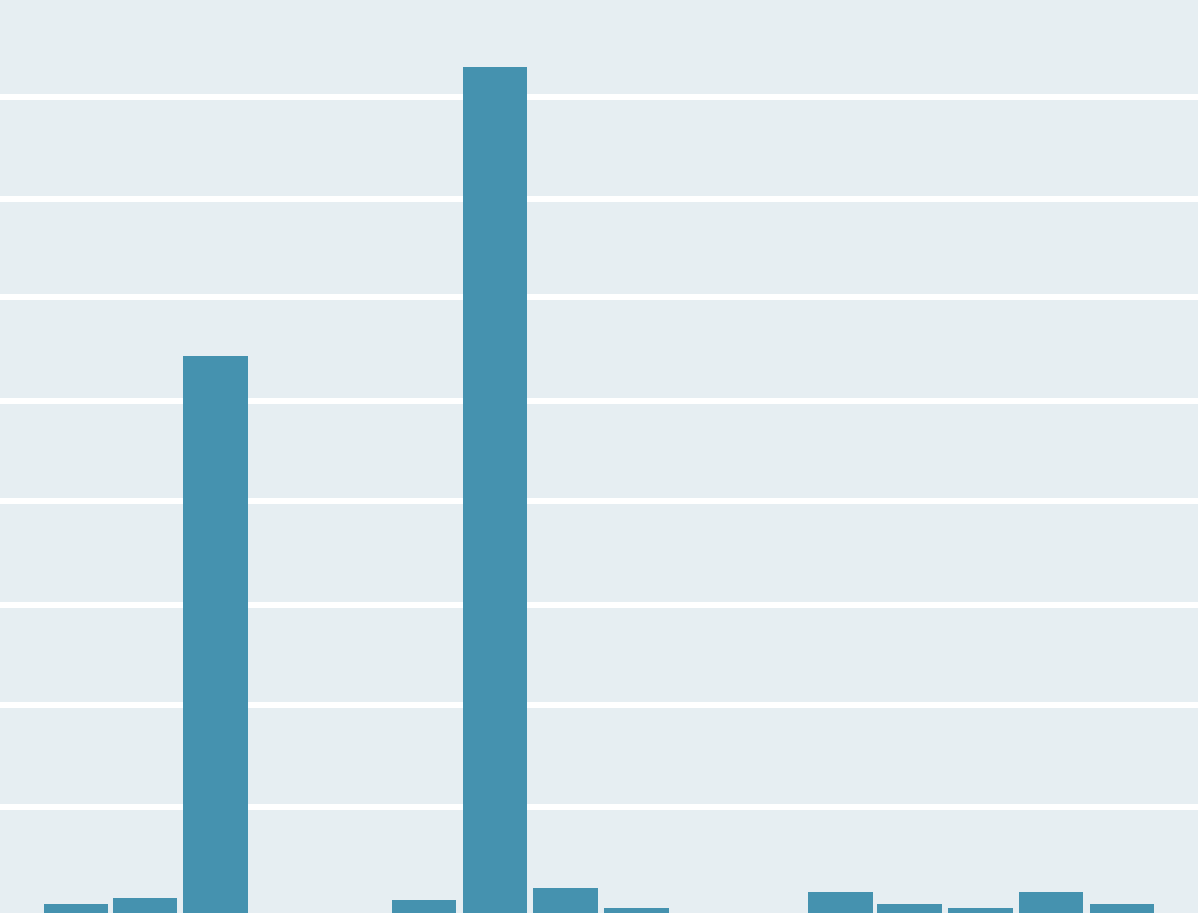}%
\label{fig:lcg_chemical_b}%
}\hfill%
\subfigure[]{%
\includegraphics[clip,trim=130mm 20mm 140mm 4mm, height=0.18\textwidth]{figure/source/exp/chemical/codebook/fork.pdf}%
\label{fig:lcg_chemical_c}%
}\hspace{5mm}%
\subfigure[]{%
\includegraphics[clip,trim=40mm 20mm 70mm 10mm, height=0.18\textwidth]{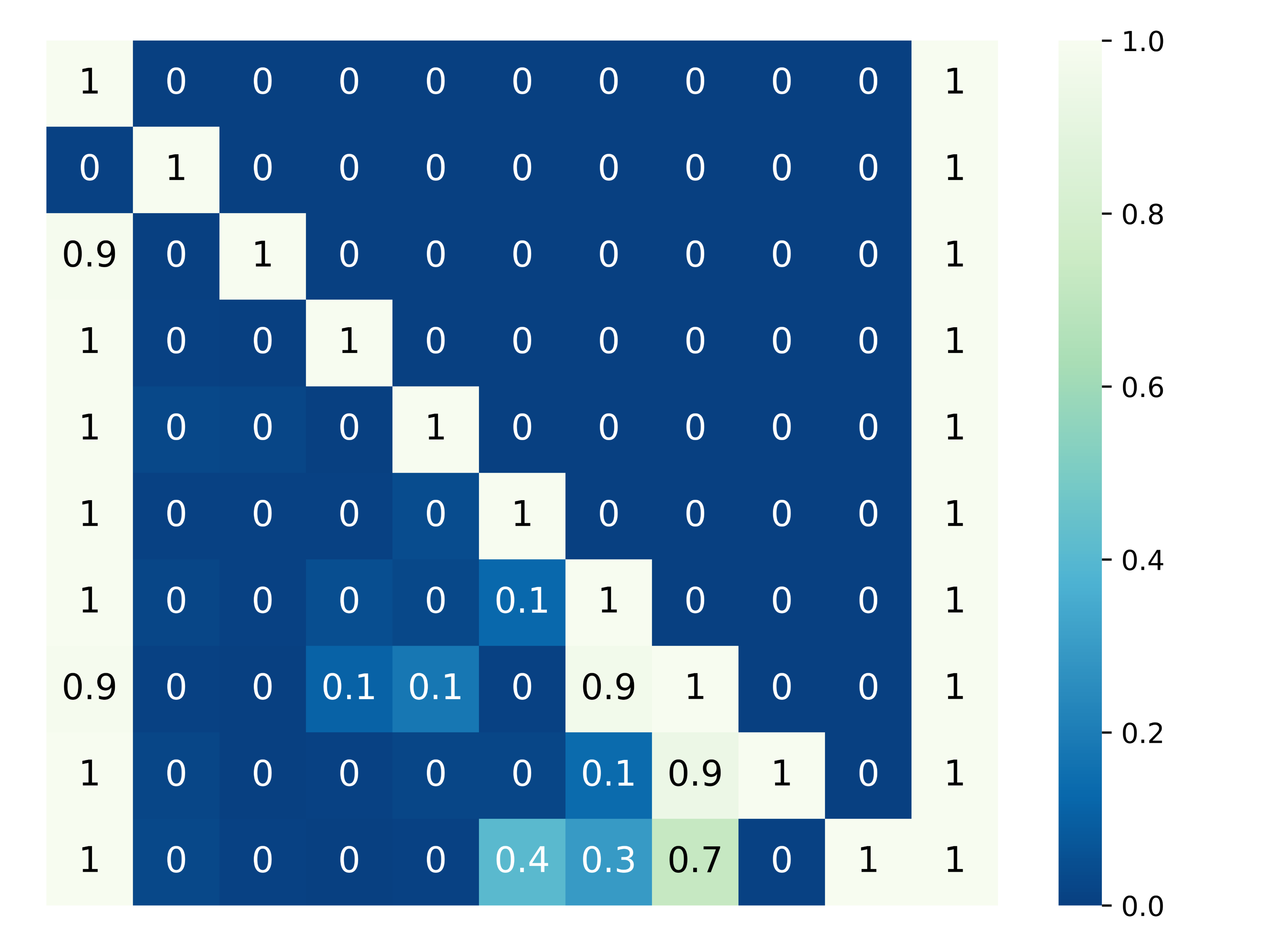}%
\label{fig:lcg_chemical_d}%
}
\vspace{-2mm}
\caption{(a,b) Codebook histogram on (a) ID states during training and (b) OOD states during the test in {Chemical} (\textit{full-fork}). 
(c) True causal graph of the \textit{fork} structure. 
(d) Learned LCG corresponding to the most used code in (b).}
\vspace{-5mm}
\label{fig:lcg_chemical_k16}
\end{figure*}

%% file: figure/lcg_chemical_analysis_k04_s1.tex
\begin{figure*}[t!]
\centering
\subfigure[]{%
\includegraphics[clip,trim=4mm 0mm 4mm 0, height=0.15\textwidth]{figure/source/exp/chemical/codebook_analysis/k04-s1-suc/histogram-all.pdf}%
\label{fig:lcg_chemical_analysis_k04_s1_a}%
}
\hspace{5mm}%
\subfigure[]{%
\includegraphics[clip,trim=4mm 0mm 4mm 0,height=0.15\textwidth]{figure/source/exp/chemical/codebook_analysis/k04-s1-suc/histogram-local.pdf}%
\label{fig:lcg_chemical_analysis_k04_s1_b}%
}
\hspace{5mm}%
\subfigure[]{%
\includegraphics[clip,trim=4mm 0mm 4mm 0,height=0.15\textwidth]{figure/source/exp/chemical/codebook_analysis/k04-s1-suc/histogram-local-ood.pdf}%
\label{fig:lcg_chemical_analysis_k04_s1_c}%
}
\hfill%
\\
\vspace{-7pt}
\subfigure[]{%
\includegraphics[clip,trim=130mm 20mm 140mm 4mm, height=0.2\textwidth]{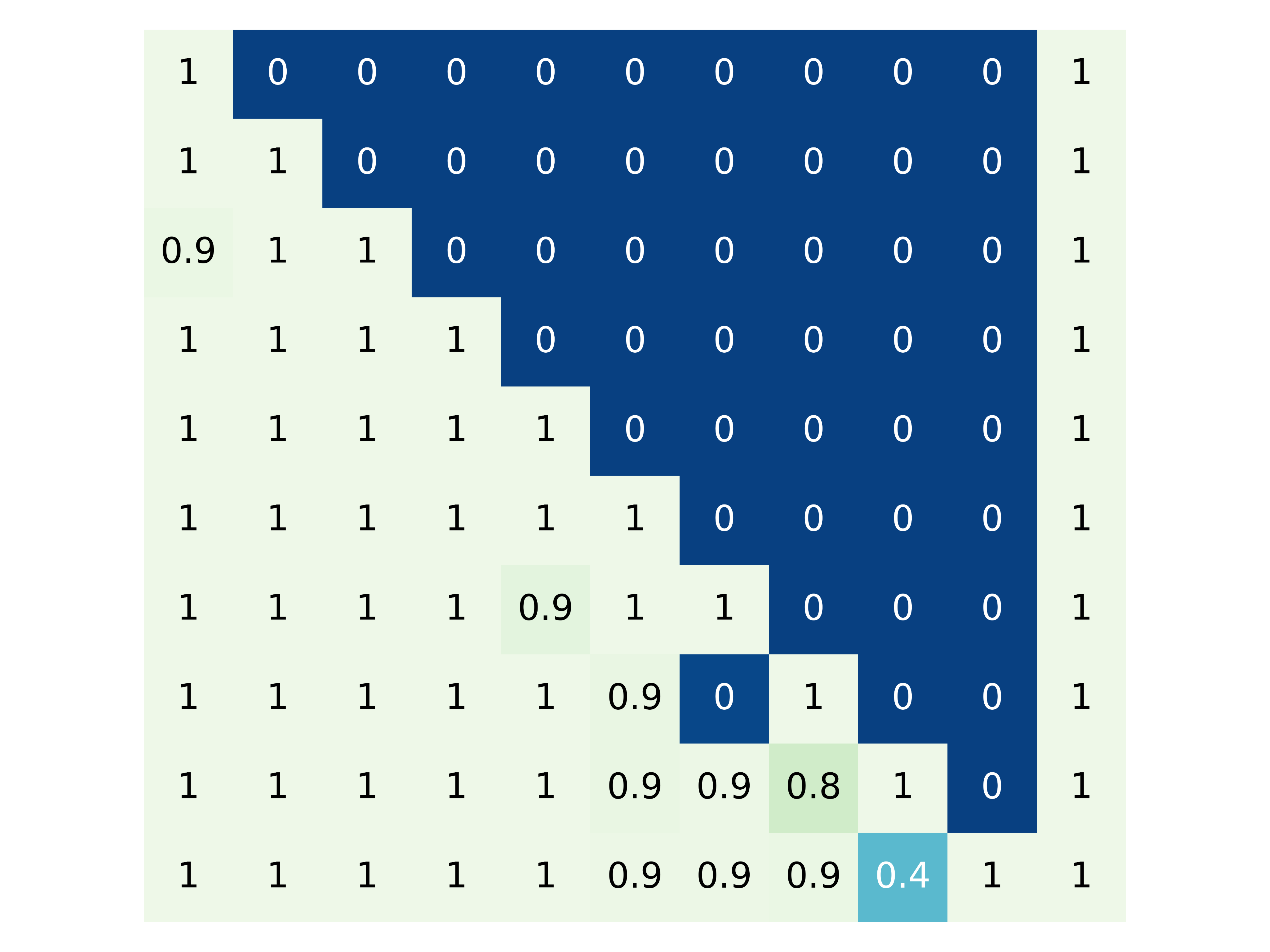}%
\label{fig:lcg_chemical_analysis_k04_s1_d}%
}
\hfill
\subfigure[]{%
\includegraphics[clip,trim=130mm 20mm 140mm 4mm, height=0.2\textwidth]{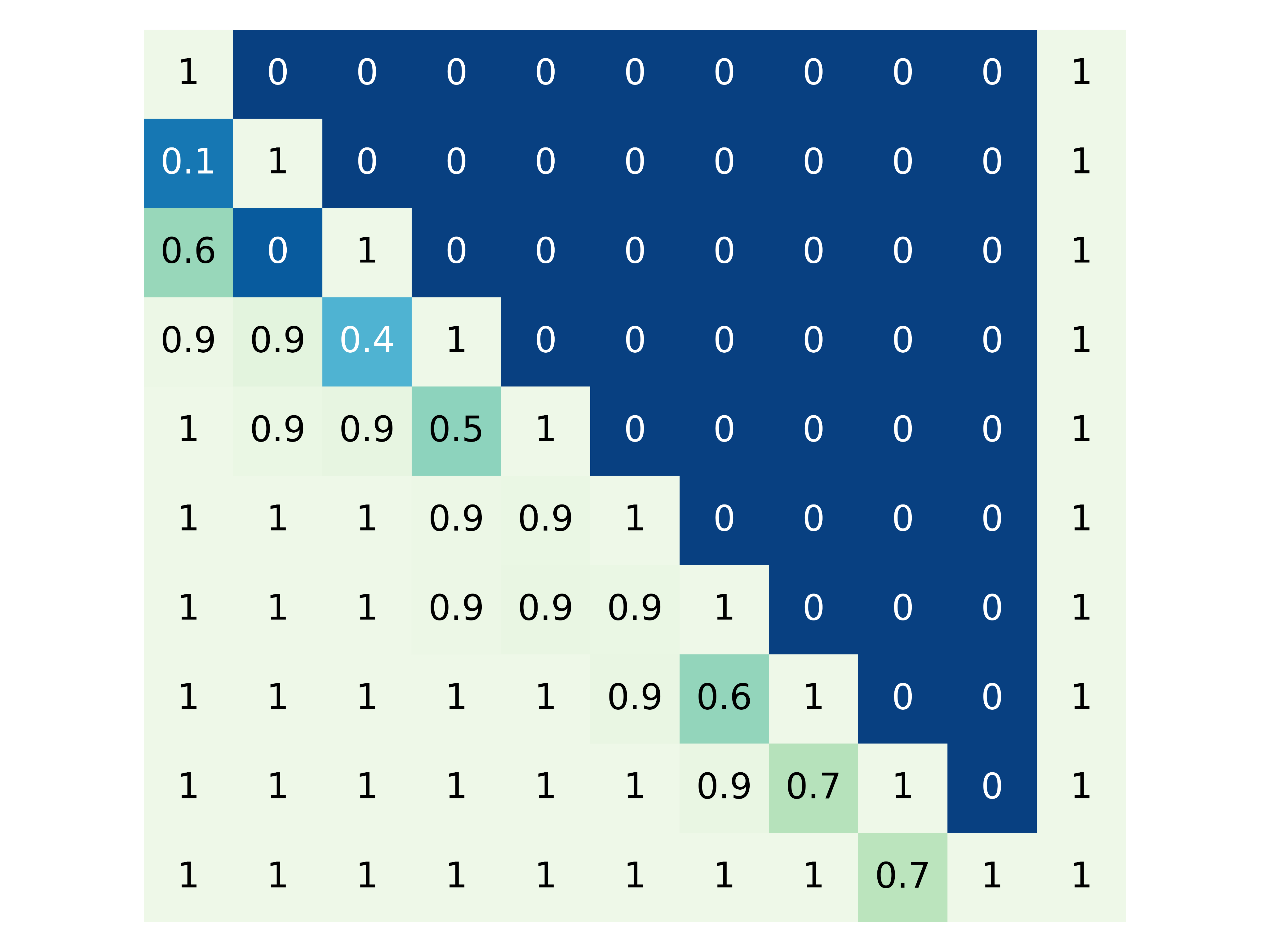}%
\label{fig:lcg_chemical_analysis_k04_s1_e}%
}
\hfill
\subfigure[]{%
\includegraphics[clip,trim=130mm 20mm 140mm 4mm, height=0.2\textwidth]{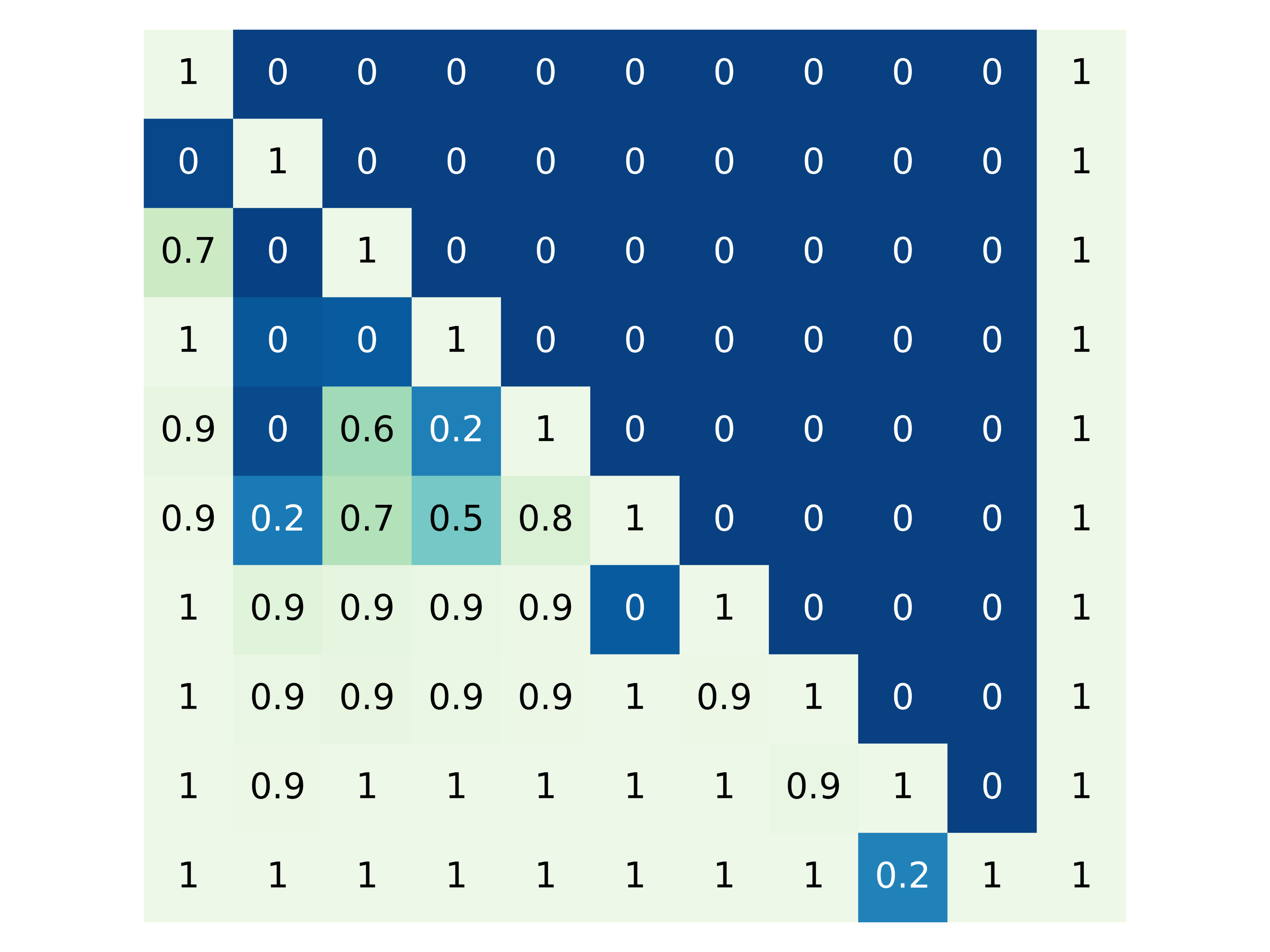}%
\label{fig:lcg_chemical_analysis_k04_s1_f}%
}
\hfill
\subfigure[]{%
\includegraphics[clip,trim=40mm 20mm 70mm 10mm, height=0.2\textwidth]{figure/source/exp/chemical/codebook_analysis/k04-s1-suc/code-03.pdf}%
\label{fig:lcg_chemical_analysis_k04_s1_g}%
}
\vspace{-7pt}
\caption{Analysis of LCGs learned by our method with quantization degree of 4 in Chemical (\textit{full-fork}) environment.
(a-c) Codebook histogram on (a) ID states, (b) ID states on local structure \textit{fork}, and (c) OOD states on local structure.
(d-g) Learned LCGs. The descriptions of the histograms are also applied to \Cref{fig:lcg_chemical_analysis_k04_s4,fig:lcg_magnetic_analysis_k04_s1,fig:lcg_magnetic_analysis_k04_s4,fig:lcg_chemical_analysis_k02_s1,fig:lcg_chemical_analysis_k02_s2}. }
\vspace{-7pt}
\label{fig:lcg_chemical_analysis_k04_s1}
\end{figure*}

%% file: figure/lcg_chemical_analysis_k04_s4.tex
\begin{figure*}[t!]
\centering
\subfigure[]{%
\includegraphics[clip,trim=4mm 0mm 4mm 0, height=0.15\textwidth]{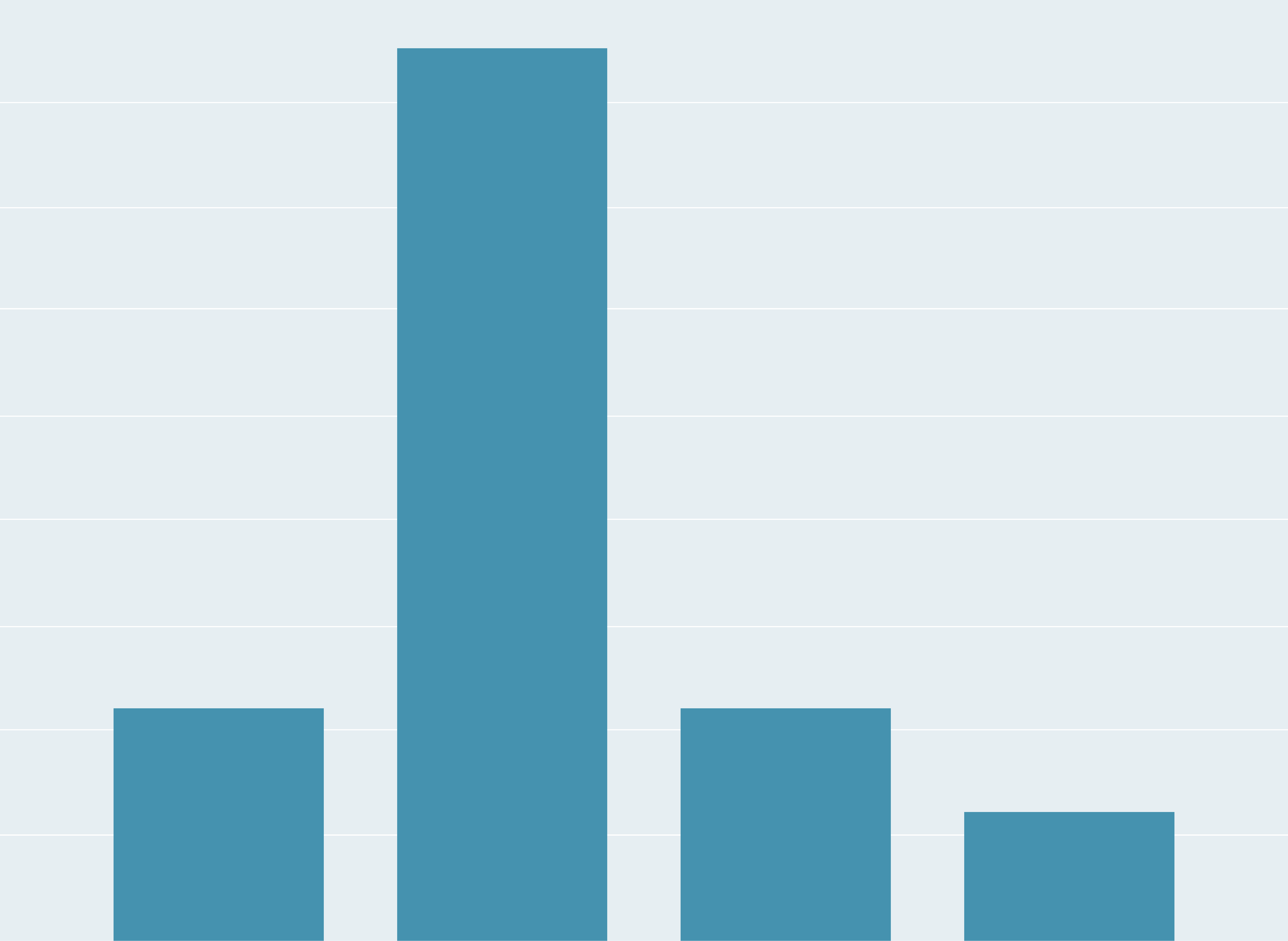}%
\label{fig:lcg_chemical_analysis_k04_s4_a}%
}
\hspace{5mm}%
\subfigure[]{%
\includegraphics[clip,trim=4mm 0mm 4mm 0,height=0.15\textwidth]{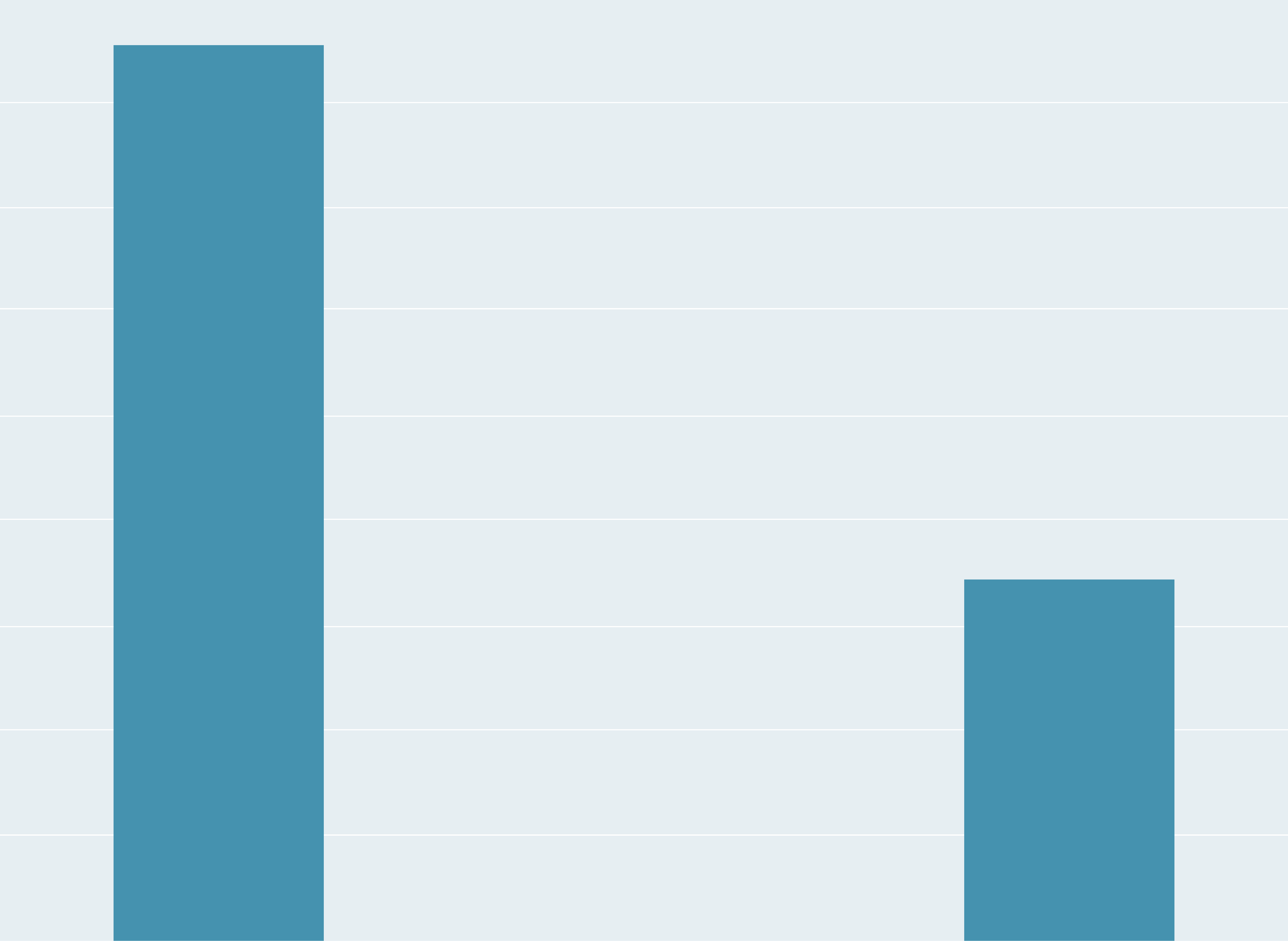}%
\label{fig:lcg_chemical_analysis_k04_s4_b}%
}
\hspace{5mm}%
\subfigure[]{%
\includegraphics[clip,trim=4mm 0mm 4mm 0,height=0.15\textwidth]{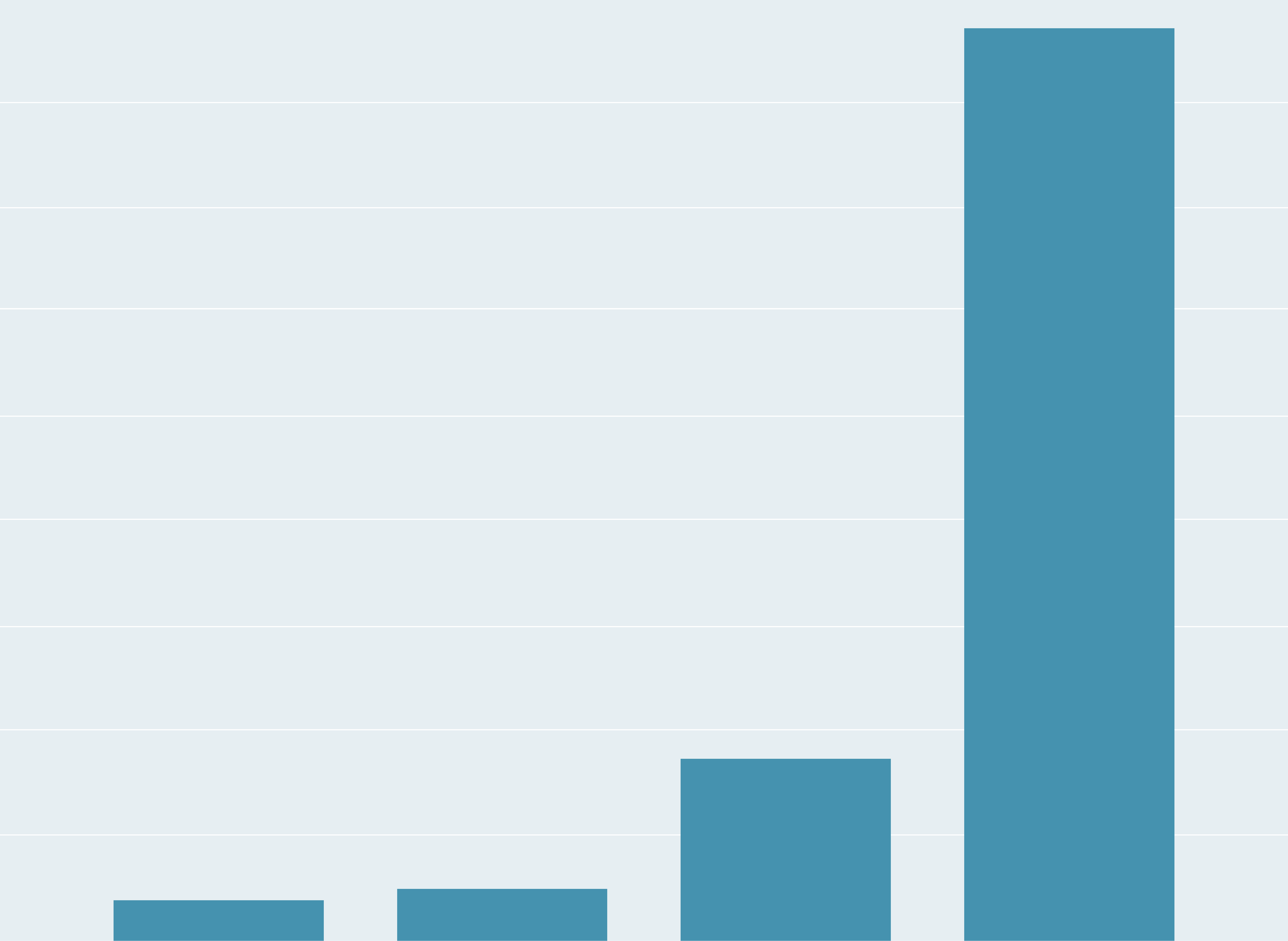}%
\label{fig:lcg_chemical_analysis_k04_s4_c}%
}
\hfill%
\\
\vspace{-7pt}
\subfigure[]{%
\includegraphics[clip,trim=130mm 20mm 140mm 4mm, height=0.2\textwidth]{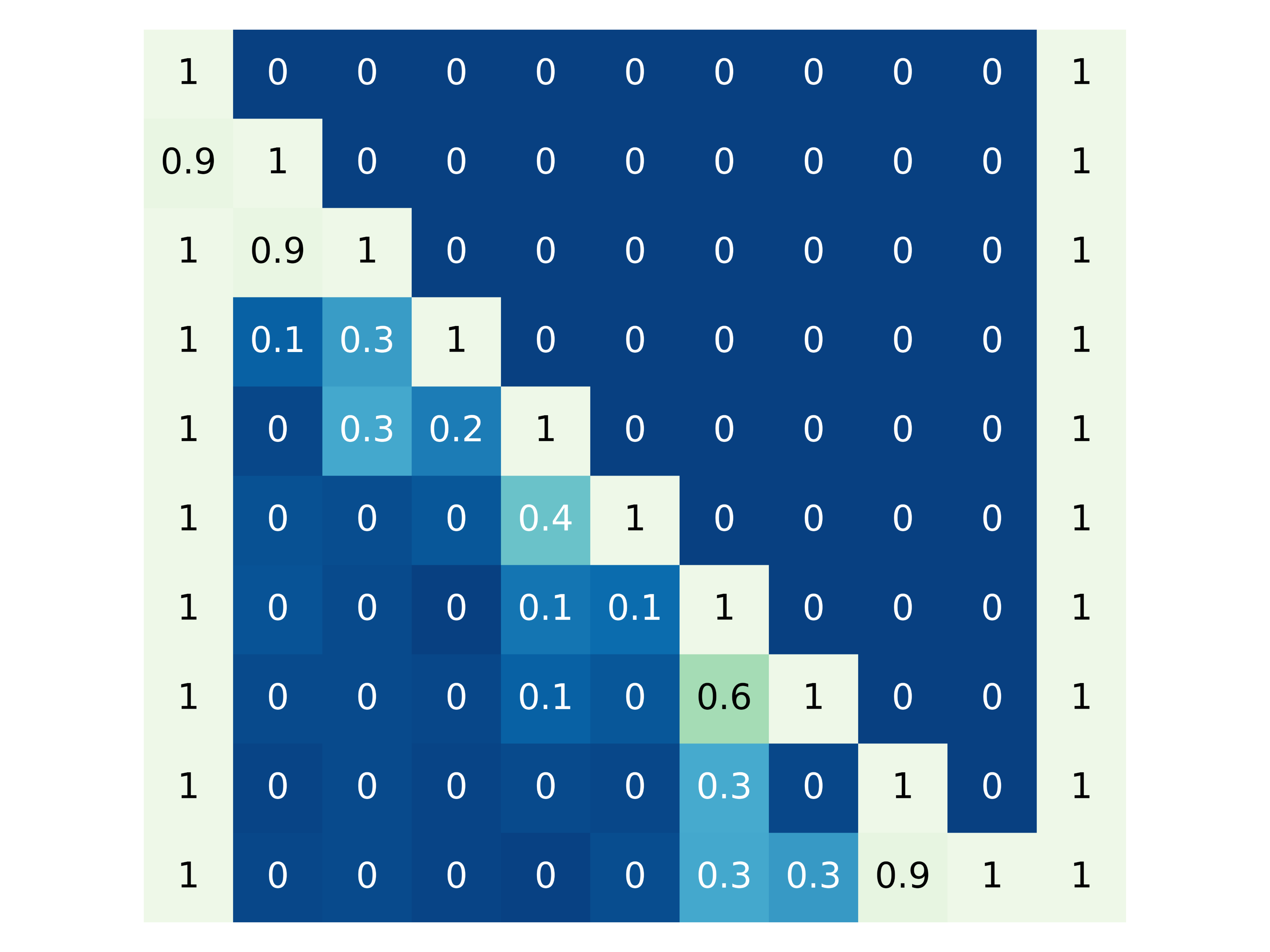}%
\label{fig:lcg_chemical_analysis_k04_s4_d}%
}
\hfill
\subfigure[]{%
\includegraphics[clip,trim=130mm 20mm 140mm 4mm, height=0.2\textwidth]{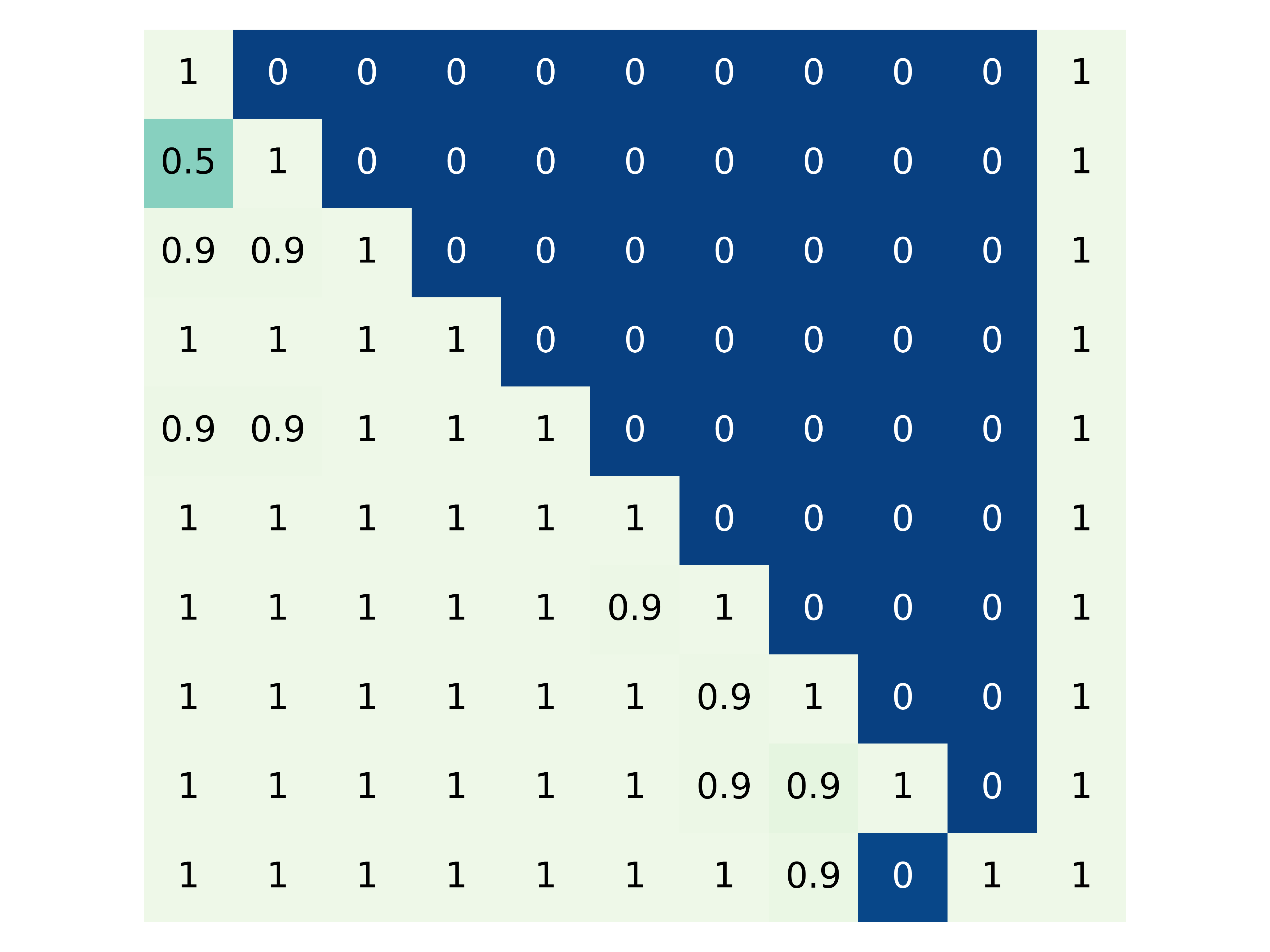}%
\label{fig:lcg_chemical_analysis_k04_s4_e}%
}
\hfill
\subfigure[]{%
\includegraphics[clip,trim=130mm 20mm 140mm 4mm, height=0.2\textwidth]{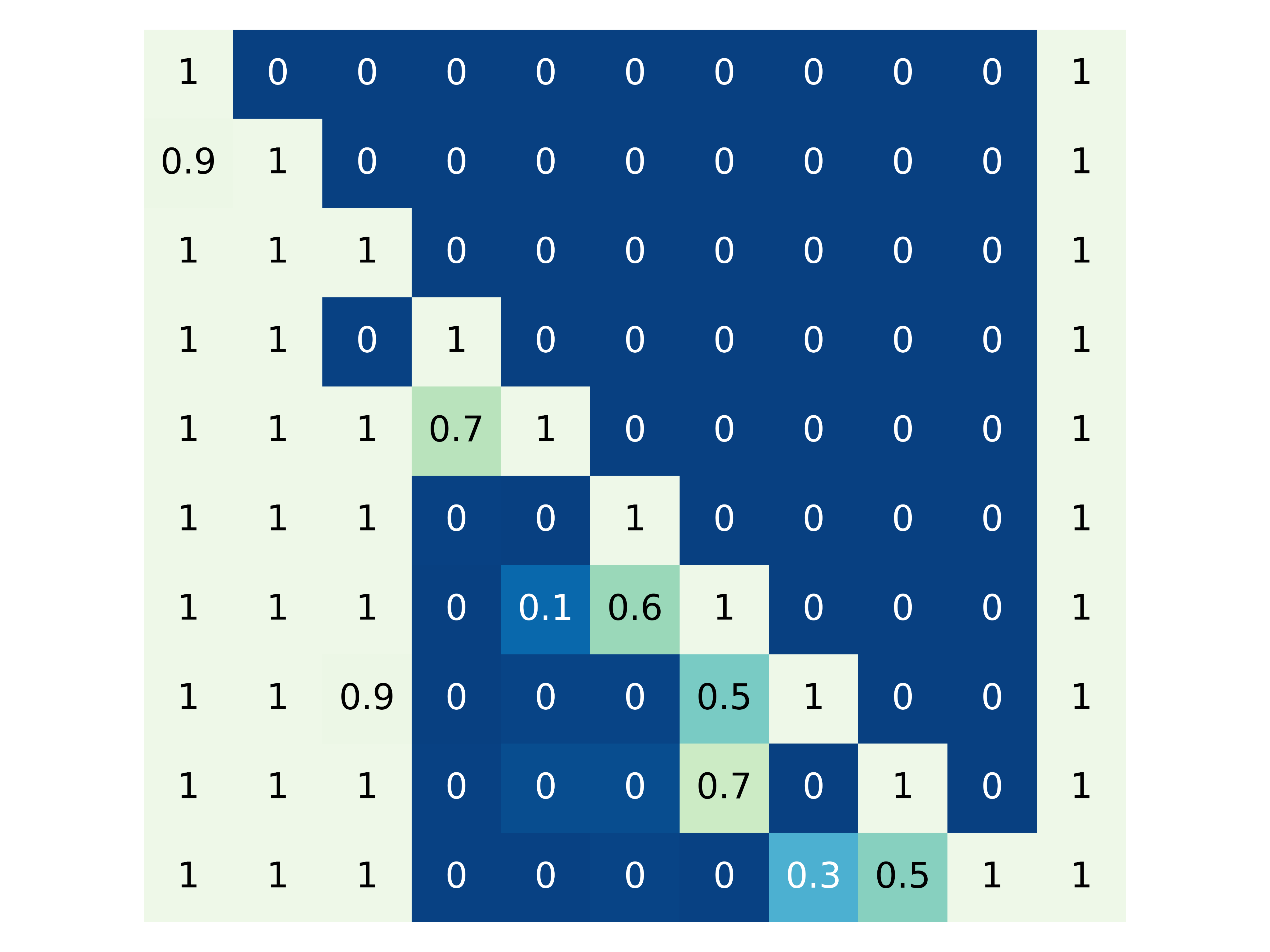}%
\label{fig:lcg_chemical_analysis_k04_s4_f}%
}
\hfill
\subfigure[]{%
\includegraphics[clip,trim=40mm 20mm 70mm 10mm, height=0.2\textwidth]{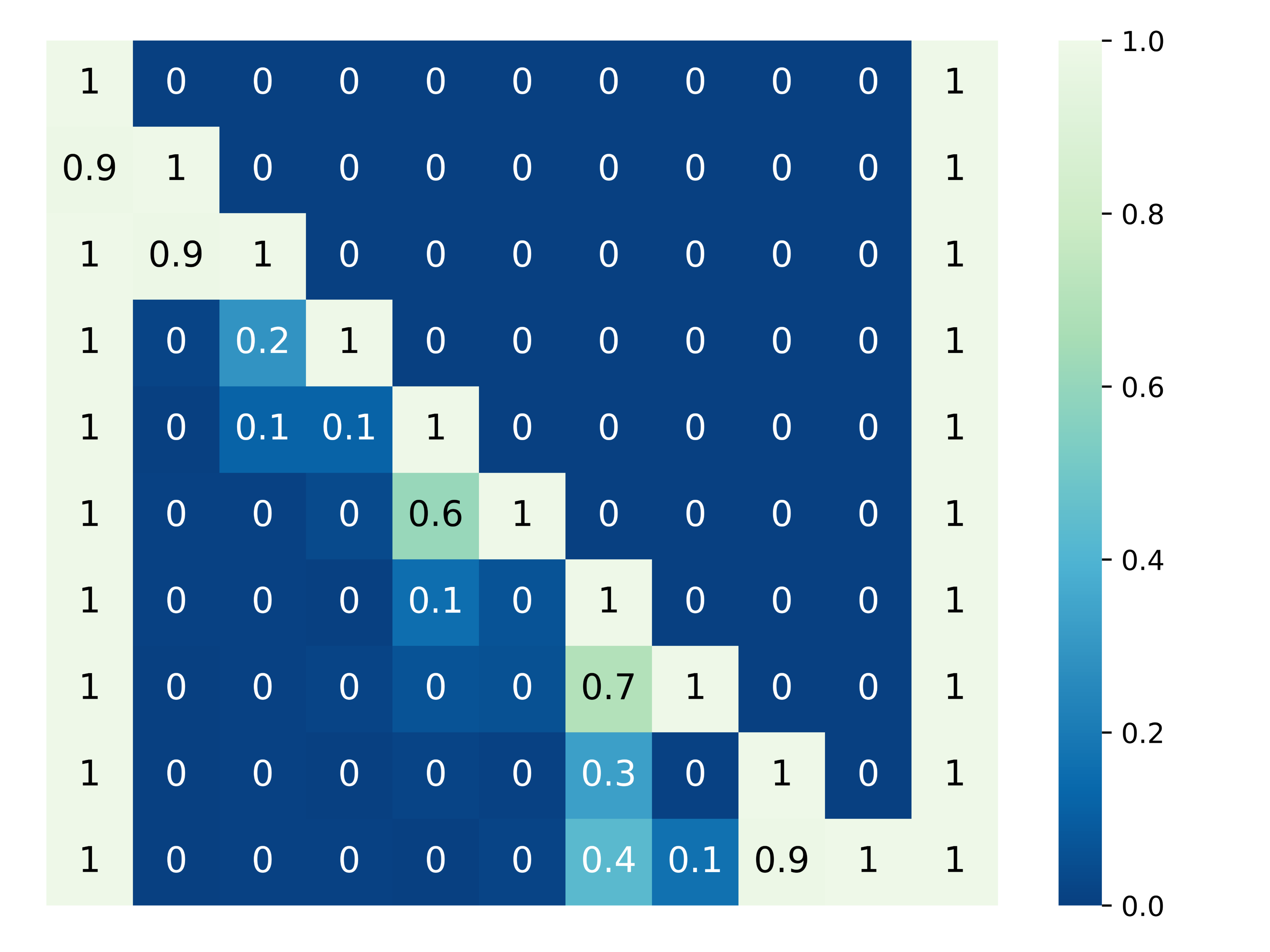}%
\label{fig:lcg_chemical_analysis_k04_s4_g}%
}
\vspace{-7pt}
\caption{Another sample run of our method with quantization degree of 4 in Chemical (\textit{full-fork}).
}
\vspace{-7pt}
\label{fig:lcg_chemical_analysis_k04_s4}
\end{figure*}

%% file: figure/lcg_magnetic_analysis_k04_s1.tex
\begin{figure*}[t!]
\centering
\subfigure[]{%
\includegraphics[clip,trim=4mm 0mm 4mm 0, height=0.15\textwidth]{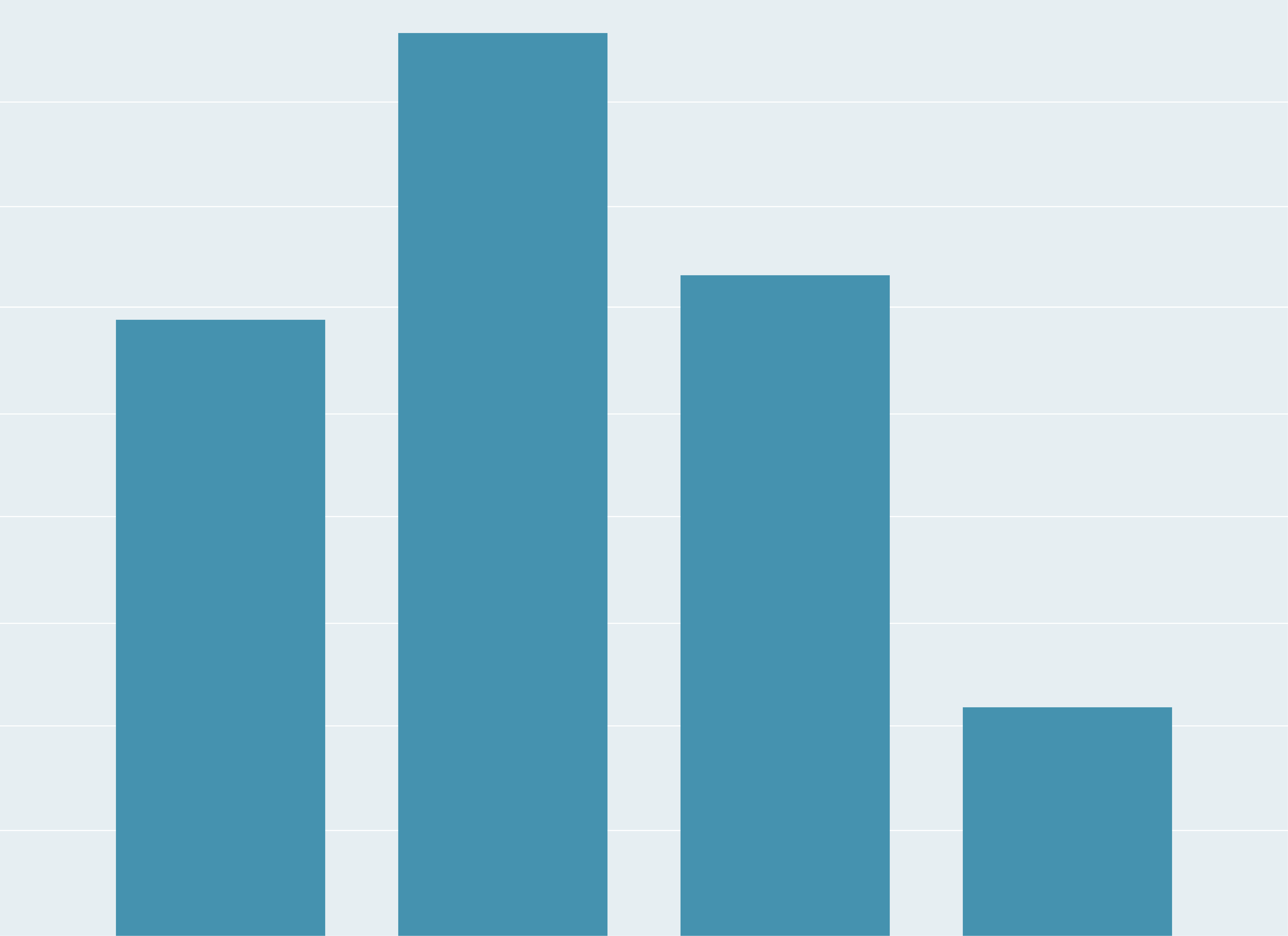}%
\label{fig:lcg_magnetic_analysis_k04_s1_a}%
}
\hspace{5mm}%
\subfigure[]{%
\includegraphics[clip,trim=4mm 0mm 4mm 0,height=0.15\textwidth]{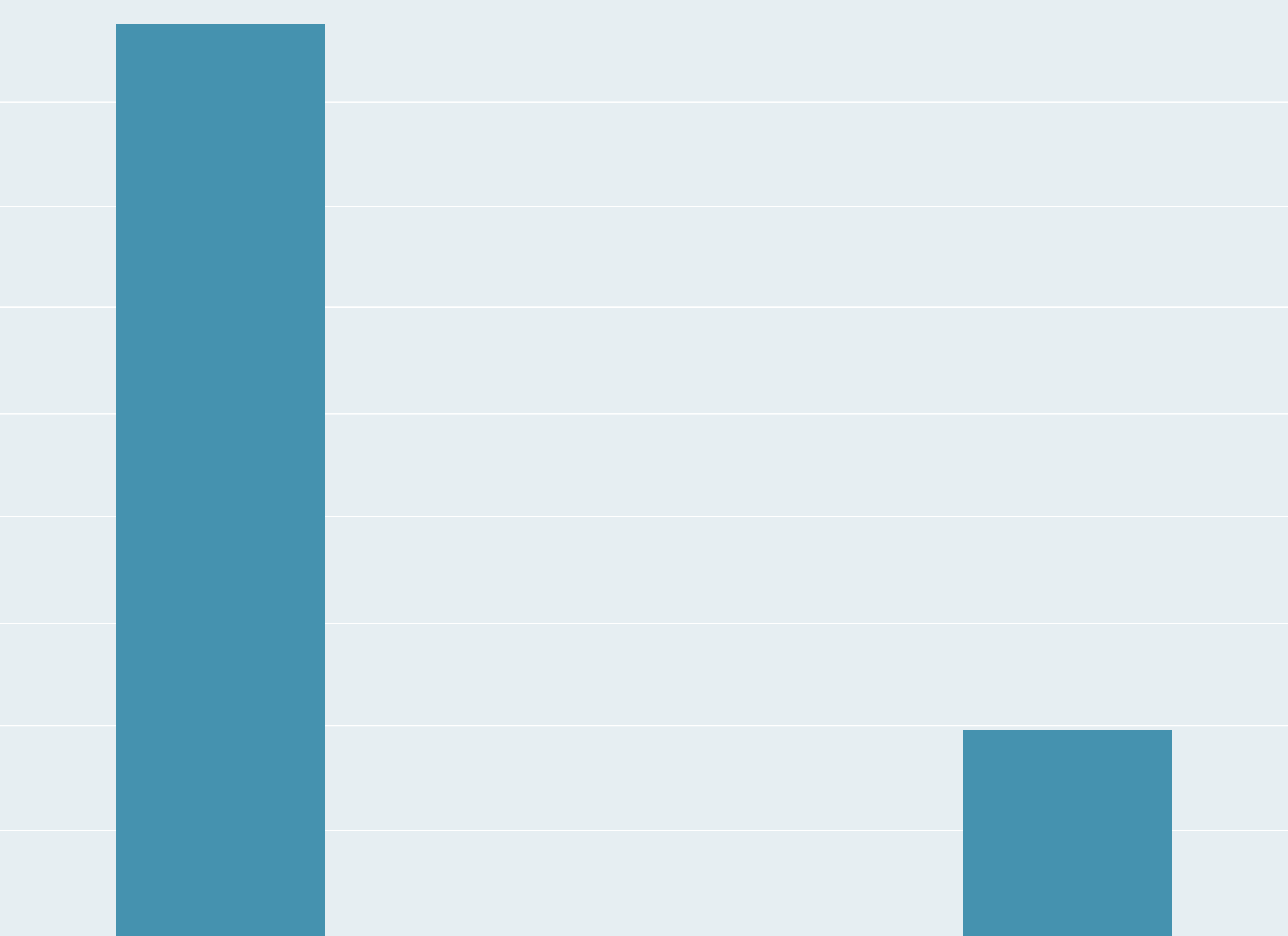}%
\label{fig:lcg_magnetic_analysis_k04_s1_b}%
}
\hspace{5mm}%
\subfigure[]{%
\includegraphics[clip,trim=4mm 0mm 4mm 0,height=0.15\textwidth]{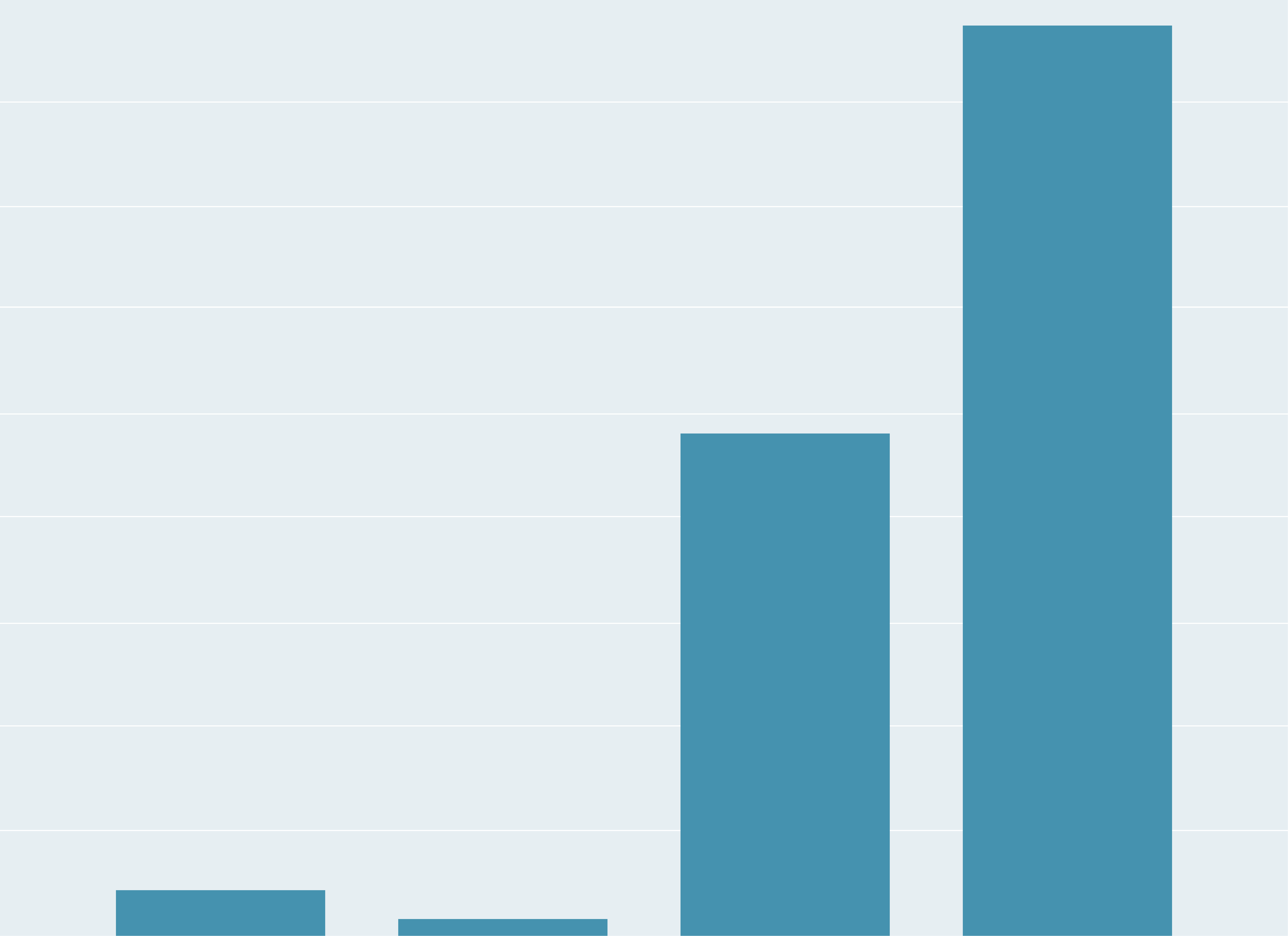}%
\label{fig:lcg_magnetic_analysis_k04_s1_c}%
}
\hfill%
\\
\vspace{-7pt}
\subfigure[]{%
\includegraphics[clip,trim=130mm 20mm 140mm 4mm, height=0.2\textwidth]{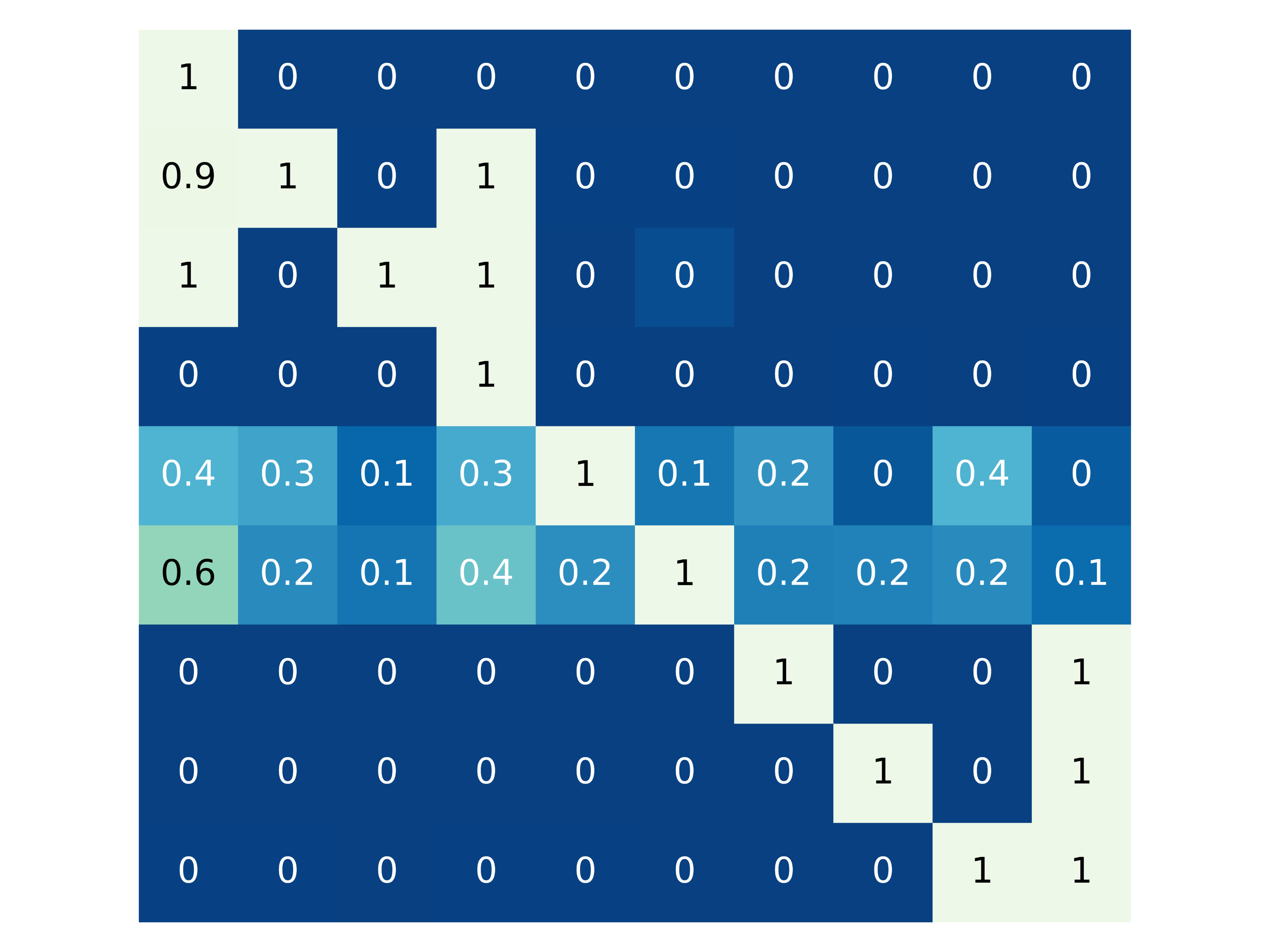}%
\label{fig:lcg_magnetic_analysis_k04_s1_d}%
}
\hfill
\subfigure[]{%
\includegraphics[clip,trim=130mm 20mm 140mm 4mm, height=0.2\textwidth]{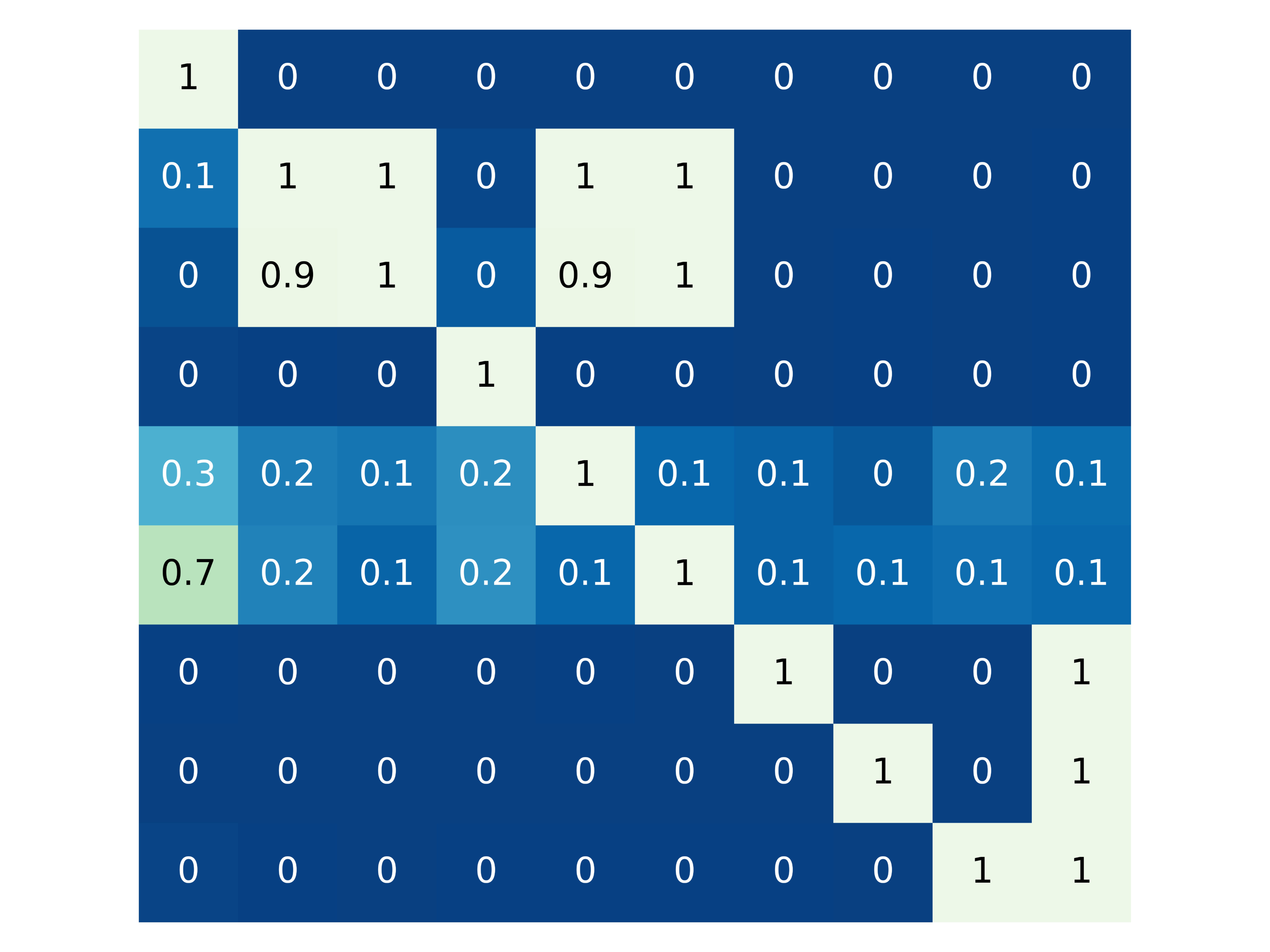}%
\label{fig:lcg_magnetic_analysis_k04_s1_e}%
}
\hfill
\subfigure[]{%
\includegraphics[clip,trim=130mm 20mm 140mm 4mm, height=0.2\textwidth]{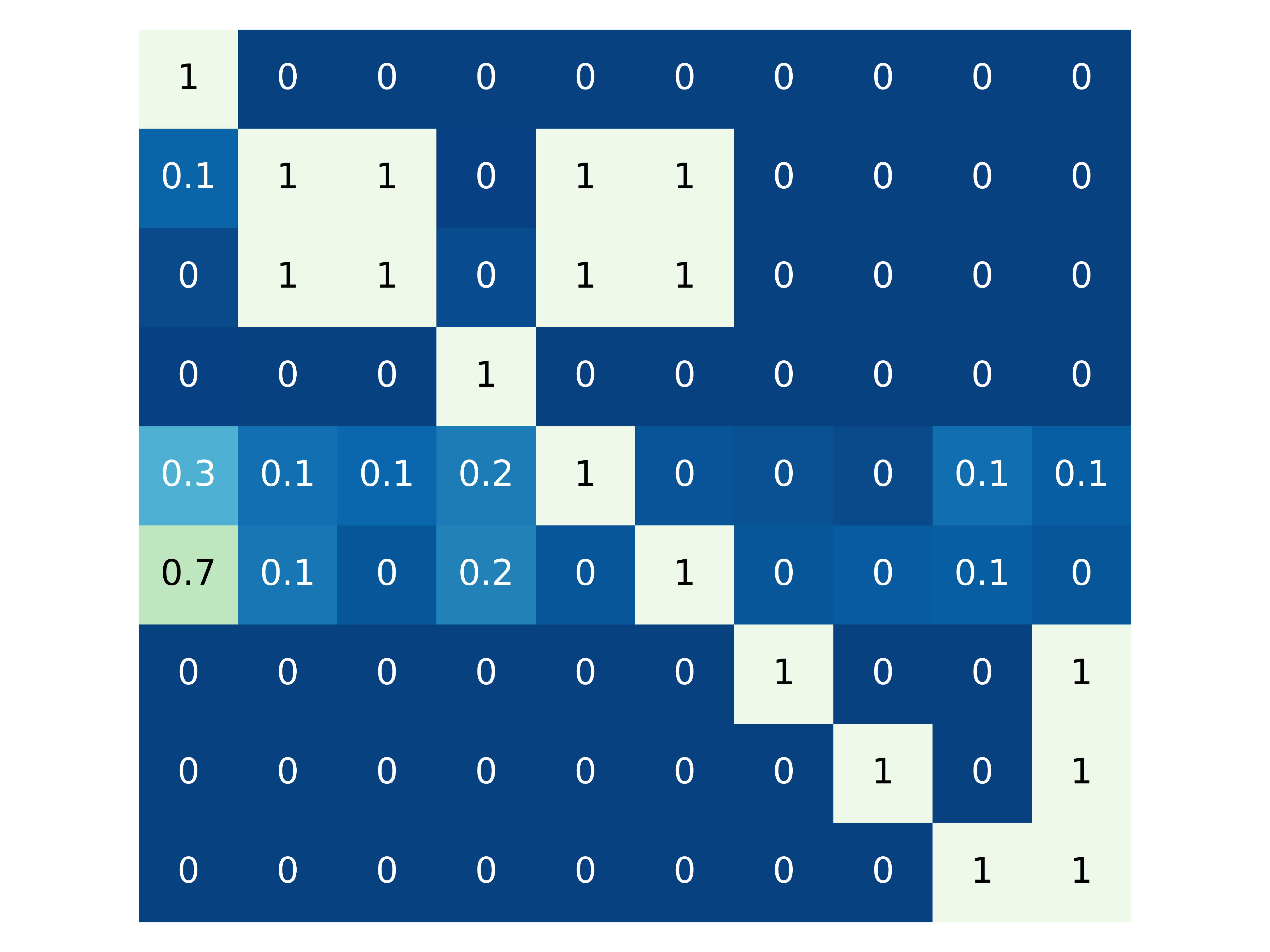}%
\label{fig:lcg_magnetic_analysis_k04_s1_f}%
}
\hfill
\subfigure[]{%
\includegraphics[clip,trim=40mm 20mm 70mm 10mm, height=0.2\textwidth]{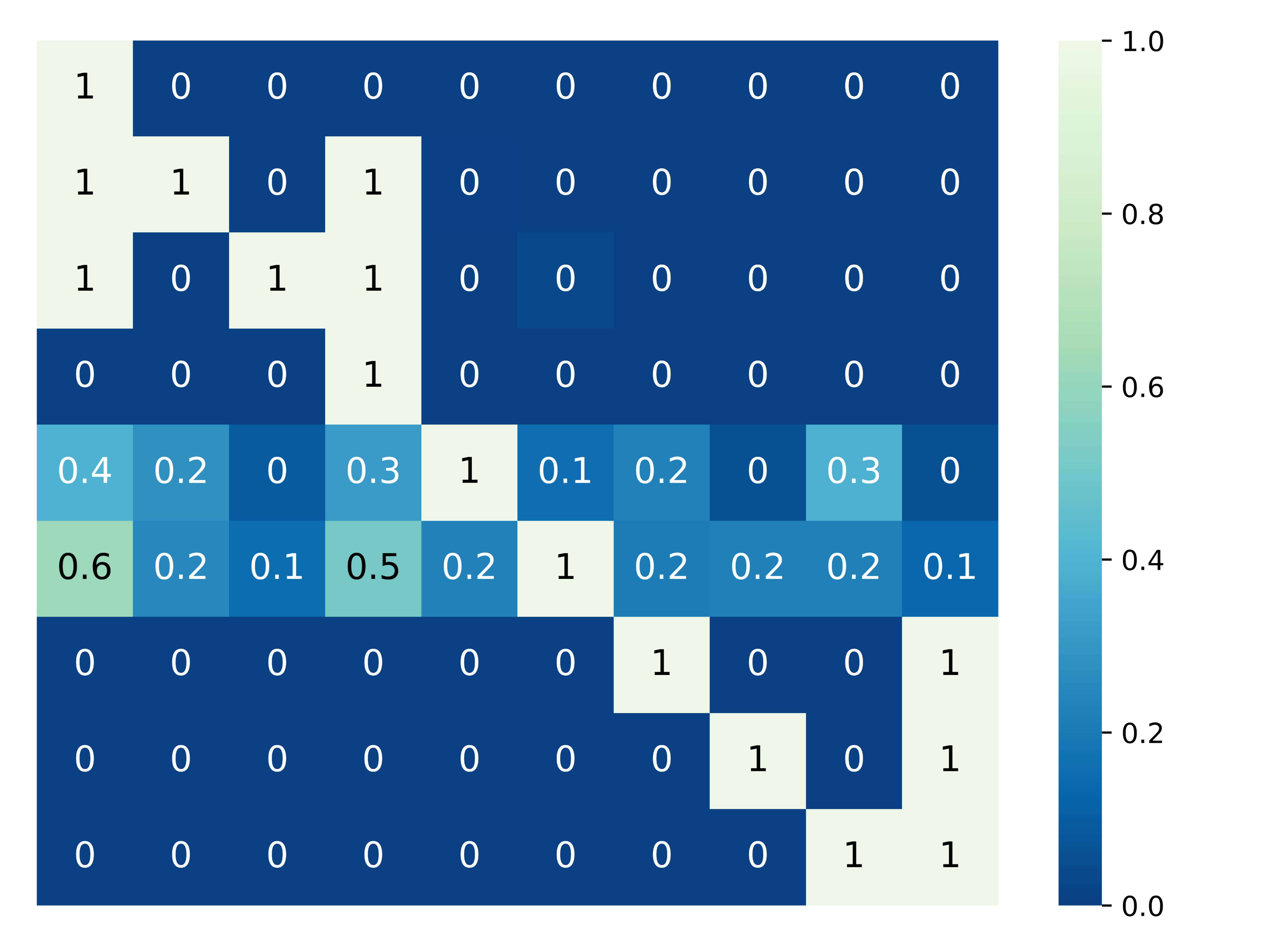}%
\label{fig:lcg_magnetic_analysis_k04_s1_g}%
}
\vspace{-7pt}
\caption{Analysis of LCGs learned by our method with quantization degree of 4 in Magnetic.
}
\vspace{-2pt}
\label{fig:lcg_magnetic_analysis_k04_s1}
\end{figure*}

%% file: figure/lcg_magnetic_analysis_k04_s4.tex
\begin{figure*}[t!]
\centering
\subfigure[]{%
\includegraphics[clip,trim=4mm 0mm 4mm 0, height=0.15\textwidth]{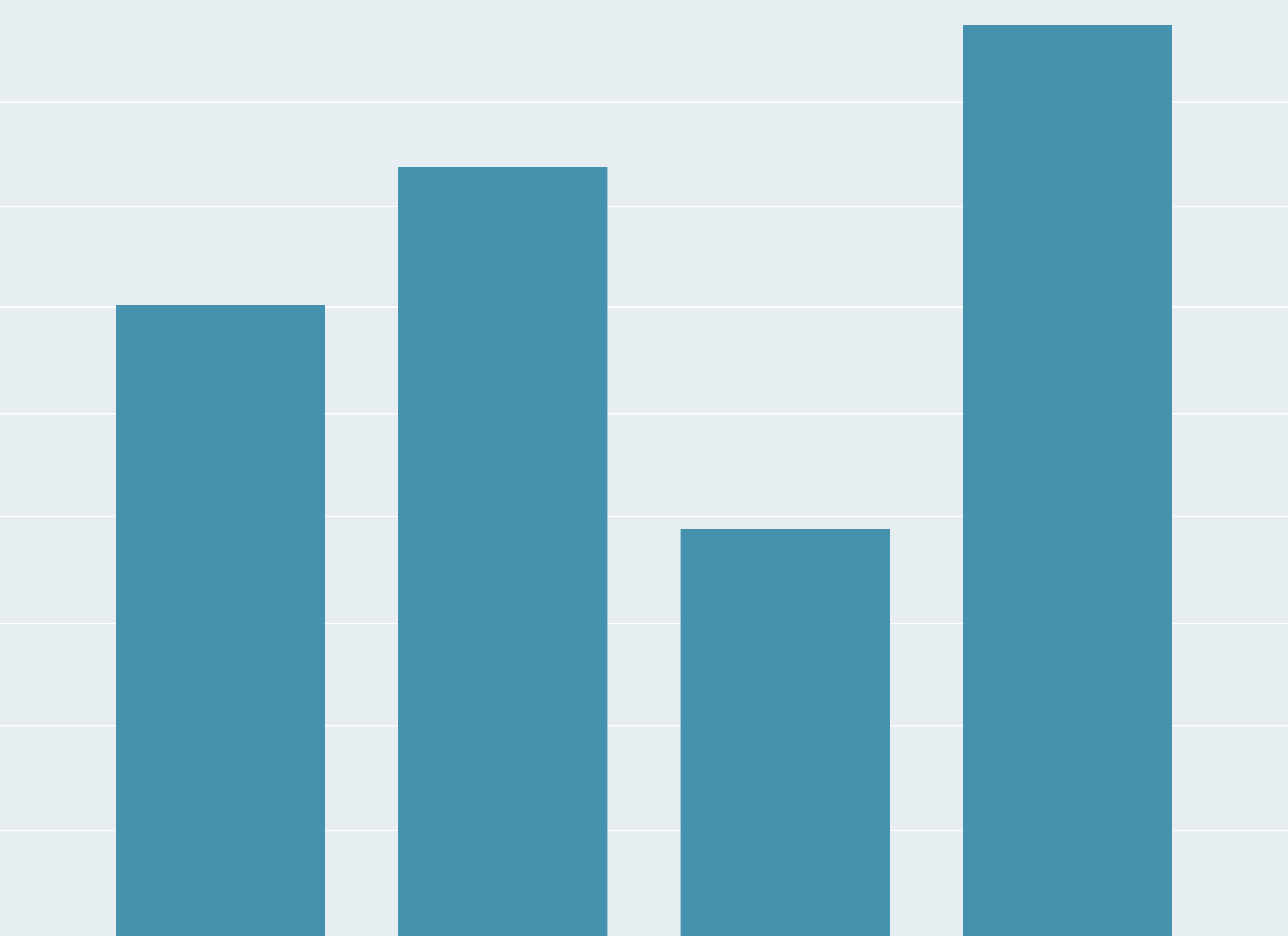}%
\label{fig:lcg_magnetic_analysis_k04_s4_a}%
}
\hspace{5mm}%
\subfigure[]{%
\includegraphics[clip,trim=4mm 0mm 4mm 0,height=0.15\textwidth]{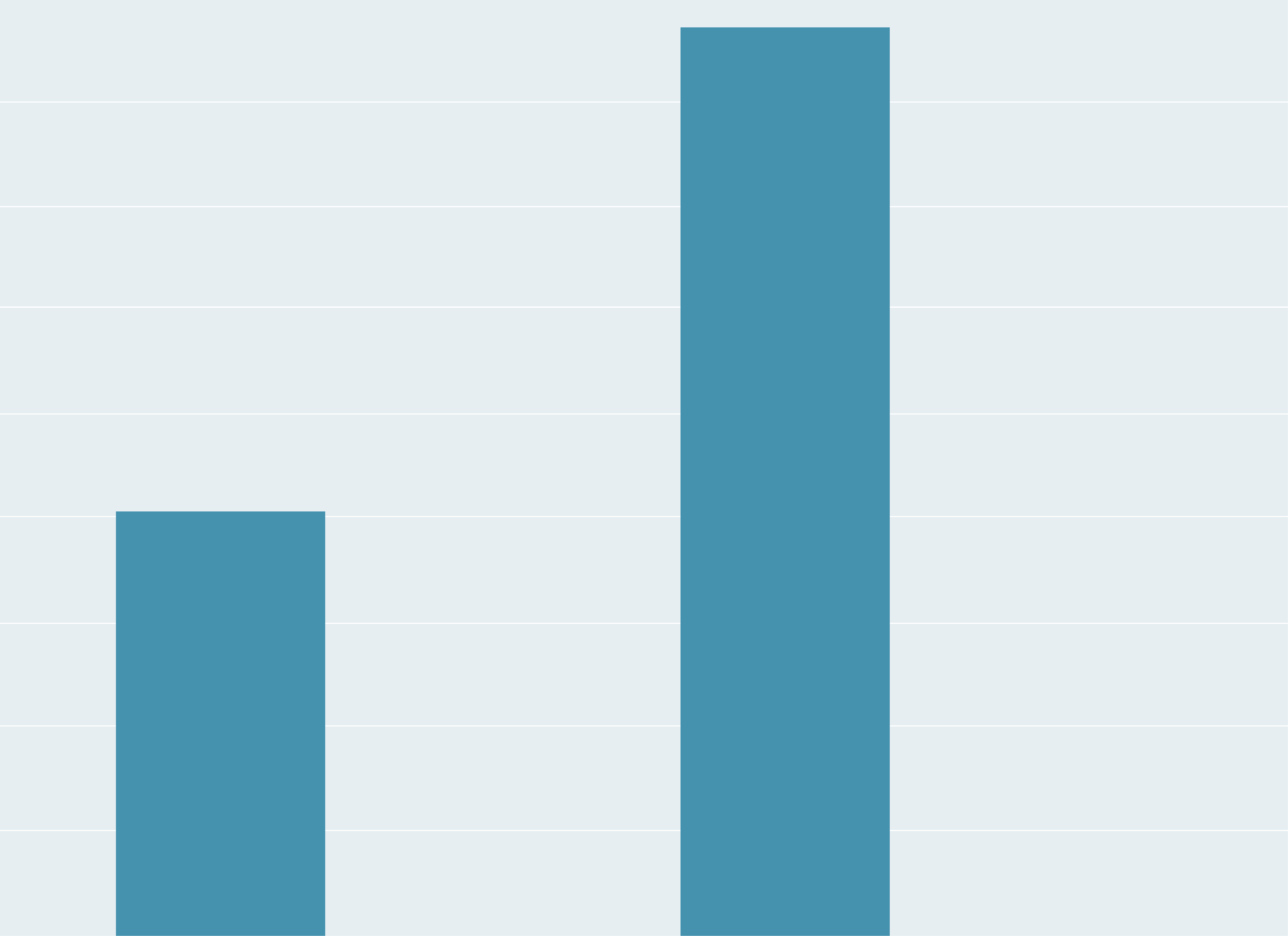}%
\label{fig:lcg_magnetic_analysis_k04_s4_b}%
}
\hspace{5mm}%
\subfigure[]{%
\includegraphics[clip,trim=4mm 0mm 4mm 0,height=0.15\textwidth]{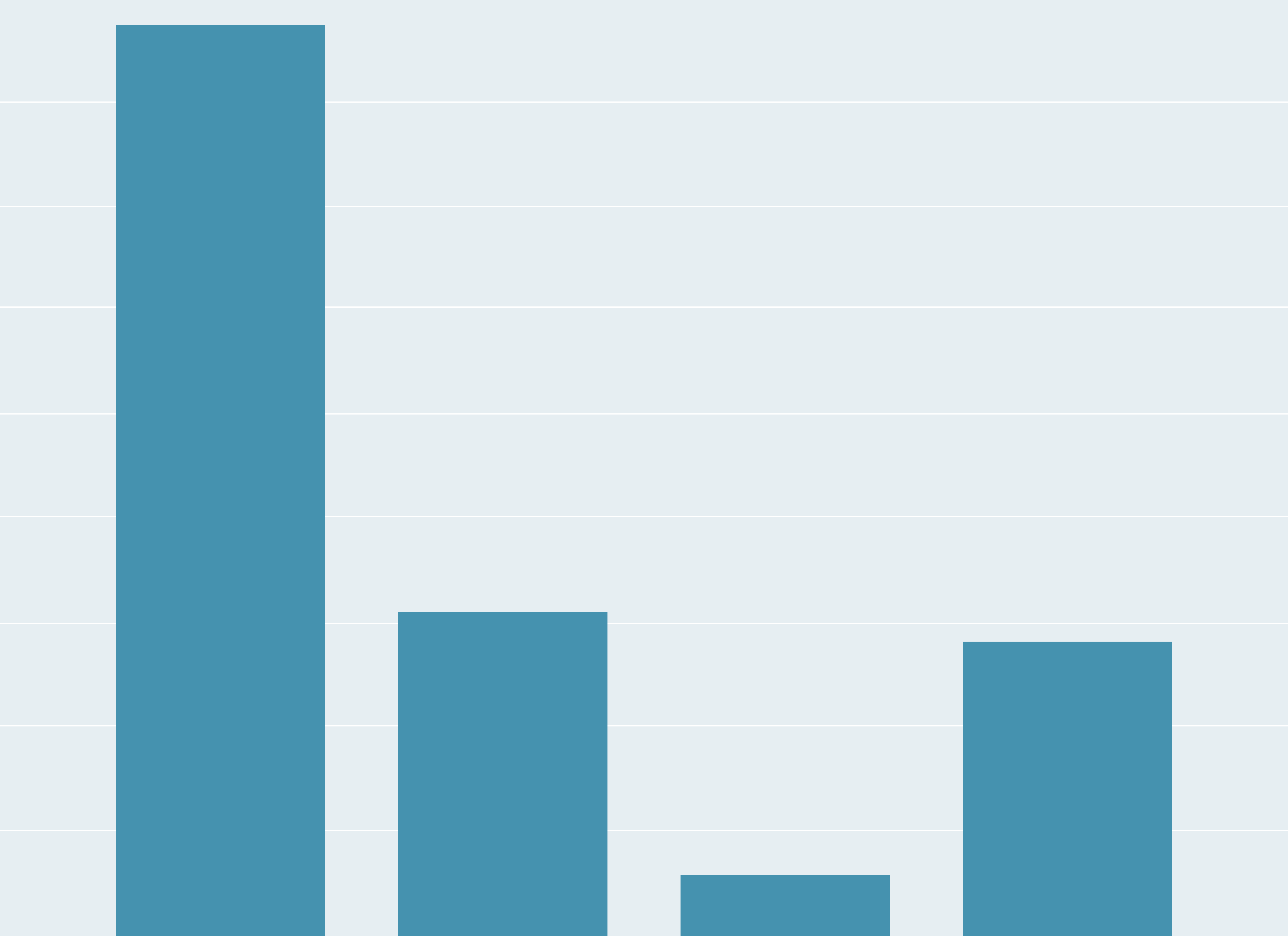}%
\label{fig:lcg_magnetic_analysis_k04_s4_c}%
}
\hfill%
\\
\vspace{-7pt}
\subfigure[]{%
\includegraphics[clip,trim=130mm 20mm 140mm 4mm, height=0.2\textwidth]{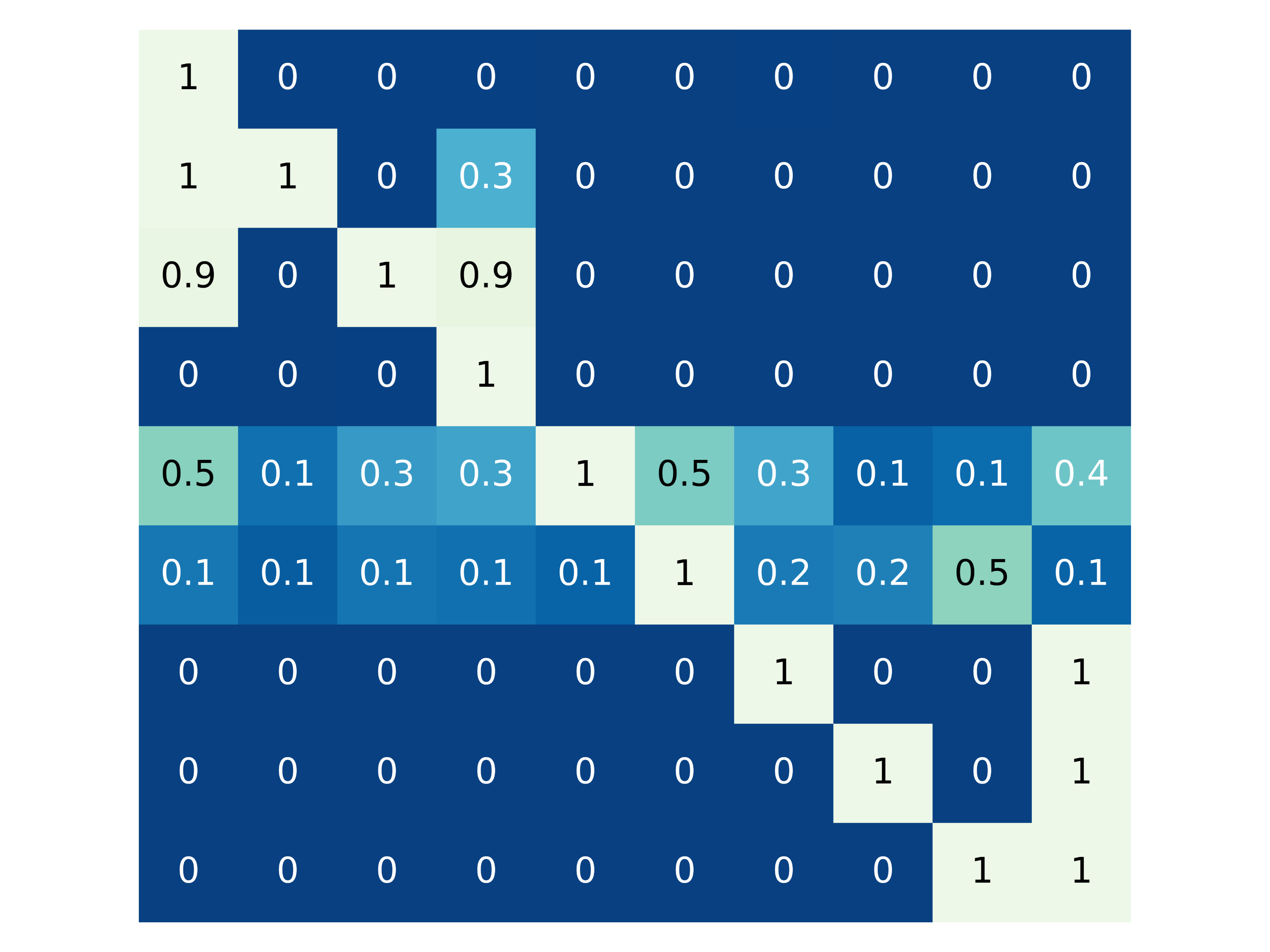}%
\label{fig:lcg_magnetic_analysis_k04_s4_d}%
}
\hfill
\subfigure[]{%
\includegraphics[clip,trim=130mm 20mm 140mm 4mm, height=0.2\textwidth]{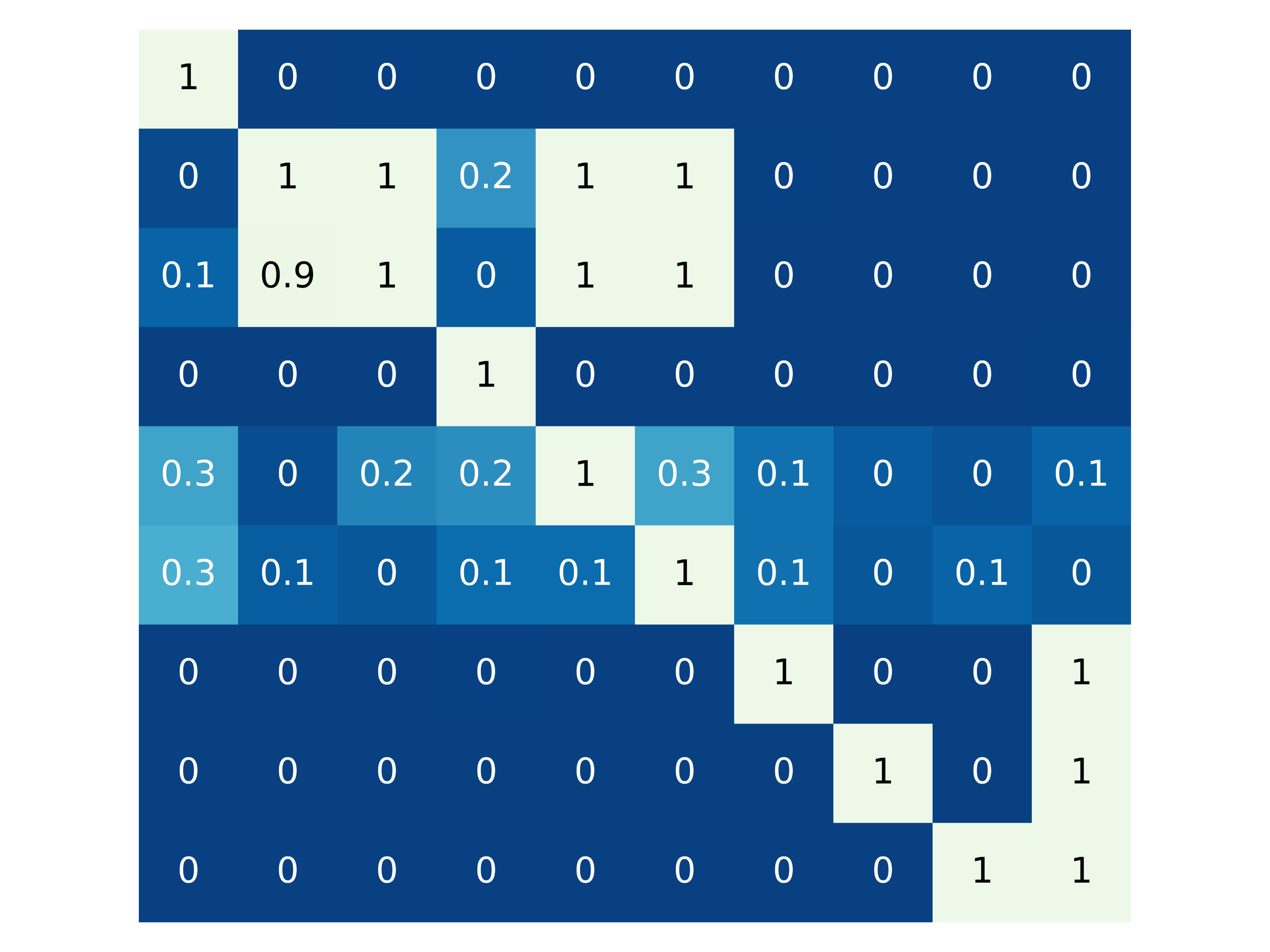}%
\label{fig:lcg_magnetic_analysis_k04_s4_e}%
}
\hfill
\subfigure[]{%
\includegraphics[clip,trim=130mm 20mm 140mm 4mm, height=0.2\textwidth]{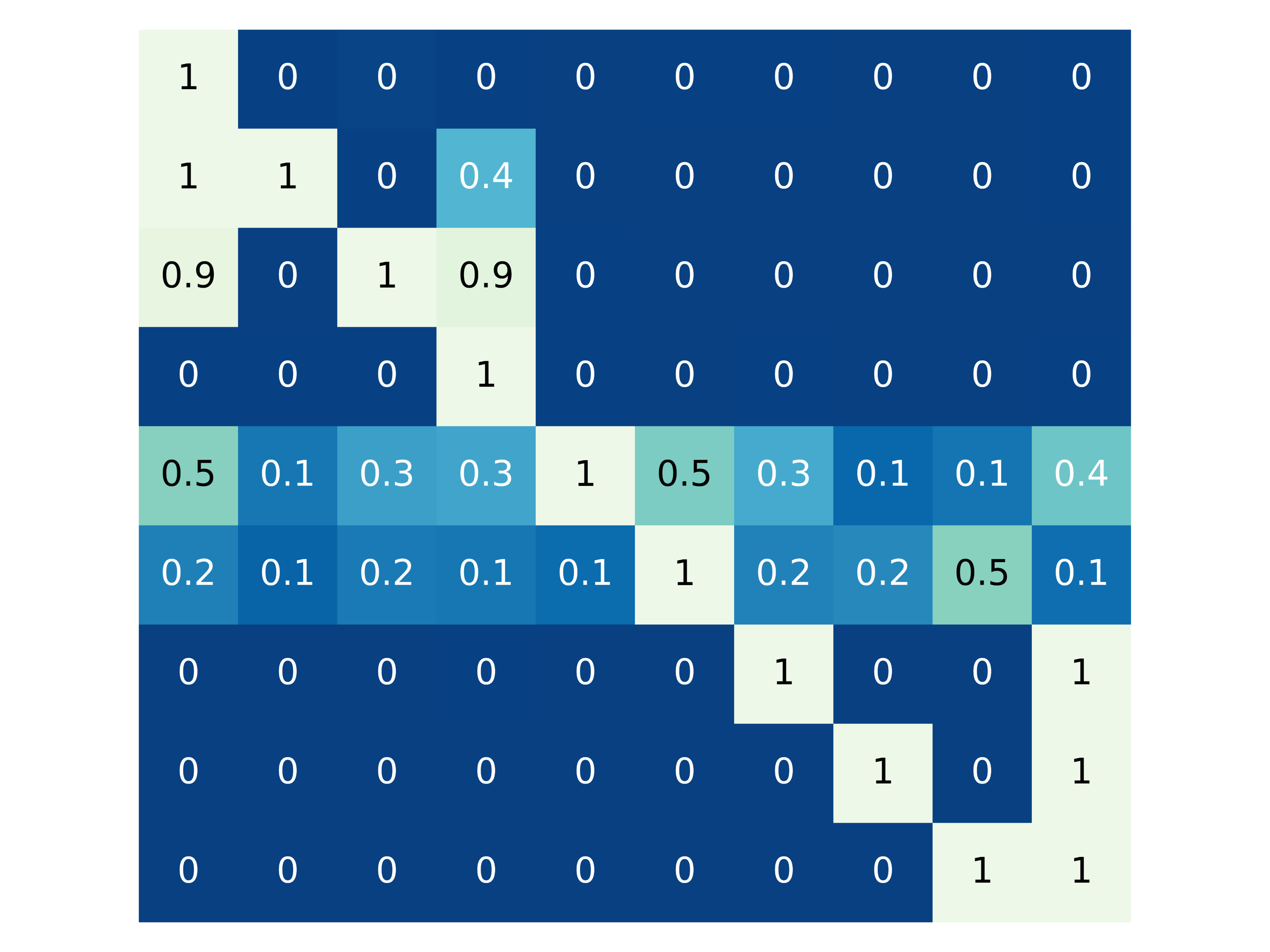}%
\label{fig:lcg_magnetic_analysis_k04_s4_f}%
}
\hfill
\subfigure[]{%
\includegraphics[clip,trim=40mm 20mm 70mm 10mm, height=0.2\textwidth]{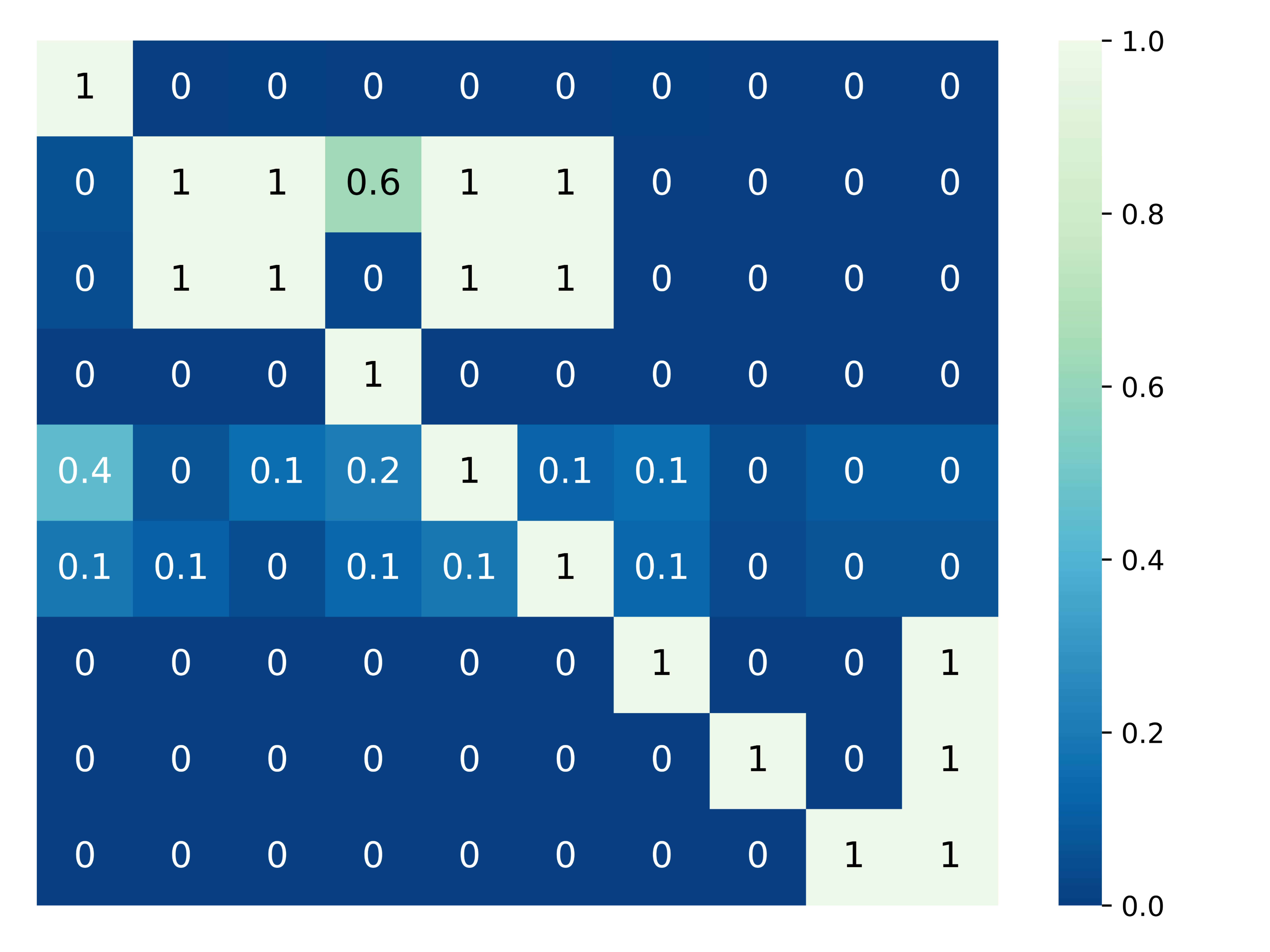}%
\label{fig:lcg_magnetic_analysis_k04_s4_g}%
}
\vspace{-7pt}
\caption{Another sample run of our method with quantization degree of 4 in Magnetic.
}
\vspace{-2pt}
\label{fig:lcg_magnetic_analysis_k04_s4}
\end{figure*}

%% file: figure/lcg_magnetic_analysis_k16_s4.tex
\begin{figure*}[t!]
\centering
\subfigure[]{%
\includegraphics[clip,trim=130mm 20mm 140mm 4mm, height=0.2\textwidth]{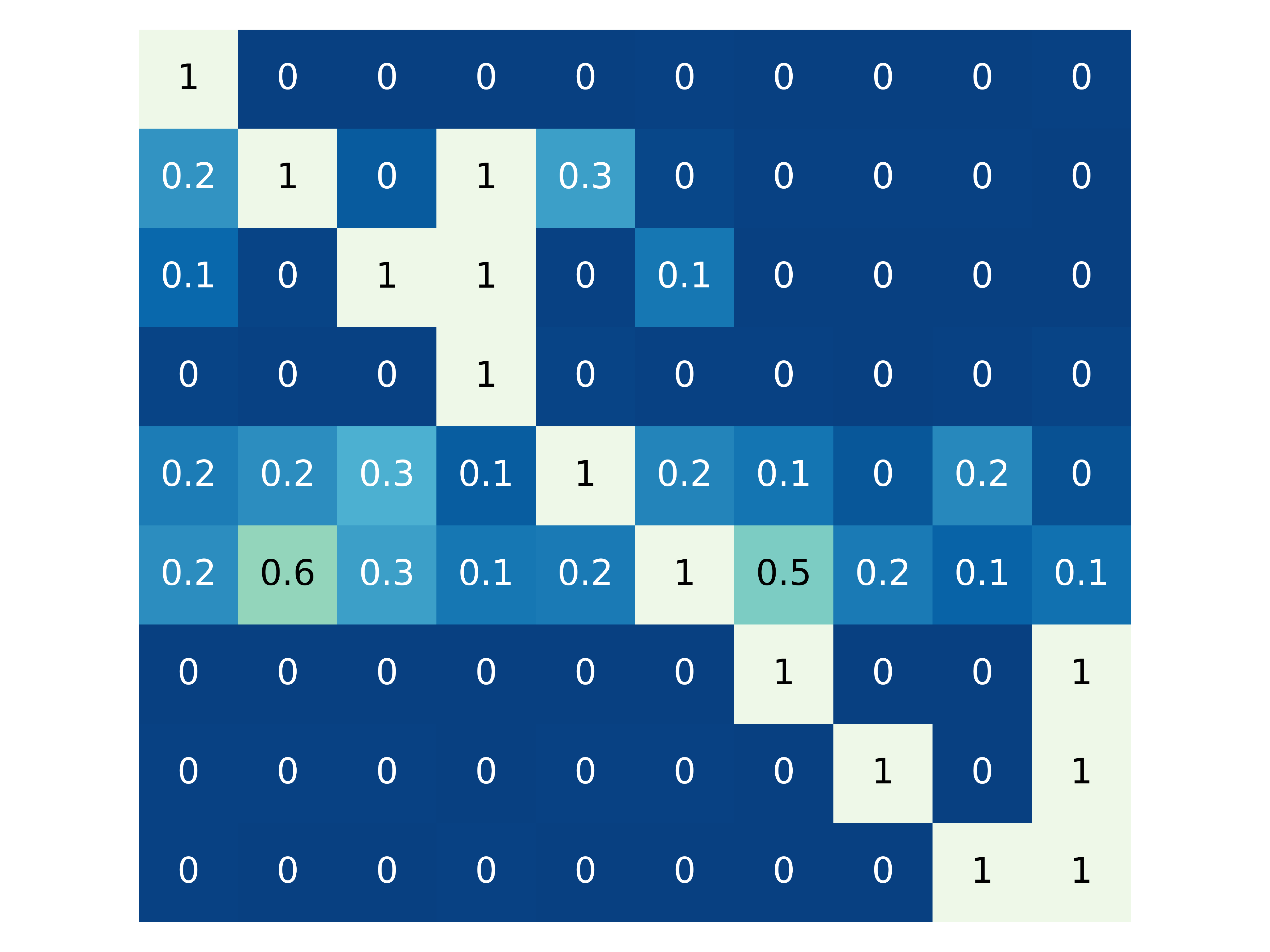}%
\label{fig:lcg_magnetic_analysis_k16_s4_a}%
}
\hspace{5mm}%
\subfigure[]{%
\includegraphics[clip,trim=130mm 20mm 140mm 4mm, height=0.2\textwidth]{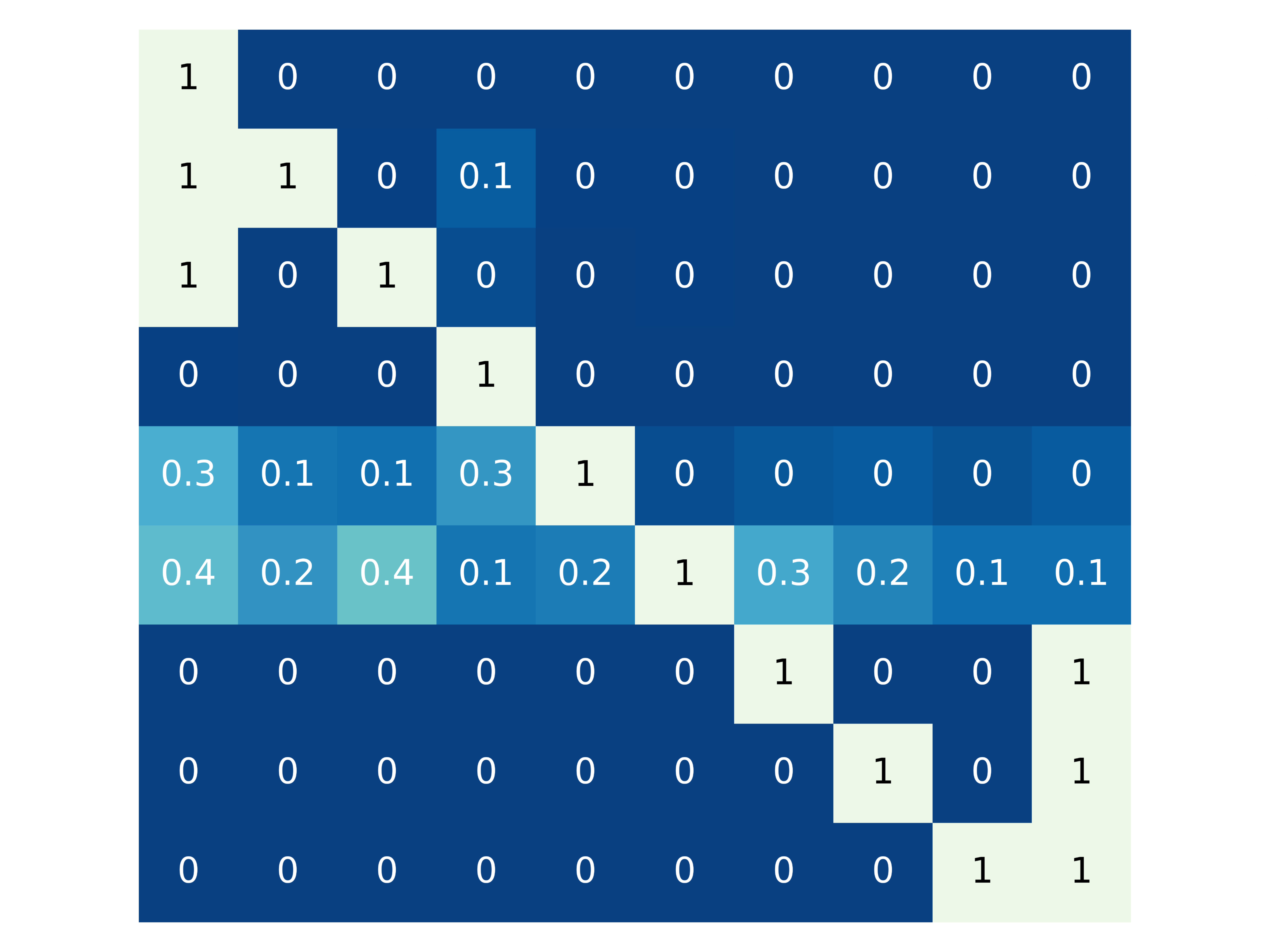}%
\label{fig:lcg_magnetic_analysis_k16_s4_b}%
}
\vspace{-3pt}
\caption{More fine-grained LCGs learned by our method with quantization degree of 16 in Magnetic.}
\vspace{-2pt}
\label{fig:lcg_magnetic_analysis_k16_s4}
\end{figure*}

%% file: figure/lcg_chemical_analysis_k02_s1.tex
\begin{figure*}[t!]
\centering
\subfigure[]{%
\includegraphics[clip,trim=120mm 0mm 120mm 0, height=0.15\textwidth]{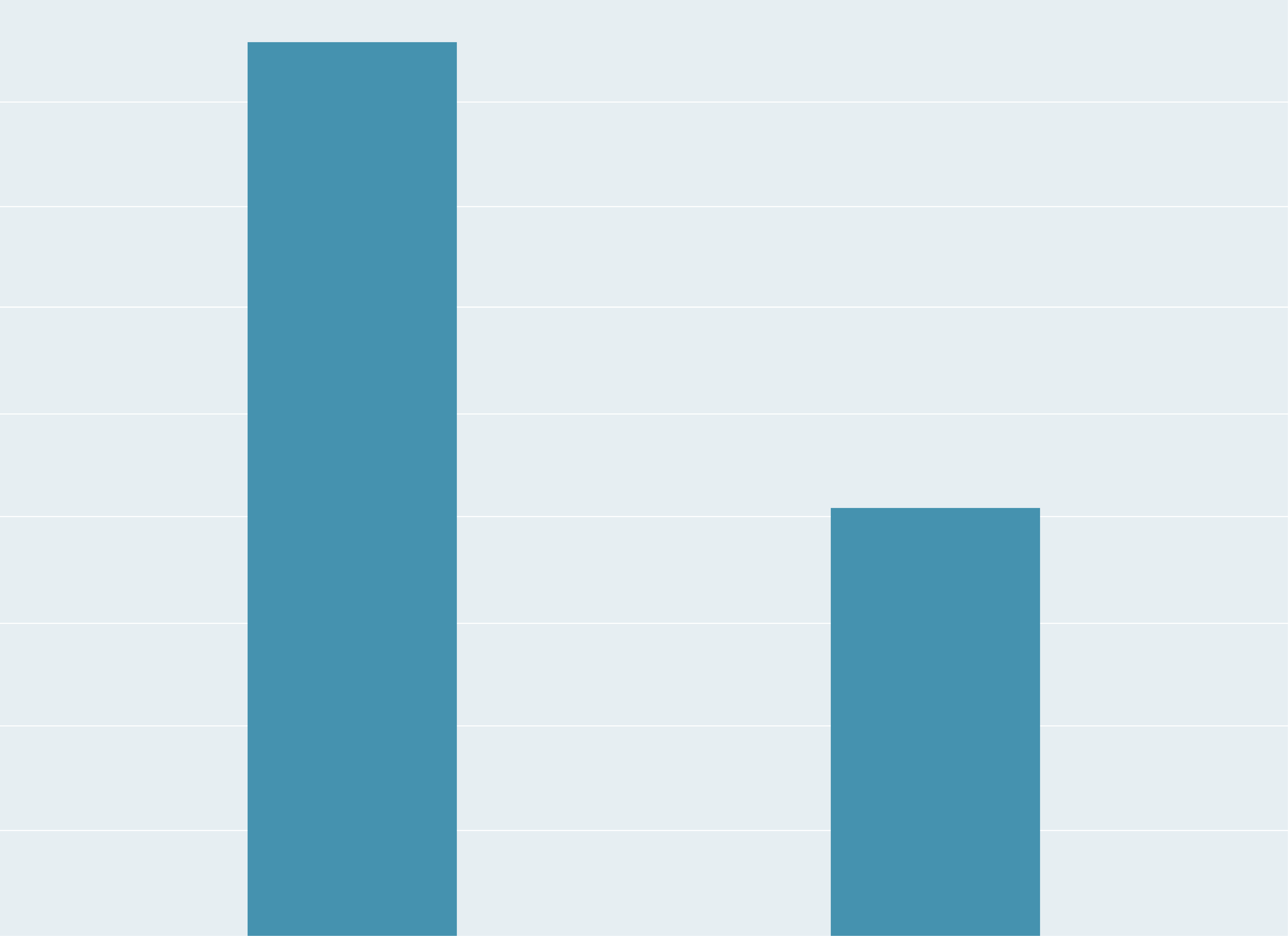}%
\label{fig:lcg_chemical_analysis_k02_s1_a}%
}
\hfill
\subfigure[]{%
\includegraphics[clip,trim=120mm 0mm 120mm 0,height=0.15\textwidth]{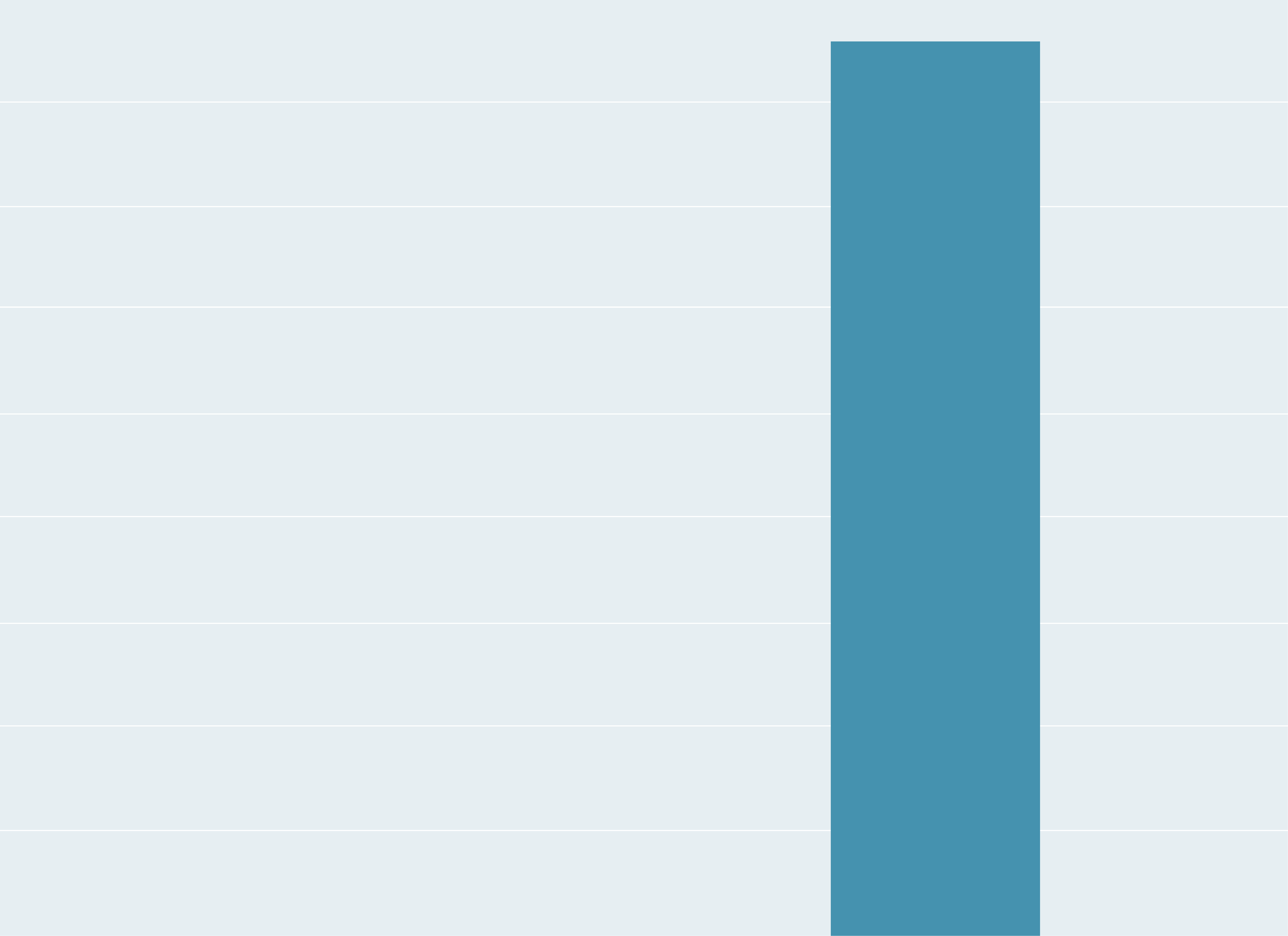}%
\label{fig:lcg_chemical_analysis_k02_s1_b}%
}
\hfill
\subfigure[]{%
\includegraphics[clip,trim=120mm 0mm 120mm 0,height=0.15\textwidth]{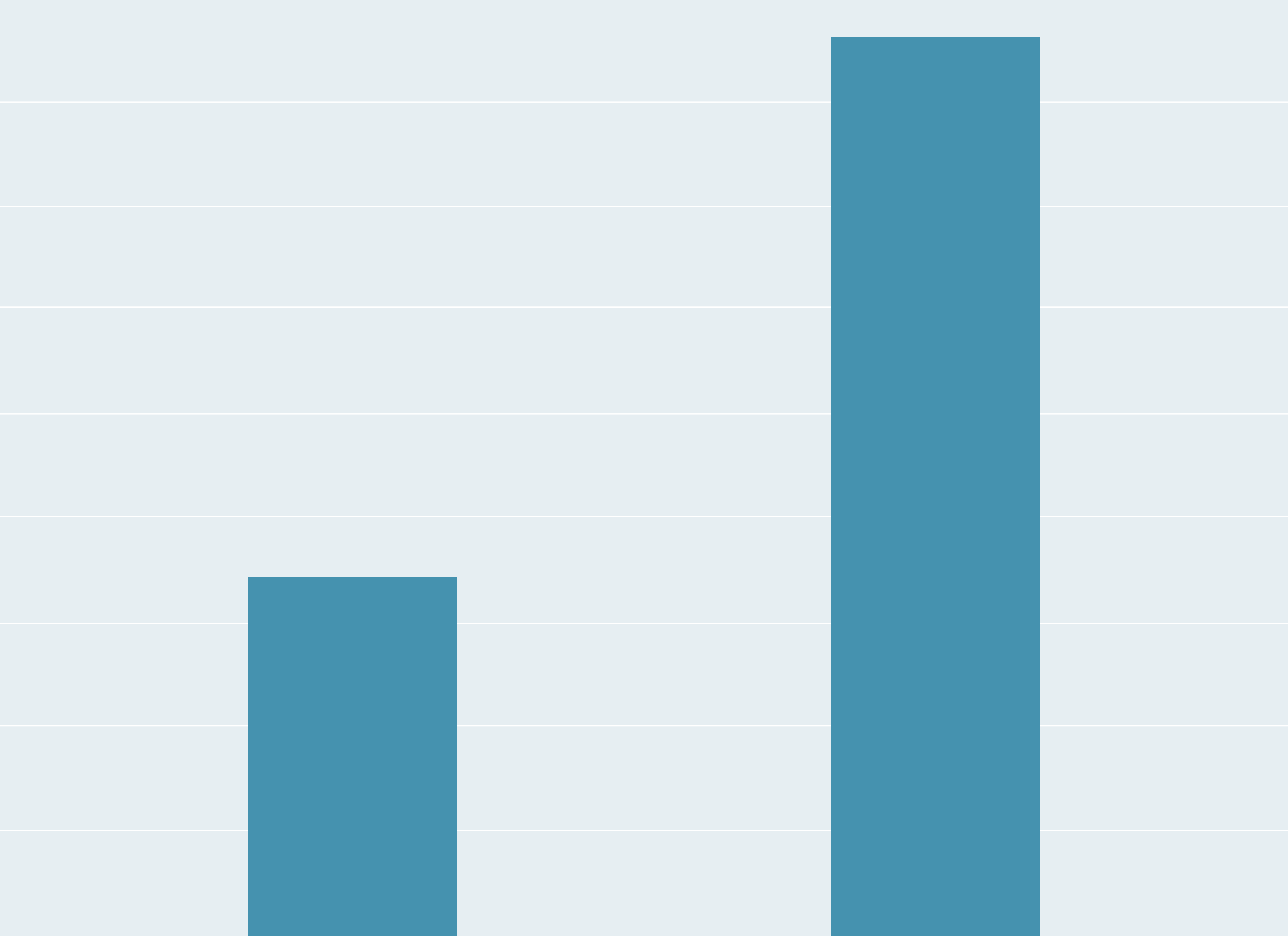}%
\label{fig:lcg_chemical_analysis_k02_s1_c}%
}
\hfill%
\subfigure[]{%
\includegraphics[clip,trim=130mm 20mm 140mm 4mm, height=0.2\textwidth]{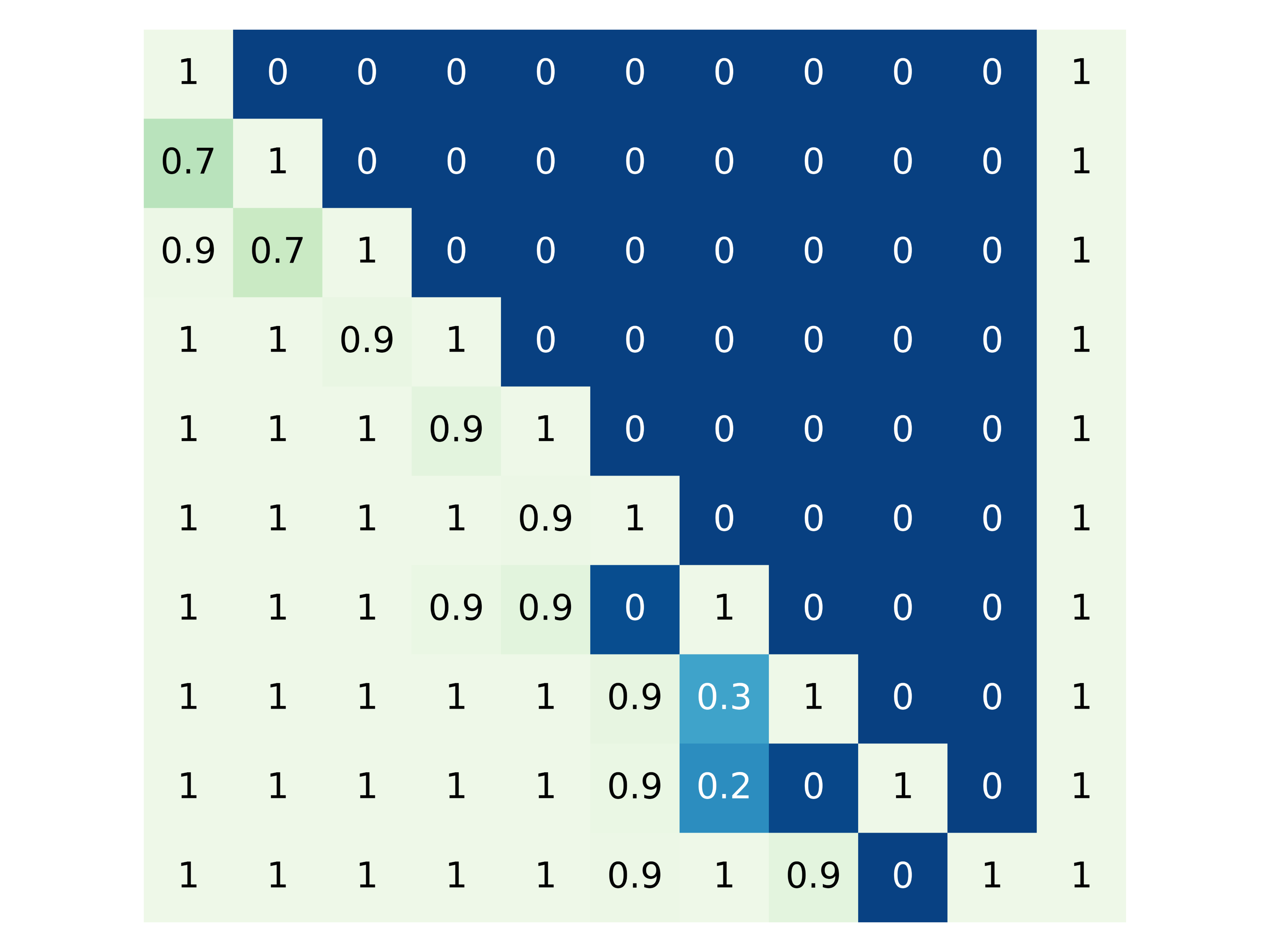}%
\label{fig:lcg_chemical_analysis_k02_s1_d}%
}
\hfill
\subfigure[]{%
\includegraphics[clip,trim=40mm 20mm 70mm 10mm, height=0.2\textwidth]{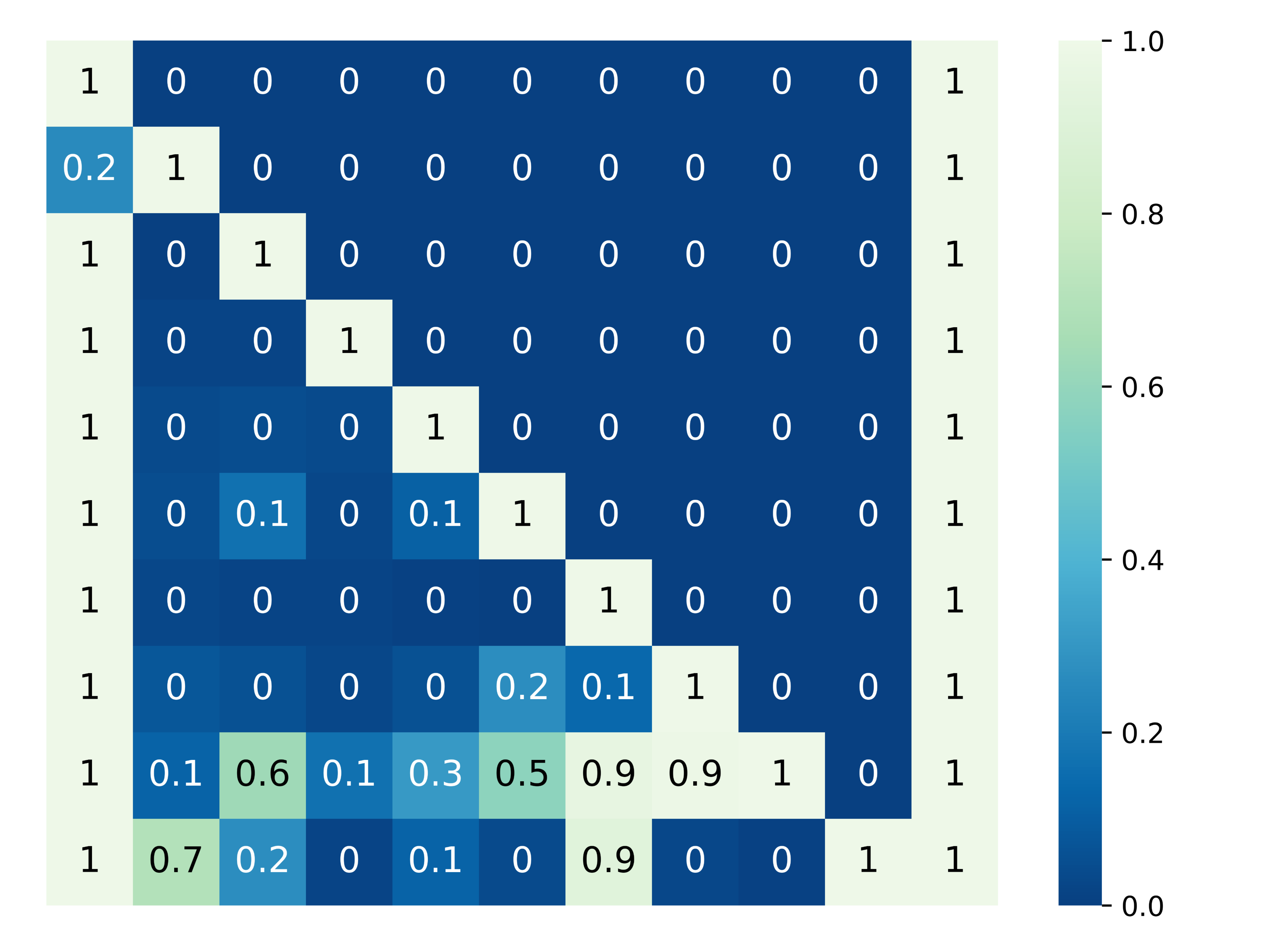}%
\label{fig:lcg_chemical_analysis_k02_s1_e}%
}
\vspace{-7pt}
\caption{Analysis of LCGs learned by our method with quantization degree of 2 in Chemical (\textit{full-fork}).
}
\vspace{-2pt}
\label{fig:lcg_chemical_analysis_k02_s1}
\end{figure*}

%% file: figure/lcg_chemical_analysis_k02_s2.tex
\begin{figure*}[t!]
\centering
\subfigure[]{%
\includegraphics[clip,trim=120mm 0mm 120mm 0, height=0.15\textwidth]{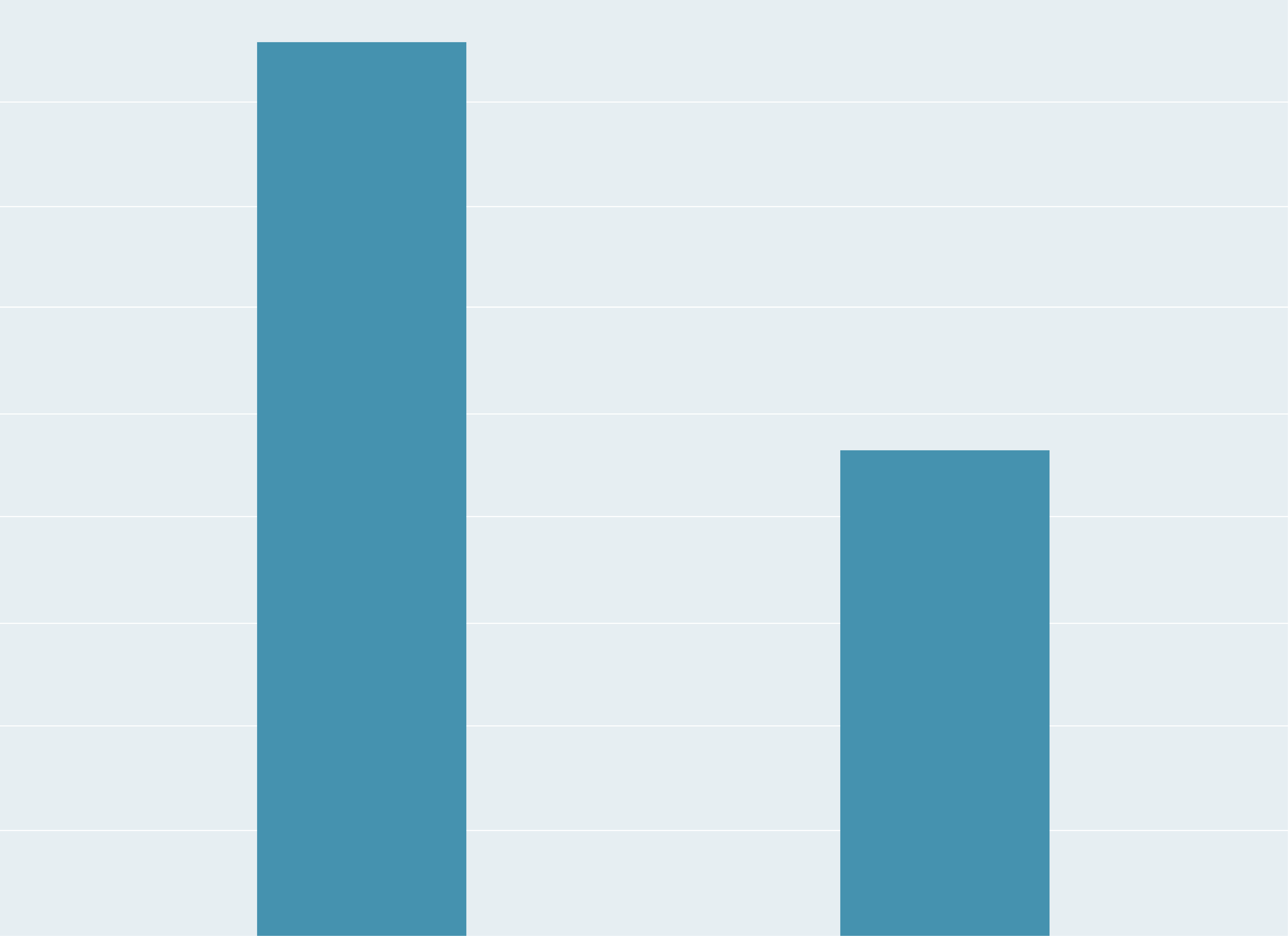}%
\label{fig:lcg_chemical_analysis_k02_s2_a}%
}
\hfill
\subfigure[]{%
\includegraphics[clip,trim=120mm 0mm 120mm 0,height=0.15\textwidth]{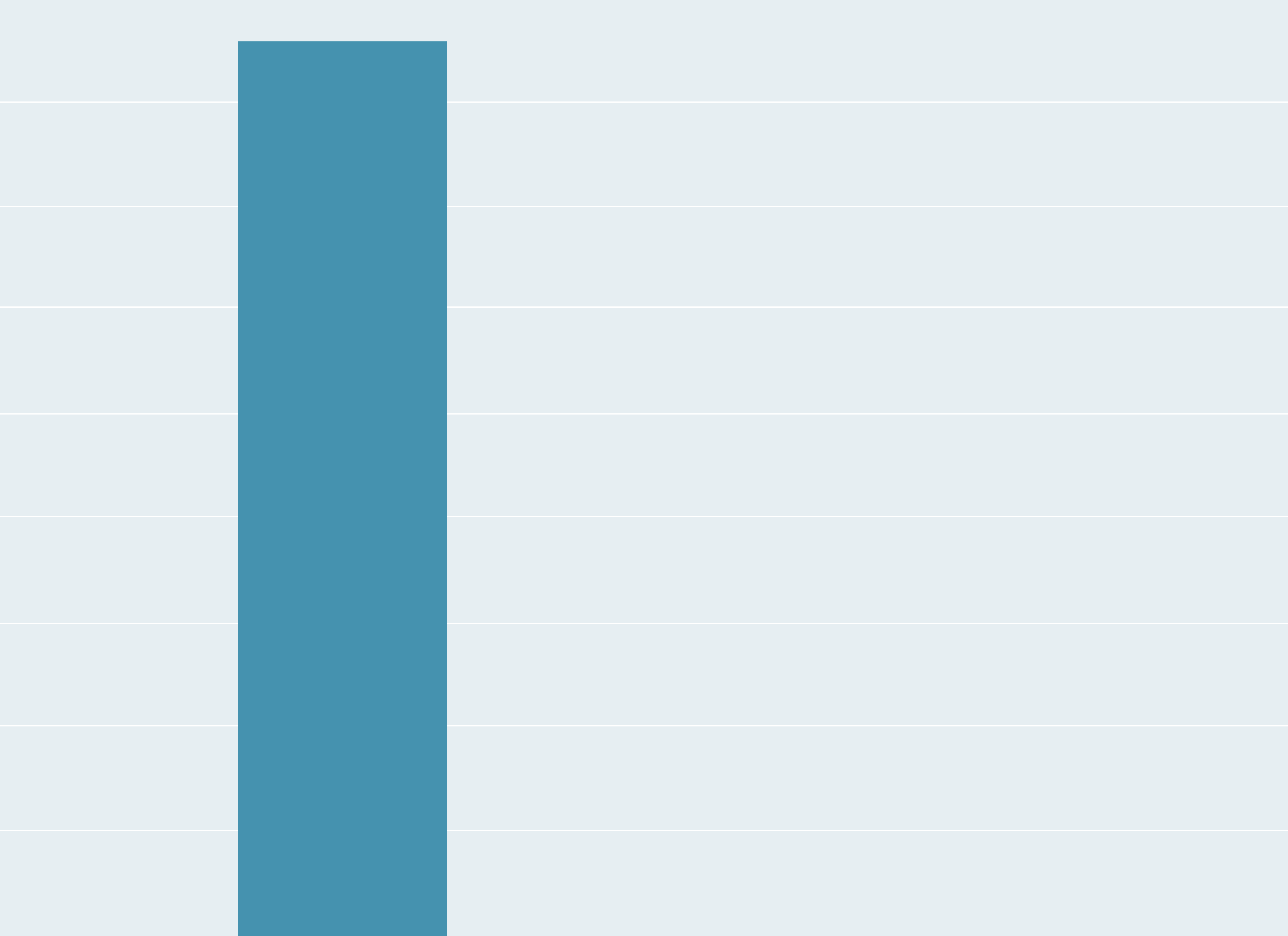}%
\label{fig:lcg_chemical_analysis_k02_s2_b}%
}
\hfill
\subfigure[]{%
\includegraphics[clip,trim=120mm 0mm 120mm 0,height=0.15\textwidth]{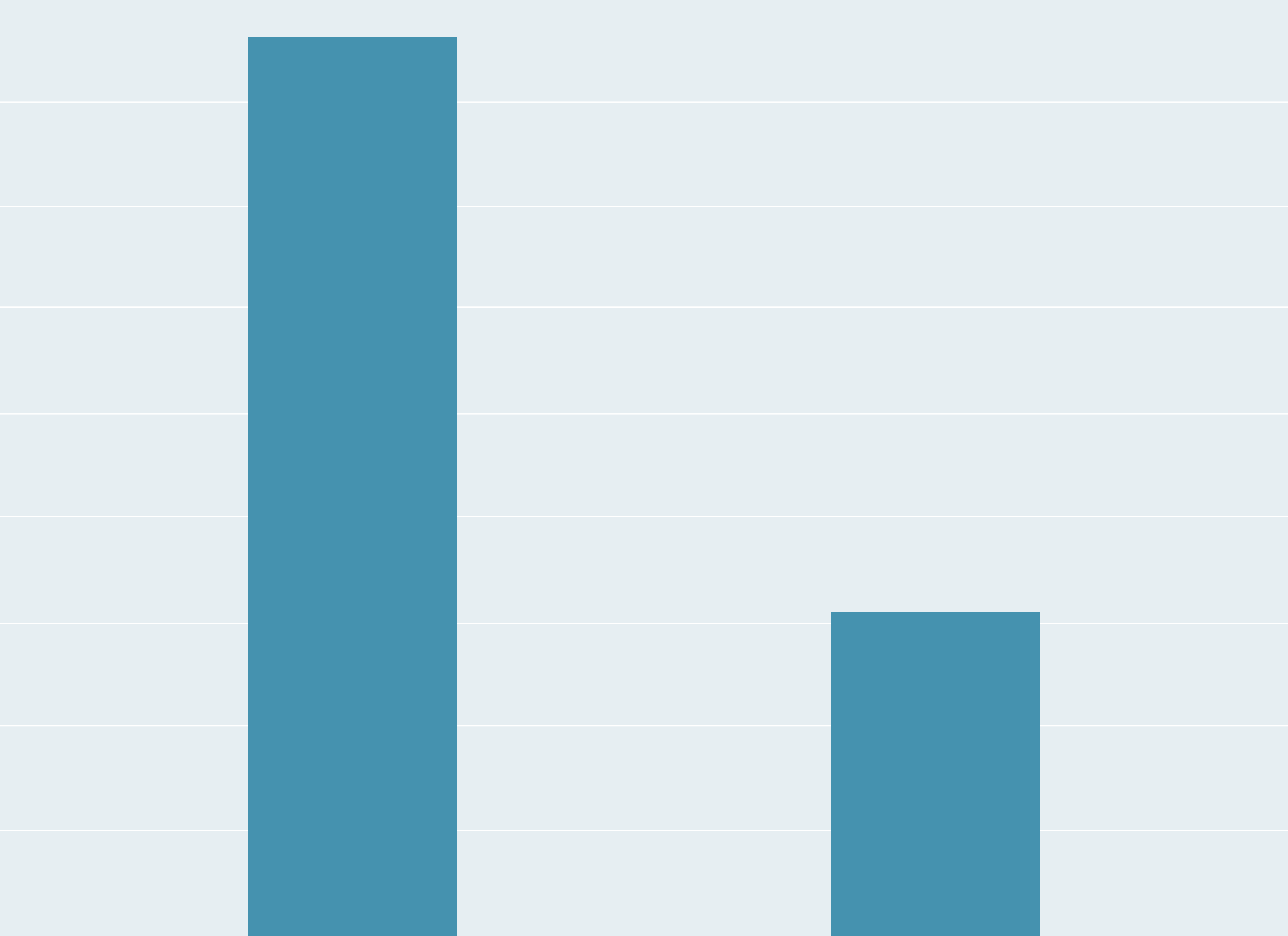}%
\label{fig:lcg_chemical_analysis_k02_s2_c}%
}
\hfill%
\subfigure[]{%
\includegraphics[clip,trim=130mm 20mm 140mm 4mm, height=0.2\textwidth]{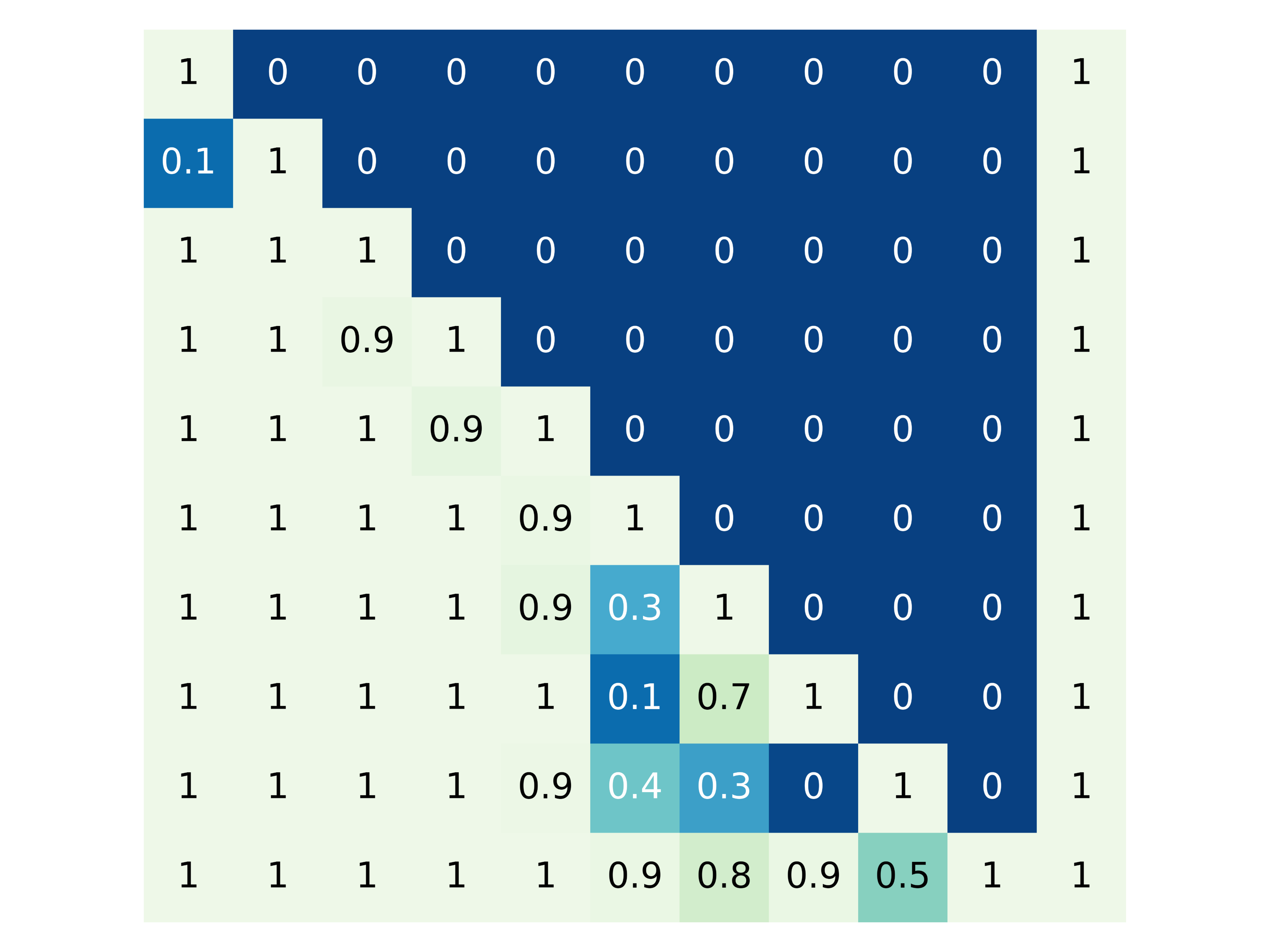}%
\label{fig:lcg_chemical_analysis_k02_s2_d}%
}
\hfill
\subfigure[]{%
\includegraphics[clip,trim=40mm 20mm 70mm 10mm, height=0.2\textwidth]{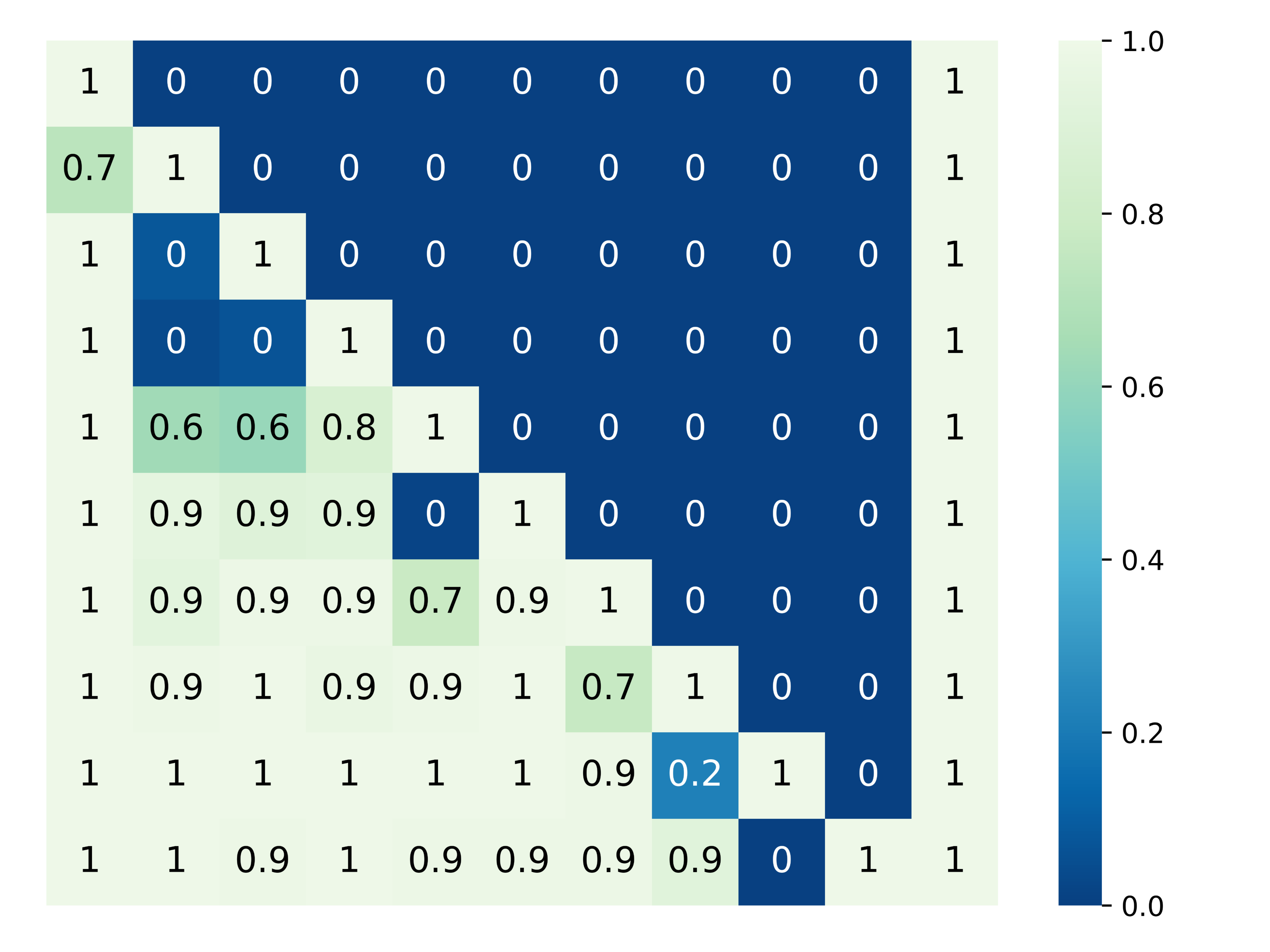}%
\label{fig:lcg_chemical_analysis_k02_s2_e}%
}
\vspace{-7pt}
\caption{{Failure case} of our method with quantization degree of 2 in Chemical (\textit{full-fork}).}
\vspace{-2pt}
\label{fig:lcg_chemical_analysis_k02_s2}
\end{figure*}

%% file: table/param_model.tex
\begin{table}[t!]
\caption{Parameters of each model.}
\label{table:param_model}
\centering
\begin{adjustbox}{width=0.57\textwidth}%
\begin{tabular}{@{}lcccc@{}}
\toprule
&
& \multicolumn{2}{c}{\makecell{Chemical}} 
& \multirow{2}{*}{\makecell{Magnetic}}
\\

\cmidrule(lr){3-4}
Models
& Parameters
& \makecell[c]{\textit{full-fork}}
& \makecell[c]{\textit{full-chain}}
&
\\
\midrule						
\multirow{2}{*}{\makecell{MLP}}	&	Hidden dim	& 1024	& 1024	& 512	\\
	&	Hidden layers	& 3	& 3	& 4	\\
\midrule						
\multirow{2}{*}{\makecell{Modular}}	&	Hidden dim	& 128	& 128	& 128	\\
	&	Hidden layers	& 4	& 4	& 4	\\
\midrule						
\multirow{6}{*}{\makecell{GNN}}	&	Node attribute dim	& 256	& 256	& 256	\\
	&	Node network hidden dim	& 512	& 512	& 512	\\
	&	Node network hidden layers	& 3	& 3	& 3	\\
	&	Edge attribute dim	& 256	& 256	& 256	\\
	&	Edge network hidden dim	& 512	& 512	& 512	\\
	&	Edge network hidden layers	& 3	& 3	& 3	\\
\midrule						
\multirow{8}{*}{\makecell{NPS}}	&	Number of rules	& 20	& 20	& 15	\\
	&	Cond selector dim	& 128	& 128	& 128	\\
	&	Rule embedding dim	& 128	& 128	& 128	\\
	&	Rule selector dim	& 128	& 128	& 128	\\
	&	Feature encoder hidden dim	& 128	& 128	& 128	\\
	&	Feature encoder hidden layers	& 2	& 2	& 2	\\
	&	Rule network hidden dim	& 128	& 128	& 128	\\
	&	Rule network hidden layers	& 3	& 3	& 3	\\
\midrule						
\multirow{8}{*}{\makecell{CDL}}	&	Hidden dim	& 128	& 128	& 128	\\
	&	Hidden layers	& 4	& 4	& 4	\\
	&	CMI threshold	& 0.001	& 0.001	& 0.001	\\
	&	CMI optimization frequency	& 10	& 10	& 10	\\
	&	CMI evaluation frequency	& 10	& 10	& 10	\\
	&	CMI evaluation step size	& 1	& 1	& 1	\\
	&	CMI evaluation batch size	& 256	& 256	& 256	\\
	&	EMA discount	& 0.9	& 0.9	& 0.99	\\
\midrule						
\multirow{3}{*}{\makecell{Grader}}	&	Feature embedding dim	& 128	& 128	& N/A	\\
	&	GRU hidden dim	& 128	& 128	& N/A	\\
	&	Causal discovery frequency	& 10	& 10	& N/A	\\
\midrule						
\multirow{2}{*}{\makecell{Oracle}}	&	Hidden dim	& 128	& 128	& 128	\\
	&	Hidden layers	& 4	& 4	& 5	\\
\midrule						
\multirow{4}{*}{\makecell{NCD}}	&	Hidden dim	& 128	& 128	& 128	\\
	&	Hidden layers	& 4	& 4	& 5	\\
	&	Auxiliary network hidden dim	& 128	& 128	& 128	\\
	&	Auxiliary network hidden layers	& 2	& 2	& 2	\\
\midrule						
\multirow{6}{*}{\makecell{Ours}}	&	Hidden dim	& 128	& 128	& 128	\\
	&	Hidden layers	& 4	& 4	& 5	\\
	&	VQ encoder	& [128, 64]	& [128, 64]	& [128, 64]	\\
	&	VQ decoder	& [32]	& [32]	& [32]	\\
	&	Codebook size	& 16	& 16	& 16	\\
	&	Code dimension	& 16	& 16	& 16	\\
\bottomrule
\end{tabular}
\end{adjustbox}
\end{table}

%% file: figure/curve_train.tex
\begin{figure*}[t]
\centering
\includegraphics[clip,trim=0 0mm 0 0,width=0.8\textwidth]{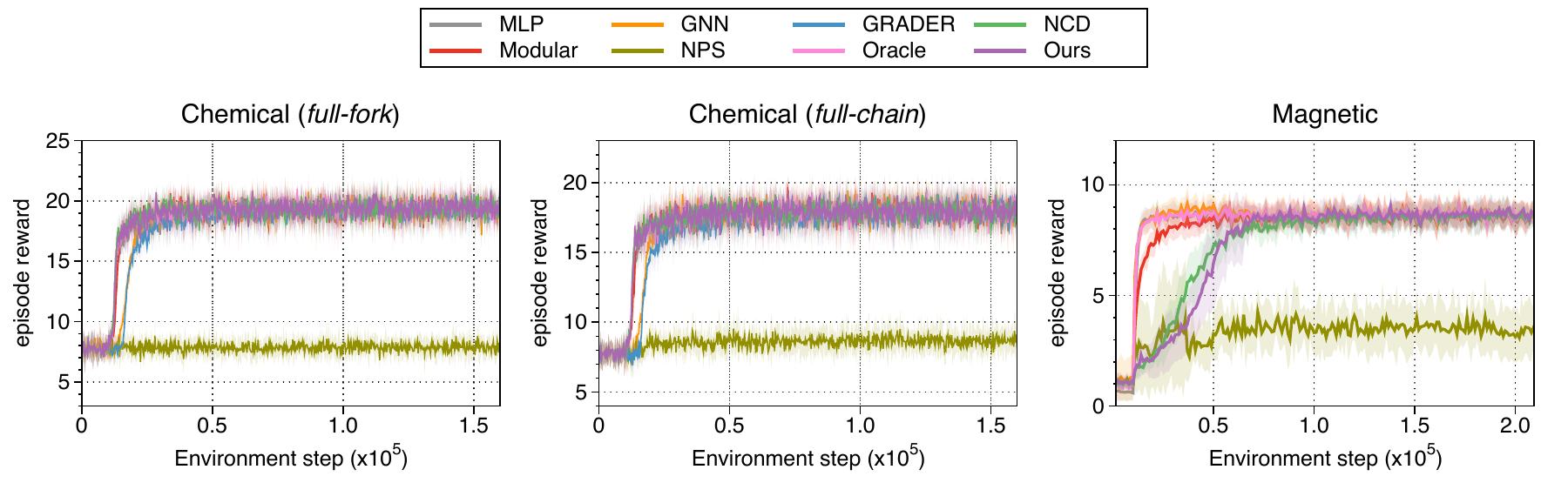}
\vspace{-3mm}
\caption{Learning curves during training as measured by the episode reward.}
\vspace{-2mm}
\label{fig:learning_curve_train}
\end{figure*}

%% file: figure/curve_chemical_fork.tex
\begin{figure*}[t]
\centering
\includegraphics[clip,trim=0 0mm 0 0mm,width=0.8\textwidth]{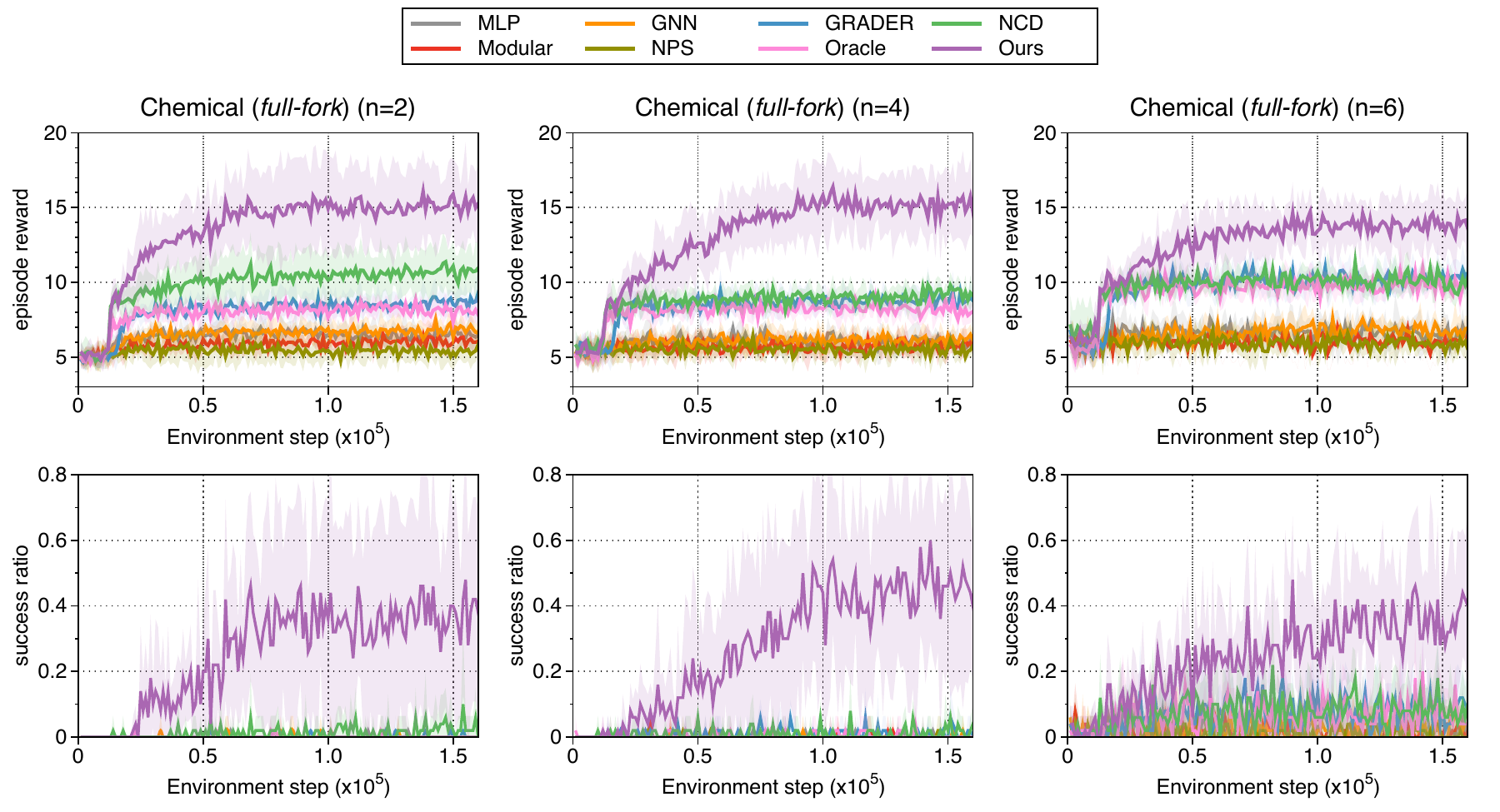}
\vspace{-3mm}
\caption{Learning curves on downstream tasks in Chemical (\textit{full-fork}) as measured on the episode reward (\textbf{top}) and success rate (\textbf{bottom}).}
\vspace{-2mm}
\label{fig:learning_curve_fork}
\end{figure*}

%% file: figure/curve_chemical_chain.tex
\begin{figure*}[t]
\centering
\includegraphics[clip,trim=0 0mm 0 0mm,width=0.8\textwidth]{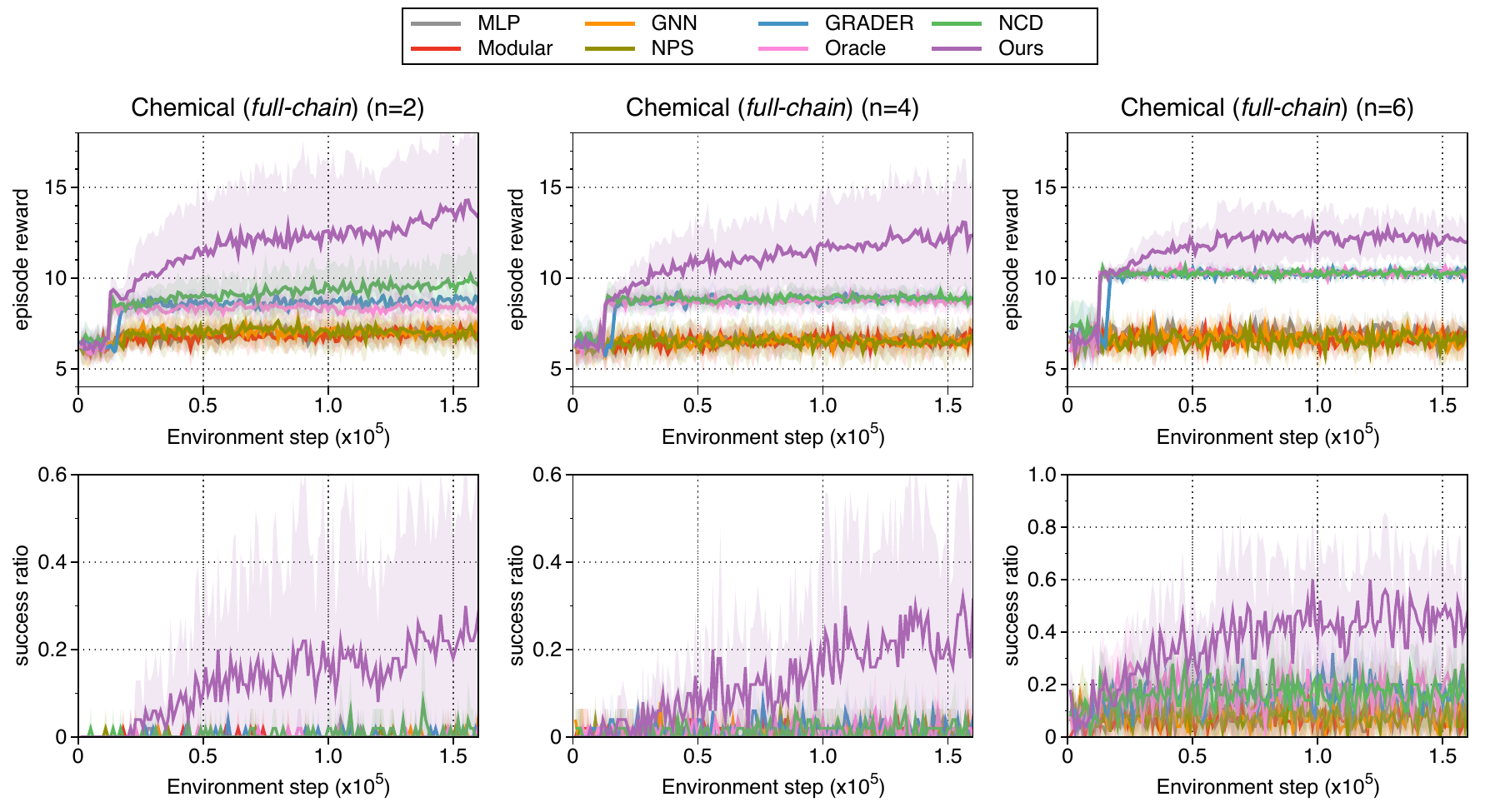}
\vspace{-3mm}
\caption{Learning curves on downstream tasks in Chemical (\textit{full-chain}) as measured on the episode reward (\textbf{top}) and success rate (\textbf{bottom}).
}
\label{fig:learning_curve_chain}
\end{figure*}

%% file: main.bbl
\begin{thebibliography}{}

\end{thebibliography}


\begin{thebibliography}{80}
\providecommand{\natexlab}[1]{#1}
\providecommand{\url}[1]{\texttt{#1}}
\expandafter\ifx\csname urlstyle\endcsname\relax
  \providecommand{\doi}[1]{doi: #1}\else
  \providecommand{\doi}{doi: \begingroup \urlstyle{rm}\Url}\fi

\bibitem[Acid \& de~Campos(2003)Acid and de~Campos]{acid2003searching}
Acid, S. and de~Campos, L.~M.
\newblock Searching for bayesian network structures in the space of restricted acyclic partially directed graphs.
\newblock \emph{Journal of Artificial Intelligence Research}, 18:\penalty0 445--490, 2003.

\bibitem[Barash \& Friedman(2001)Barash and Friedman]{barash2001context}
Barash, Y. and Friedman, N.
\newblock Context-specific bayesian clustering for gene expression data.
\newblock In \emph{Proceedings of the fifth annual international conference on Computational biology}, pp.\  12--21, 2001.

\bibitem[Bareinboim et~al.(2015)Bareinboim, Forney, and Pearl]{bareinboim2015bandits}
Bareinboim, E., Forney, A., and Pearl, J.
\newblock Bandits with unobserved confounders: A causal approach.
\newblock \emph{Advances in Neural Information Processing Systems}, 28, 2015.

\bibitem[Bica et~al.(2021)Bica, Jarrett, and van~der Schaar]{bica2021invariant}
Bica, I., Jarrett, D., and van~der Schaar, M.
\newblock Invariant causal imitation learning for generalizable policies.
\newblock \emph{Advances in Neural Information Processing Systems}, 34:\penalty0 3952--3964, 2021.

\bibitem[Bongers et~al.(2018)Bongers, Blom, and Mooij]{bongers2018causal}
Bongers, S., Blom, T., and Mooij, J.~M.
\newblock Causal modeling of dynamical systems.
\newblock \emph{arXiv preprint arXiv:1803.08784}, 2018.

\bibitem[Boutilier et~al.(2013)Boutilier, Friedman, Goldszmidt, and Koller]{boutilier2013contextspecific}
Boutilier, C., Friedman, N., Goldszmidt, M., and Koller, D.
\newblock Context-specific independence in bayesian networks.
\newblock \emph{CoRR}, abs/1302.3562, 2013.

\bibitem[Brouillard et~al.(2020)Brouillard, Lachapelle, Lacoste, Lacoste-Julien, and Drouin]{brouillard2020differentiable}
Brouillard, P., Lachapelle, S., Lacoste, A., Lacoste-Julien, S., and Drouin, A.
\newblock Differentiable causal discovery from interventional data.
\newblock \emph{Advances in Neural Information Processing Systems}, 33:\penalty0 21865--21877, 2020.

\bibitem[Buesing et~al.(2019)Buesing, Weber, Zwols, Heess, Racaniere, Guez, and Lespiau]{buesing2018woulda}
Buesing, L., Weber, T., Zwols, Y., Heess, N., Racaniere, S., Guez, A., and Lespiau, J.-B.
\newblock Woulda, coulda, shoulda: Counterfactually-guided policy search.
\newblock In \emph{International Conference on Learning Representations}, 2019.

\bibitem[Camacho \& Alba(2013)Camacho and Alba]{camacho2013model}
Camacho, E.~F. and Alba, C.~B.
\newblock \emph{Model predictive control}.
\newblock Springer science \& business media, 2013.

\bibitem[Chitnis et~al.(2021)Chitnis, Silver, Kim, Kaelbling, and Lozano-Perez]{chitnis2021camps}
Chitnis, R., Silver, T., Kim, B., Kaelbling, L., and Lozano-Perez, T.
\newblock Camps: Learning context-specific abstractions for efficient planning in factored mdps.
\newblock In \emph{Conference on Robot Learning}, pp.\  64--79. PMLR, 2021.

\bibitem[Chung et~al.(2014)Chung, Gulcehre, Cho, and Bengio]{chung2014empirical}
Chung, J., Gulcehre, C., Cho, K., and Bengio, Y.
\newblock Empirical evaluation of gated recurrent neural networks on sequence modeling.
\newblock \emph{arXiv preprint arXiv:1412.3555}, 2014.

\bibitem[Dal et~al.(2018)Dal, Laarman, and Lucas]{dal2018parallel}
Dal, G.~H., Laarman, A.~W., and Lucas, P.~J.
\newblock Parallel probabilistic inference by weighted model counting.
\newblock In \emph{International Conference on Probabilistic Graphical Models}, pp.\  97--108. PMLR, 2018.

\bibitem[De~Haan et~al.(2019)De~Haan, Jayaraman, and Levine]{de2019causal}
De~Haan, P., Jayaraman, D., and Levine, S.
\newblock Causal confusion in imitation learning.
\newblock \emph{Advances in Neural Information Processing Systems}, 32, 2019.

\bibitem[Ding et~al.(2022)Ding, Lin, Li, and Zhao]{ding2022generalizing}
Ding, W., Lin, H., Li, B., and Zhao, D.
\newblock Generalizing goal-conditioned reinforcement learning with variational causal reasoning.
\newblock In \emph{Advances in Neural Information Processing Systems}, 2022.

\bibitem[Edwards \& Toma(1985)Edwards and Toma]{edwards1985fast}
Edwards, D. and Toma, H.
\newblock A fast procedure for model search in multidimensional contingency tables.
\newblock \emph{Biometrika}, 72:\penalty0 339--351, 1985.

\bibitem[Feng \& Magliacane(2023)Feng and Magliacane]{feng2023learning}
Feng, F. and Magliacane, S.
\newblock Learning dynamic attribute-factored world models for efficient multi-object reinforcement learning.
\newblock In \emph{Advances in Neural Information Processing Systems}, volume~36, pp.\  19117--19144, 2023.

\bibitem[Feng et~al.(2022)Feng, Huang, Zhang, and Magliacane]{feng2022factored}
Feng, F., Huang, B., Zhang, K., and Magliacane, S.
\newblock Factored adaptation for non-stationary reinforcement learning.
\newblock In \emph{Advances in Neural Information Processing Systems}, 2022.

\bibitem[Goyal et~al.(2021{\natexlab{a}})Goyal, Didolkar, Ke, Blundell, Beaudoin, Heess, Mozer, and Bengio]{goyal2021neural}
Goyal, A., Didolkar, A.~R., Ke, N.~R., Blundell, C., Beaudoin, P., Heess, N., Mozer, M.~C., and Bengio, Y.
\newblock Neural production systems.
\newblock In \emph{Advances in Neural Information Processing Systems}, 2021{\natexlab{a}}.

\bibitem[Goyal et~al.(2021{\natexlab{b}})Goyal, Lamb, Gampa, Beaudoin, Blundell, Levine, Bengio, and Mozer]{goyal2021factorizing}
Goyal, A., Lamb, A., Gampa, P., Beaudoin, P., Blundell, C., Levine, S., Bengio, Y., and Mozer, M.~C.
\newblock Factorizing declarative and procedural knowledge in structured, dynamical environments.
\newblock In \emph{International Conference on Learning Representations}, 2021{\natexlab{b}}.

\bibitem[Goyal et~al.(2021{\natexlab{c}})Goyal, Lamb, Hoffmann, Sodhani, Levine, Bengio, and Sch{\"o}lkopf]{goyal2021recurrent}
Goyal, A., Lamb, A., Hoffmann, J., Sodhani, S., Levine, S., Bengio, Y., and Sch{\"o}lkopf, B.
\newblock Recurrent independent mechanisms.
\newblock In \emph{International Conference on Learning Representations}, 2021{\natexlab{c}}.

\bibitem[Hoey et~al.(1999)Hoey, St-Aubin, Hu, and Boutilier]{hoey1999spudd}
Hoey, J., St-Aubin, R., Hu, A., and Boutilier, C.
\newblock Spudd: stochastic planning using decision diagrams.
\newblock In \emph{Proceedings of the Fifteenth conference on Uncertainty in artificial intelligence}, pp.\  279--288, 1999.

\bibitem[Huang et~al.(2022)Huang, Lu, Leqi, Hern{\'a}ndez-Lobato, Glymour, Sch{\"o}lkopf, and Zhang]{huang2022action}
Huang, B., Lu, C., Leqi, L., Hern{\'a}ndez-Lobato, J.~M., Glymour, C., Sch{\"o}lkopf, B., and Zhang, K.
\newblock Action-sufficient state representation learning for control with structural constraints.
\newblock In \emph{International Conference on Machine Learning}, pp.\  9260--9279. PMLR, 2022.

\bibitem[Hwang et~al.(2023)Hwang, Kwak, Song, Zhang, and Lee]{hwang2023on}
Hwang, I., Kwak, Y., Song, Y.-J., Zhang, B.-T., and Lee, S.
\newblock On discovery of local independence over continuous variables via neural contextual decomposition.
\newblock In \emph{Conference on Causal Learning and Reasoning}, pp.\  448--472. PMLR, 2023.

\bibitem[Jamshidi et~al.(2023)Jamshidi, Akbari, and Kiyavash]{jamshidi2024causal}
Jamshidi, F., Akbari, S., and Kiyavash, N.
\newblock Causal imitability under context-specific independence relations.
\newblock In \emph{Advances in Neural Information Processing Systems}, volume~36, pp.\  26810--26830, 2023.

\bibitem[Jang et~al.(2017)Jang, Gu, and Poole]{jang2016categorical}
Jang, E., Gu, S., and Poole, B.
\newblock Categorical reparameterization with gumbel-softmax.
\newblock In \emph{International Conference on Learning Representations}, 2017.

\bibitem[Kaiser et~al.(2020)Kaiser, Babaeizadeh, Mi{\l}os, Osi{\'n}ski, Campbell, Czechowski, Erhan, Finn, Kozakowski, Levine, et~al.]{Kaiser2020Model}
Kaiser, {\L}., Babaeizadeh, M., Mi{\l}os, P., Osi{\'n}ski, B., Campbell, R.~H., Czechowski, K., Erhan, D., Finn, C., Kozakowski, P., Levine, S., et~al.
\newblock Model based reinforcement learning for atari.
\newblock In \emph{International Conference on Learning Representations}, 2020.

\bibitem[Ke et~al.(2021)Ke, Didolkar, Mittal, Goyal, Lajoie, Bauer, Rezende, Mozer, Bengio, and Pal]{ke2021systematic}
Ke, N.~R., Didolkar, A.~R., Mittal, S., Goyal, A., Lajoie, G., Bauer, S., Rezende, D.~J., Mozer, M.~C., Bengio, Y., and Pal, C.
\newblock Systematic evaluation of causal discovery in visual model based reinforcement learning.
\newblock In \emph{Thirty-fifth Conference on Neural Information Processing Systems Datasets and Benchmarks Track (Round 2)}, 2021.

\bibitem[Kearns \& Koller(1999)Kearns and Koller]{kearns1999efficient}
Kearns, M. and Koller, D.
\newblock Efficient reinforcement learning in factored mdps.
\newblock In \emph{IJCAI}, volume~16, pp.\  740--747, 1999.

\bibitem[Killian et~al.(2022)Killian, Ghassemi, and Joshi]{killian2022counterfactually}
Killian, T.~W., Ghassemi, M., and Joshi, S.
\newblock Counterfactually guided policy transfer in clinical settings.
\newblock In \emph{Conference on Health, Inference, and Learning}, pp.\  5--31. PMLR, 2022.

\bibitem[Kipf et~al.(2020)Kipf, van~der Pol, and Welling]{Kipf2020Contrastive}
Kipf, T., van~der Pol, E., and Welling, M.
\newblock Contrastive learning of structured world models.
\newblock In \emph{International Conference on Learning Representations}, 2020.

\bibitem[Kumor et~al.(2021)Kumor, Zhang, and Bareinboim]{kumor2021sequential}
Kumor, D., Zhang, J., and Bareinboim, E.
\newblock Sequential causal imitation learning with unobserved confounders.
\newblock In \emph{Advances in Neural Information Processing Systems}, volume~34, pp.\  14669--14680, 2021.

\bibitem[Lee \& Bareinboim(2018)Lee and Bareinboim]{lee2018structural}
Lee, S. and Bareinboim, E.
\newblock Structural causal bandits: Where to intervene?
\newblock In \emph{Advances in Neural Information Processing Systems}, volume~31, 2018.

\bibitem[Lee \& Bareinboim(2020)Lee and Bareinboim]{lee2020characterizing}
Lee, S. and Bareinboim, E.
\newblock Characterizing optimal mixed policies: Where to intervene and what to observe.
\newblock In \emph{Advances in Neural Information Processing Systems}, volume~33, pp.\  8565--8576, 2020.

\bibitem[Li et~al.(2024)Li, Zhang, and Bareinboim]{li2024causally}
Li, M., Zhang, J., and Bareinboim, E.
\newblock Causally aligned curriculum learning.
\newblock In \emph{The Twelfth International Conference on Learning Representations}, 2024.

\bibitem[Li et~al.(2020)Li, Torralba, Anandkumar, Fox, and Garg]{li2020causal}
Li, Y., Torralba, A., Anandkumar, A., Fox, D., and Garg, A.
\newblock Causal discovery in physical systems from videos.
\newblock \emph{Advances in Neural Information Processing Systems}, 33:\penalty0 9180--9192, 2020.

\bibitem[L{\"o}we et~al.(2022)L{\"o}we, Madras, Zemel, and Welling]{lowe2022amortized}
L{\"o}we, S., Madras, D., Zemel, R., and Welling, M.
\newblock Amortized causal discovery: Learning to infer causal graphs from time-series data.
\newblock In \emph{Conference on Causal Learning and Reasoning}, pp.\  509--525. PMLR, 2022.

\bibitem[Lu et~al.(2018)Lu, Sch{\"o}lkopf, and Hern{\'a}ndez-Lobato]{lu2018deconfounding}
Lu, C., Sch{\"o}lkopf, B., and Hern{\'a}ndez-Lobato, J.~M.
\newblock Deconfounding reinforcement learning in observational settings.
\newblock \emph{arXiv preprint arXiv:1812.10576}, 2018.

\bibitem[Lu et~al.(2020)Lu, Huang, Wang, Hern{\'a}ndez-Lobato, Zhang, and Sch{\"o}lkopf]{lu2020sample}
Lu, C., Huang, B., Wang, K., Hern{\'a}ndez-Lobato, J.~M., Zhang, K., and Sch{\"o}lkopf, B.
\newblock Sample-efficient reinforcement learning via counterfactual-based data augmentation.
\newblock \emph{arXiv preprint arXiv:2012.09092}, 2020.

\bibitem[Lyle et~al.(2021)Lyle, Zhang, Jiang, Pineau, and Gal]{lyle2021resolving}
Lyle, C., Zhang, A., Jiang, M., Pineau, J., and Gal, Y.
\newblock Resolving causal confusion in reinforcement learning via robust exploration.
\newblock In \emph{Self-Supervision for Reinforcement Learning Workshop-ICLR}, volume 2021, 2021.

\bibitem[Maddison et~al.(2017)Maddison, Mnih, and Teh]{maddison2016concrete}
Maddison, C.~J., Mnih, A., and Teh, Y.~W.
\newblock The concrete distribution: A continuous relaxation of discrete random variables.
\newblock In \emph{International Conference on Learning Representations}, 2017.

\bibitem[Madumal et~al.(2020)Madumal, Miller, Sonenberg, and Vetere]{madumal2020explainable}
Madumal, P., Miller, T., Sonenberg, L., and Vetere, F.
\newblock Explainable reinforcement learning through a causal lens.
\newblock In \emph{Proceedings of the AAAI conference on artificial intelligence}, pp.\  2493--2500, 2020.

\bibitem[Mesnard et~al.(2021)Mesnard, Weber, Viola, Thakoor, Saade, Harutyunyan, Dabney, Stepleton, Heess, Guez, et~al.]{mesnard2021counterfactual}
Mesnard, T., Weber, T., Viola, F., Thakoor, S., Saade, A., Harutyunyan, A., Dabney, W., Stepleton, T.~S., Heess, N., Guez, A., et~al.
\newblock Counterfactual credit assignment in model-free reinforcement learning.
\newblock In \emph{International Conference on Machine Learning}, pp.\  7654--7664. PMLR, 2021.

\bibitem[Mutti et~al.(2023)Mutti, De~Santi, Rossi, Calderon, Bronstein, and Restelli]{mutti2023provably}
Mutti, M., De~Santi, R., Rossi, E., Calderon, J.~F., Bronstein, M., and Restelli, M.
\newblock Provably efficient causal model-based reinforcement learning for systematic generalization.
\newblock In \emph{Proceedings of the AAAI Conference on Artificial Intelligence}, pp.\  9251--9259, 2023.

\bibitem[Nair et~al.(2019)Nair, Zhu, Savarese, and Fei-Fei]{nair2019causal}
Nair, S., Zhu, Y., Savarese, S., and Fei-Fei, L.
\newblock Causal induction from visual observations for goal directed tasks.
\newblock \emph{arXiv preprint arXiv:1910.01751}, 2019.

\bibitem[Oberst \& Sontag(2019)Oberst and Sontag]{oberst2019counterfactual}
Oberst, M. and Sontag, D.
\newblock Counterfactual off-policy evaluation with gumbel-max structural causal models.
\newblock In \emph{International Conference on Machine Learning}, pp.\  4881--4890. PMLR, 2019.

\bibitem[Ozair et~al.(2021)Ozair, Li, Razavi, Antonoglou, Van Den~Oord, and Vinyals]{ozair2021vector}
Ozair, S., Li, Y., Razavi, A., Antonoglou, I., Van Den~Oord, A., and Vinyals, O.
\newblock Vector quantized models for planning.
\newblock In \emph{International Conference on Machine Learning}, pp.\  8302--8313. PMLR, 2021.

\bibitem[Pearl(2009)]{pearl2009causality}
Pearl, J.
\newblock \emph{Causality}.
\newblock Cambridge university press, 2009.

\bibitem[Peters et~al.(2017)Peters, Janzing, and Sch{\"o}lkopf]{peters2017elements}
Peters, J., Janzing, D., and Sch{\"o}lkopf, B.
\newblock \emph{Elements of causal inference: foundations and learning algorithms}.
\newblock The MIT Press, 2017.

\bibitem[Pitis et~al.(2020)Pitis, Creager, and Garg]{pitis2020counterfactual}
Pitis, S., Creager, E., and Garg, A.
\newblock Counterfactual data augmentation using locally factored dynamics.
\newblock \emph{Advances in Neural Information Processing Systems}, 33, 2020.

\bibitem[Pitis et~al.(2022)Pitis, Creager, Mandlekar, and Garg]{pitis2022mocoda}
Pitis, S., Creager, E., Mandlekar, A., and Garg, A.
\newblock Moco{DA}: Model-based counterfactual data augmentation.
\newblock In \emph{Advances in Neural Information Processing Systems}, 2022.

\bibitem[Poole(1998)]{poole1998context}
Poole, D.
\newblock Context-specific approximation in probabilistic inference.
\newblock In \emph{Proceedings of the Fourteenth conference on Uncertainty in artificial intelligence}, pp.\  447--454, 1998.

\bibitem[Ramsey et~al.(2006)Ramsey, Spirtes, and Zhang]{ramsey2006adjacency}
Ramsey, J., Spirtes, P., and Zhang, J.
\newblock Adjacency-faithfulness and conservative causal inference.
\newblock In \emph{Proceedings of the Twenty-Second Conference on Uncertainty in Artificial Intelligence}, pp.\  401--408, 2006.

\bibitem[Razavi et~al.(2019)Razavi, van~den Oord, and Vinyals]{razavi2019generating}
Razavi, A., van~den Oord, A., and Vinyals, O.
\newblock Generating diverse high-fidelity images with vq-vae-2.
\newblock In \emph{Advances in Neural Information Processing Systems}, volume~32, 2019.

\bibitem[Rezende et~al.(2020)Rezende, Danihelka, Papamakarios, Ke, Jiang, Weber, Gregor, Merzic, Viola, Wang, et~al.]{rezende2020causally}
Rezende, D.~J., Danihelka, I., Papamakarios, G., Ke, N.~R., Jiang, R., Weber, T., Gregor, K., Merzic, H., Viola, F., Wang, J., et~al.
\newblock Causally correct partial models for reinforcement learning.
\newblock \emph{arXiv preprint arXiv:2002.02836}, 2020.

\bibitem[Rubinstein \& Kroese(2004)Rubinstein and Kroese]{rubinstein2004cross}
Rubinstein, R.~Y. and Kroese, D.~P.
\newblock \emph{The cross-entropy method: a unified approach to combinatorial optimization, Monte-Carlo simulation, and machine learning}, volume 133.
\newblock Springer, 2004.

\bibitem[Sch{\"o}lkopf et~al.(2021)Sch{\"o}lkopf, Locatello, Bauer, Ke, Kalchbrenner, Goyal, and Bengio]{scholkopf2021toward}
Sch{\"o}lkopf, B., Locatello, F., Bauer, S., Ke, N.~R., Kalchbrenner, N., Goyal, A., and Bengio, Y.
\newblock Toward causal representation learning.
\newblock \emph{Proceedings of the IEEE}, 109\penalty0 (5):\penalty0 612--634, 2021.

\bibitem[Schrittwieser et~al.(2020)Schrittwieser, Antonoglou, Hubert, Simonyan, Sifre, Schmitt, Guez, Lockhart, Hassabis, Graepel, et~al.]{schrittwieser2020mastering}
Schrittwieser, J., Antonoglou, I., Hubert, T., Simonyan, K., Sifre, L., Schmitt, S., Guez, A., Lockhart, E., Hassabis, D., Graepel, T., et~al.
\newblock Mastering atari, go, chess and shogi by planning with a learned model.
\newblock \emph{Nature}, 588\penalty0 (7839):\penalty0 604--609, 2020.

\bibitem[Seitzer et~al.(2021)Seitzer, Sch{\"o}lkopf, and Martius]{seitzer2021causal}
Seitzer, M., Sch{\"o}lkopf, B., and Martius, G.
\newblock Causal influence detection for improving efficiency in reinforcement learning.
\newblock In \emph{Advances in Neural Information Processing Systems}, 2021.

\bibitem[Sontakke et~al.(2021)Sontakke, Mehrjou, Itti, and Sch{\"o}lkopf]{sontakke2021causal}
Sontakke, S.~A., Mehrjou, A., Itti, L., and Sch{\"o}lkopf, B.
\newblock Causal curiosity: Rl agents discovering self-supervised experiments for causal representation learning.
\newblock In \emph{International conference on machine learning}, pp.\  9848--9858. PMLR, 2021.

\bibitem[Spirtes et~al.(2000)Spirtes, Glymour, Scheines, and Heckerman]{spirtes2000causation}
Spirtes, P., Glymour, C.~N., Scheines, R., and Heckerman, D.
\newblock \emph{Causation, prediction, and search}.
\newblock MIT press, 2000.

\bibitem[Sutton \& Barto(2018)Sutton and Barto]{sutton2018reinforcement}
Sutton, R.~S. and Barto, A.~G.
\newblock \emph{Reinforcement learning: An introduction}.
\newblock MIT press, 2018.

\bibitem[Takida et~al.(2022)Takida, Shibuya, Liao, Lai, Ohmura, Uesaka, Murata, Takahashi, Kumakura, and Mitsufuji]{takida2022sq}
Takida, Y., Shibuya, T., Liao, W., Lai, C.-H., Ohmura, J., Uesaka, T., Murata, N., Takahashi, S., Kumakura, T., and Mitsufuji, Y.
\newblock Sq-vae: Variational bayes on discrete representation with self-annealed stochastic quantization.
\newblock In \emph{International Conference on Machine Learning}, pp.\  20987--21012. PMLR, 2022.

\bibitem[Tikka et~al.(2019)Tikka, Hyttinen, and Karvanen]{tikka2020identifying}
Tikka, S., Hyttinen, A., and Karvanen, J.
\newblock Identifying causal effects via context-specific independence relations.
\newblock \emph{Advances in Neural Information Processing Systems}, 32:\penalty0 2804--2814, 2019.

\bibitem[Tomar et~al.(2021)Tomar, Zhang, Calandra, Taylor, and Pineau]{tomar2021model}
Tomar, M., Zhang, A., Calandra, R., Taylor, M.~E., and Pineau, J.
\newblock Model-invariant state abstractions for model-based reinforcement learning.
\newblock \emph{arXiv preprint arXiv:2102.09850}, 2021.

\bibitem[Van Den~Oord et~al.(2017)Van Den~Oord, Vinyals, et~al.]{van2017neural}
Van Den~Oord, A., Vinyals, O., et~al.
\newblock Neural discrete representation learning.
\newblock \emph{Advances in neural information processing systems}, 30, 2017.

\bibitem[Volodin et~al.(2020)Volodin, Wichers, and Nixon]{volodin2020resolving}
Volodin, S., Wichers, N., and Nixon, J.
\newblock Resolving spurious correlations in causal models of environments via interventions.
\newblock \emph{arXiv preprint arXiv:2002.05217}, 2020.

\bibitem[Wang et~al.(2021)Wang, Xiao, Zhu, and Stone]{wang2021task}
Wang, Z., Xiao, X., Zhu, Y., and Stone, P.
\newblock Task-independent causal state abstraction.
\newblock In \emph{Proceedings of the 35th International Conference on Neural Information Processing Systems, Robot Learning workshop}, 2021.

\bibitem[Wang et~al.(2022)Wang, Xiao, Xu, Zhu, and Stone]{wang2022causal}
Wang, Z., Xiao, X., Xu, Z., Zhu, Y., and Stone, P.
\newblock Causal dynamics learning for task-independent state abstraction.
\newblock In \emph{International Conference on Machine Learning}, pp.\  23151--23180. PMLR, 2022.

\bibitem[Wang et~al.(2023)Wang, Hu, Stone, and Mart{\'\i}n-Mart{\'\i}n]{wang2023elden}
Wang, Z., Hu, J., Stone, P., and Mart{\'\i}n-Mart{\'\i}n, R.
\newblock {ELDEN}: Exploration via local dependencies.
\newblock In \emph{Thirty-seventh Conference on Neural Information Processing Systems}, 2023.

\bibitem[Williams et~al.(2020)Williams, Ringer, Ash, MacLeod, Dougherty, and Hughes]{williams2020hierarchical}
Williams, W., Ringer, S., Ash, T., MacLeod, D., Dougherty, J., and Hughes, J.
\newblock Hierarchical quantized autoencoders.
\newblock \emph{Advances in Neural Information Processing Systems}, 33:\penalty0 4524--4535, 2020.

\bibitem[Yang et~al.(2018)Yang, Katcoff, and Uhler]{yang2018characterizing}
Yang, K., Katcoff, A., and Uhler, C.
\newblock Characterizing and learning equivalence classes of causal dags under interventions.
\newblock In \emph{International Conference on Machine Learning}, pp.\  5541--5550. PMLR, 2018.

\bibitem[Yao et~al.(2022)Yao, Chen, and Zhang]{yao2022learning}
Yao, W., Chen, G., and Zhang, K.
\newblock Learning latent causal dynamics.
\newblock \emph{arXiv preprint arXiv:2202.04828}, 2022.

\bibitem[Yoon et~al.(2023)Yoon, Wu, Bae, and Ahn]{yoon2023investigation}
Yoon, J., Wu, Y.-F., Bae, H., and Ahn, S.
\newblock An investigation into pre-training object-centric representations for reinforcement learning.
\newblock In \emph{International Conference on Machine Learning}, pp.\  40147--40174. PMLR, 2023.

\bibitem[Zadaianchuk et~al.(2021)Zadaianchuk, Seitzer, and Martius]{zadaianchuk2021selfsupervised}
Zadaianchuk, A., Seitzer, M., and Martius, G.
\newblock Self-supervised visual reinforcement learning with object-centric representations.
\newblock In \emph{International Conference on Learning Representations}, 2021.

\bibitem[Zhang et~al.(2019)Zhang, Lipton, Pineda, Azizzadenesheli, Anandkumar, Itti, Pineau, and Furlanello]{zhang2019learning}
Zhang, A., Lipton, Z.~C., Pineda, L., Azizzadenesheli, K., Anandkumar, A., Itti, L., Pineau, J., and Furlanello, T.
\newblock Learning causal state representations of partially observable environments.
\newblock \emph{arXiv preprint arXiv:1906.10437}, 2019.

\bibitem[Zhang et~al.(2020{\natexlab{a}})Zhang, Lyle, Sodhani, Filos, Kwiatkowska, Pineau, Gal, and Precup]{zhang2020invariant}
Zhang, A., Lyle, C., Sodhani, S., Filos, A., Kwiatkowska, M., Pineau, J., Gal, Y., and Precup, D.
\newblock Invariant causal prediction for block mdps.
\newblock In \emph{International Conference on Machine Learning}, pp.\  11214--11224. PMLR, 2020{\natexlab{a}}.

\bibitem[Zhang et~al.(2020{\natexlab{b}})Zhang, Kumor, and Bareinboim]{zhang2020causal}
Zhang, J., Kumor, D., and Bareinboim, E.
\newblock Causal imitation learning with unobserved confounders.
\newblock \emph{Advances in neural information processing systems}, 33:\penalty0 12263--12274, 2020{\natexlab{b}}.

\bibitem[Zhang \& Poole(1999)Zhang and Poole]{zhang1999role}
Zhang, N.~L. and Poole, D.
\newblock On the role of context-specific independence in probabilistic inference.
\newblock In \emph{16th International Joint Conference on Artificial Intelligence, IJCAI 1999, Stockholm, Sweden}, volume~2, pp.\  1288, 1999.

\bibitem[Zholus et~al.(2022)Zholus, Ivchenkov, and Panov]{zholus2022factorized}
Zholus, A., Ivchenkov, Y., and Panov, A.
\newblock Factorized world models for learning causal relationships.
\newblock In \emph{ICLR2022 Workshop on the Elements of Reasoning: Objects, Structure and Causality}, 2022.

\bibitem[Zhu et~al.(2020)Zhu, Wong, Mandlekar, Mart{\'\i}n-Mart{\'\i}n, Joshi, Nasiriany, and Zhu]{zhu2020robosuite}
Zhu, Y., Wong, J., Mandlekar, A., Mart{\'\i}n-Mart{\'\i}n, R., Joshi, A., Nasiriany, S., and Zhu, Y.
\newblock robosuite: A modular simulation framework and benchmark for robot learning.
\newblock \emph{arXiv preprint arXiv:2009.12293}, 2020.

\end{thebibliography}
